%% file: main_arxiv.tex
\documentclass[10pt,twocolumn,letterpaper]{article}

\usepackage{cvpr}              %

\usepackage{graphicx}
\usepackage{amsmath}
\usepackage{amssymb}
\usepackage{booktabs}
\usepackage{mathtools}
\usepackage{esvect}

\usepackage[export]{adjustbox}
\usepackage{pbox}
\usepackage[accsupp]{axessibility}  %

\usepackage{color}
\definecolor{citecolor}{RGB}{30,130,255}
\usepackage[pagebackref=true,breaklinks=true,letterpaper=true,colorlinks,citecolor=citecolor,bookmarks=false]{hyperref}

\usepackage[capitalize]{cleveref}
\crefname{section}{Sec.}{Secs.}
\Crefname{section}{Section}{Sections}
\Crefname{table}{Table}{Tables}
\crefname{table}{Tab.}{Tabs.}

\def\cvprPaperID{****}
\def\confName{CVPR}
\def\confYear{2023}

\newlength{\ww}

\newcommand\blfootnote[1]{%
  \begingroup
  \renewcommand\thefootnote{}\footnote{#1}%
  \addtocounter{footnote}{-1}%
  \endgroup
}

\begin{document}

\newcommand{\name}{SpaText\xspace}

\title{\name: Spatio-Textual Representation for Controllable Image Generation}

\author
{
    Omri Avrahami$^{1,2}$ 
    \hspace{5.5mm} Thomas Hayes$^{1}$
    \hspace{5.5mm} Oran Gafni$^{1}$
    \hspace{5.5mm} Sonal Gupta$^{1}$
    \\
    \hspace{1mm} Yaniv Taigman$^{1}$
    \hspace{1mm} Devi Parikh$^{1}$
    \hspace{1mm} Dani Lischinski$^{2}$
    \hspace{1mm} Ohad Fried$^{3}$
    \hspace{1mm} Xi Yin$^{1}$
    \vspace{1mm}
    \\
    \normalsize{$^{1}$Meta AI}
    \hspace{5mm} \normalsize{$^{2}$The Hebrew University of Jerusalem}
    \hspace{5mm} \normalsize{$^{3}$Reichman University}
}

\def\ShowNotes{}
\input{macros.tex}
\twocolumn[{
\renewcommand\twocolumn[1][]{#1}%
\maketitle
\input{figures/method_examples/teaser.tex}
}]

\input{sections/abstract.tex}
\blfootnote{Project page is available at: \href{https://omriavrahami.com/spatext}{https://omriavrahami.com/spatext}}
\input{sections/introduction.tex}
\input{sections/related_work.tex}
\input{sections/method.tex}
\input{sections/experiments.tex}
\input{sections/limitations.tex}

\textbf{Acknowledgments} We thank Uriel Singer, Adam Polyak, Yuval Kirstain, Shelly Sheynin and Oron Ashual for their valuable help and feedback. This work was supported in part by the Israel Science Foundation (grants No. 1574/21, 2492/20, and 3611/21).

\appendix
\input{sections/appendix/additional_examples.tex}
\input{sections/appendix/implementation_details.tex}
\input{sections/appendix/additional_experiments.tex}
\input{sections/appendix/additional_related_work.tex}

\clearpage
{\small
\bibliographystyle{ieee_fullname}
\bibliography{egbib}
}

\end{document}

%% file: macros.tex
\newcommand{\ignorethis}[1]{}
\newcommand{\redund}[1]{#1}

\newcommand{\apriori    }     {\textit{a~priori}}
\newcommand{\aposteriori}     {\textit{a~posteriori}}
\newcommand{\perse      }     {\textit{per~se}}
\newcommand{\naive      }     {{na\"{\i}ve}}
\newcommand{\Naive      }     {{Na\"{\i}ve}}

\newcommand{\Identity   }     {\mat{I}}
\newcommand{\Zero       }     {\mathbf{0}}
\newcommand{\Reals      }     {{\textrm{I\kern-0.18em R}}}
\newcommand{\isdefined  }     {\mbox{\hspace{0.5ex}:=\hspace{0.5ex}}}
\newcommand{\texthalf   }     {\ensuremath{\textstyle\frac{1}{2}}}
\newcommand{\half       }     {\ensuremath{\frac{1}{2}}}
\newcommand{\third      }     {\ensuremath{\frac{1}{3}}}
\newcommand{\fourth     }     {\ensuremath{\frac{1}{4}}}

\newcommand{\Lone} {\ensuremath{L_1}}
\newcommand{\Ltwo} {\ensuremath{L_2}}

\newcommand{\mat        } [1] {{\text{\boldmath $\mathbit{#1}$}}}
\newcommand{\Approx     } [1] {\widetilde{#1}}
\newcommand{\change     } [1] {\mbox{{\footnotesize $\Delta$} \kern-3pt}#1}

\newcommand{\Order      } [1] {O(#1)}
\newcommand{\set        } [1] {{\lbrace #1 \rbrace}}
\newcommand{\floor      } [1] {{\lfloor #1 \rfloor}}
\newcommand{\ceil       } [1] {{\lceil  #1 \rceil }}
\newcommand{\inverse    } [1] {{#1}^{-1}}
\newcommand{\transpose  } [1] {{#1}^\mathrm{T}}
\newcommand{\invtransp  } [1] {{#1}^{-\mathrm{T}}}
\newcommand{\relu       } [1] {{\lbrack #1 \rbrack_+}}

\newcommand{\abs        } [1] {{| #1 |}}
\newcommand{\Abs        } [1] {{\left| #1 \right|}}
\newcommand{\norm       } [1] {{\| #1 \|}}
\newcommand{\Norm       } [1] {{\left\| #1 \right\|}}
\newcommand{\pnorm      } [2] {\norm{#1}_{#2}}
\newcommand{\Pnorm      } [2] {\Norm{#1}_{#2}}
\newcommand{\inner      } [2] {{\langle {#1} \, | \, {#2} \rangle}}
\newcommand{\Inner      } [2] {{\left\langle \begin{array}{@{}c|c@{}}
                               \displaystyle {#1} & \displaystyle {#2}
                               \end{array} \right\rangle}}

\newcommand{\twopartdef}[4]
{
  \left\{
  \begin{array}{ll}
    #1 & \mbox{if } #2 \\
    #3 & \mbox{if } #4
  \end{array}
  \right.
}

\newcommand{\fourpartdef}[8]
{
  \left\{
  \begin{array}{ll}
    #1 & \mbox{if } #2 \\
    #3 & \mbox{if } #4 \\
    #5 & \mbox{if } #6 \\
    #7 & \mbox{if } #8
  \end{array}
  \right.
}

\newcommand{\len}[1]{\text{len}(#1)}

\newlength{\w}
\newlength{\h}
\newlength{\x}

\definecolor{darkred}{rgb}{0.7,0.1,0.1}
\definecolor{darkgreen}{rgb}{0.1,0.6,0.1}
\definecolor{cyan}{rgb}{0.7,0.0,0.7}
\definecolor{otherblue}{rgb}{0.1,0.4,0.8}
\definecolor{maroon}{rgb}{0.76,.13,.28}
\definecolor{burntorange}{rgb}{0.81,.33,0}

\ifdefined\ShowNotes
  \newcommand{\colornote}[3]{{\color{#1}\textbf{#2} #3\normalfont}}
\else
  \newcommand{\colornote}[3]{}
\fi

\newcommand {\todo}[1]{\colornote{cyan}{TODO}{#1}}
\newcommand {\ohad}[1]{\colornote{otherblue}{Ohad:}{#1}}
\newcommand {\dani}[1]{\colornote{magenta}{Dani:}{#1}}
\newcommand {\omri}[1]{\colornote{burntorange}{Omri:}{#1}}
\newcommand {\xiyin}[1]{\colornote{darkred}{Xi:}{#1}}
\newcommand {\tom}[1]{\colornote{red}{Tom:}{#1}}
\newcommand {\oran}[1]{\colornote{darkgreen}{Oran:}{#1}}
\newcommand {\sonal}[1]{\colornote{blue}{Sonal:}{#1}}
\newcommand {\yaniv}[1]{\colornote{green}{Yaniv:}{#1}}
\newcommand {\devi}[1]{\colornote{otherblue}{Devi:}{#1}}

\definecolor{maskacolor}{RGB}{242,78,112}
\newcommand {\maska}[1]{{\color{maskacolor}{#1}}}
\definecolor{maskbcolor}{RGB}{255,176,1}
\newcommand {\maskb}[1]{{\color{maskbcolor}{#1}}}
\definecolor{maskccolor}{RGB}{41,160,177}
\newcommand {\maskc}[1]{{\color{maskccolor}{#1}}}
\definecolor{maskdcolor}{RGB}{55, 146, 55}
\newcommand {\maskd}[1]{{\color{maskdcolor}{#1}}}

\newcommand {\reqs}[1]{\colornote{red}{\tiny #1}}

\newcommand {\new}[1]{{\color{red}{#1}}}

\newcommand*\rot[1]{\rotatebox{90}{#1}}

\newcommand {\newstuff}[1]{#1}

\newcommand\todosilent[1]{}

\newcommand{\woBGmask}{{w/o~bg~\&~mask}}
\newcommand{\woMask}{{w/o~mask}}

\providecommand{\keywords}[1]
{
  \textbf{\textit{Keywords---}} #1
}

\newcommand{\clipdim}{\text{d}_{\text{CLIP}}}
\newcommand{\clipimgd}{\text{CLIP}_{\text{img}}}
\newcommand{\cliptxtd}{\text{CLIP}_{\text{txt}}}
\newcommand{\tglobal}{t_\text{global}}
\newcommand{\tlocal}{t_\text{local}}
\newcommand{\clipimg}[1]{\ensuremath{\text{CLIP}_{\text{img}}({#1})}}
\newcommand{\cliptxt}[1]{\ensuremath{\text{CLIP}_{\text{txt}}({#1})}}
\newcommand{\DALLE}{{DALL$\cdot$E}}

%% file: figures/method_examples/teaser.tex
\centering
\captionsetup{type=figure}
\setlength{\tabcolsep}{1pt}
\renewcommand{\arraystretch}{0.5}
\setlength{\ww}{0.32\columnwidth}
\vspace{-2.2em}
\begin{tabular}{cccccc}

    &&&&
    \scriptsize{``in the style of}
    \\
    \scriptsize{``in the forest''} &&
    \scriptsize{``on the moon''} &&
    \scriptsize{The Starry Night''}
    \\

    \includegraphics[width=\ww,frame]{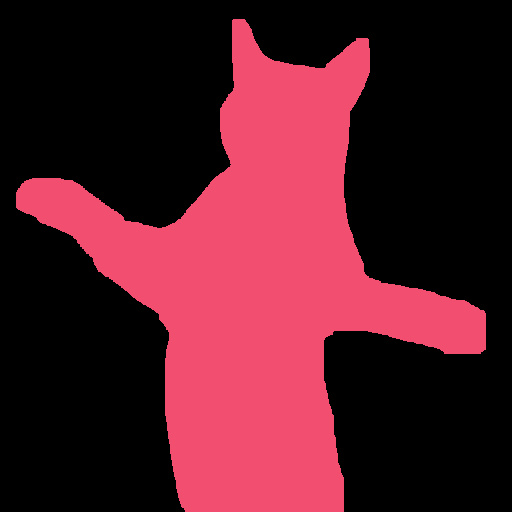} & 
    \includegraphics[width=\ww,frame]{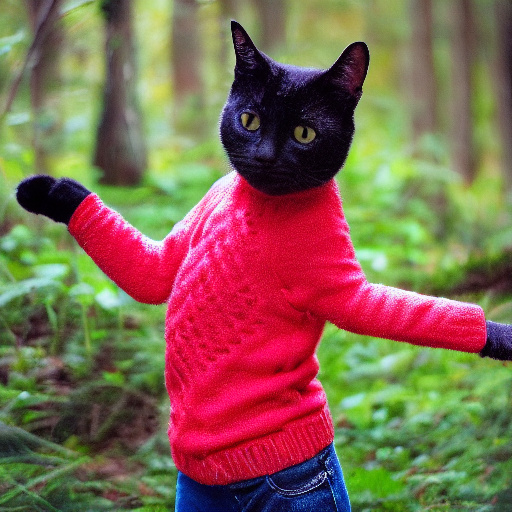} 
    \phantom{a}
    &
    
    \includegraphics[width=\ww,frame]{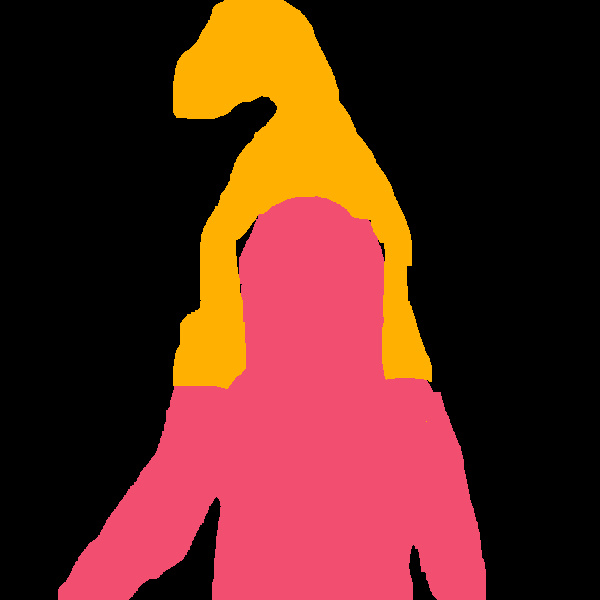} & 
    \includegraphics[width=\ww,frame]{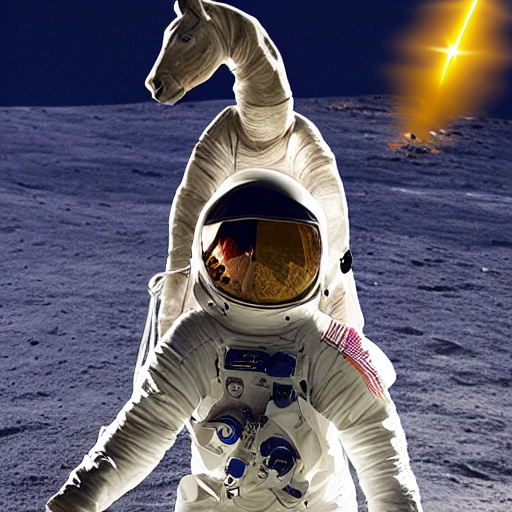}
    \phantom{a}
    &

    \includegraphics[width=\ww,frame]{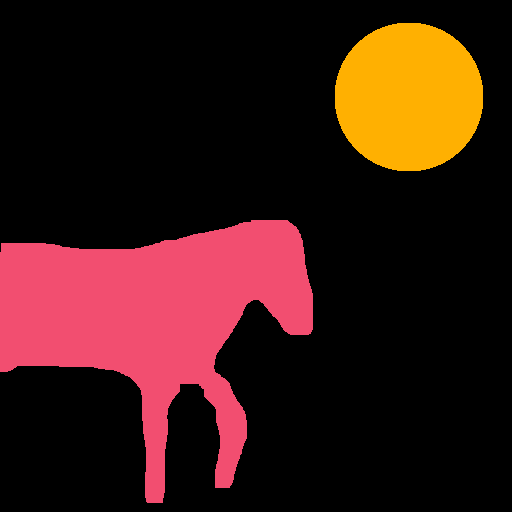} & 
    \includegraphics[width=\ww,frame]{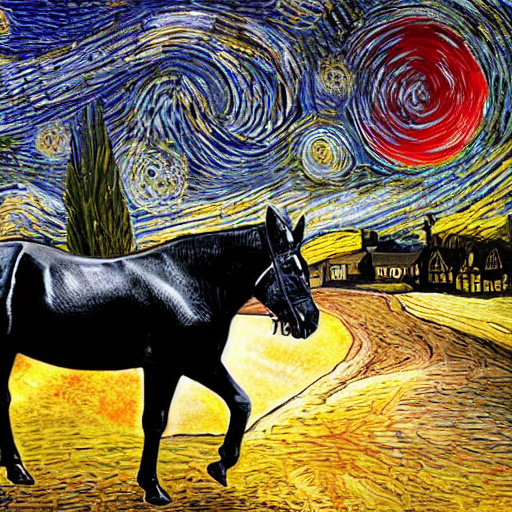}
    \\

    \begin{tabular}{c}
        \maska{\scriptsize{``a black cat with a red}} \\ 
        \maska{\scriptsize{sweater and a blue jeans''}} \\
        \\
    \end{tabular} &&

    \begin{tabular}{c}
        \maska{\scriptsize{``an astronaut''}} \\ 
        \maskb{\scriptsize{``a horse''}} \\
        \\
    \end{tabular} &&

    \begin{tabular}{c}
        \maska{\scriptsize{``a black horse''}} \\ 
        \maskb{\scriptsize{``a red full moon''}} \\
        \\
    \end{tabular} 
    \\

    \scriptsize{``in an empty room''} &&
    \scriptsize{``on a snowy day''} &&
    \scriptsize{``at the beach''}
    \\

    \includegraphics[width=\ww,frame]{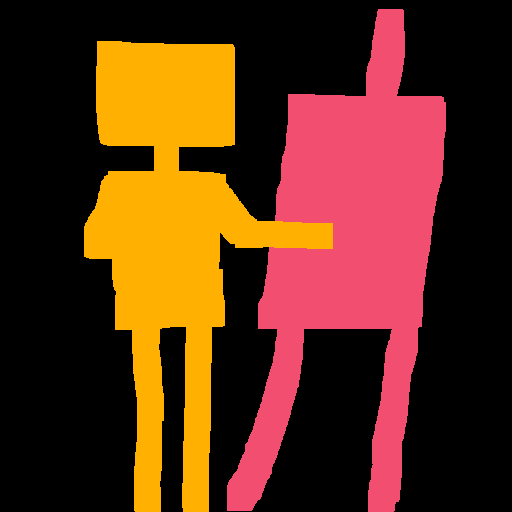} & 
    \includegraphics[width=\ww,frame]{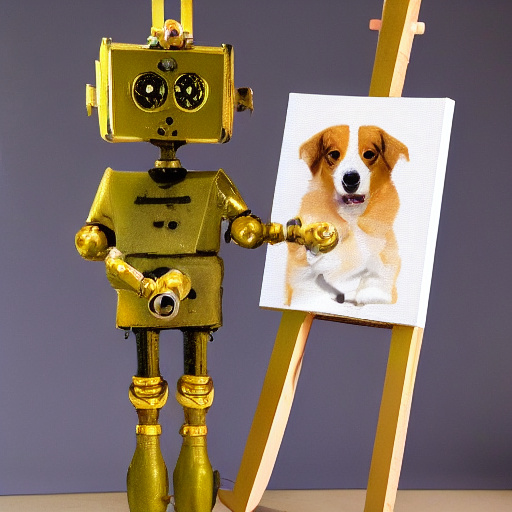} 
    \phantom{a}
    &
    
    \includegraphics[width=\ww,frame]{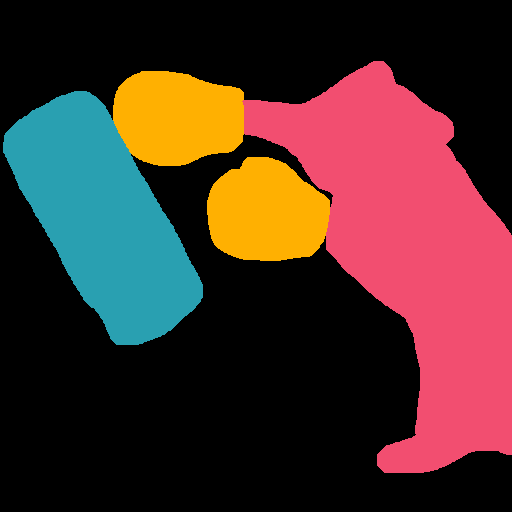} & 
    \includegraphics[width=\ww,frame]{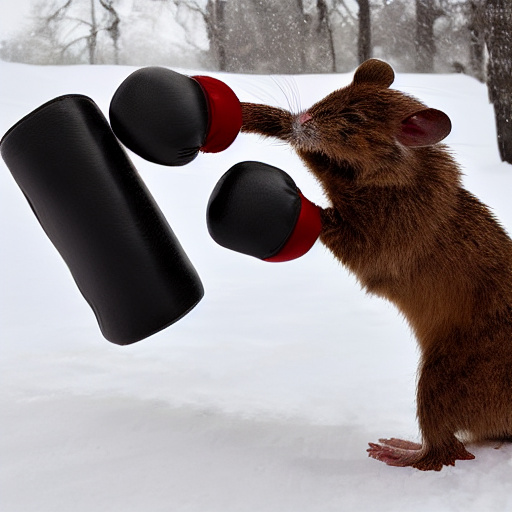}
    \phantom{a}
    &

    \includegraphics[width=\ww,frame]{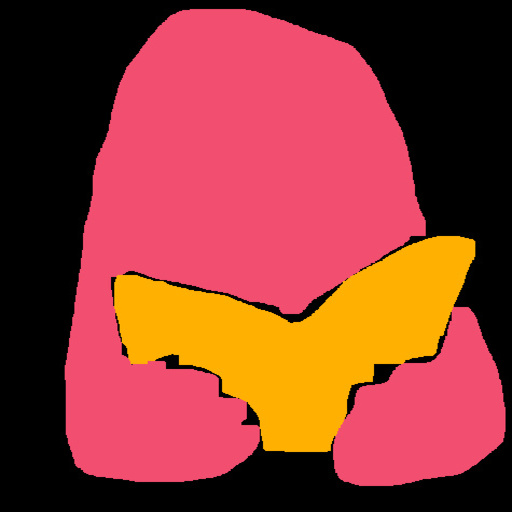} & 
    \includegraphics[width=\ww,frame]{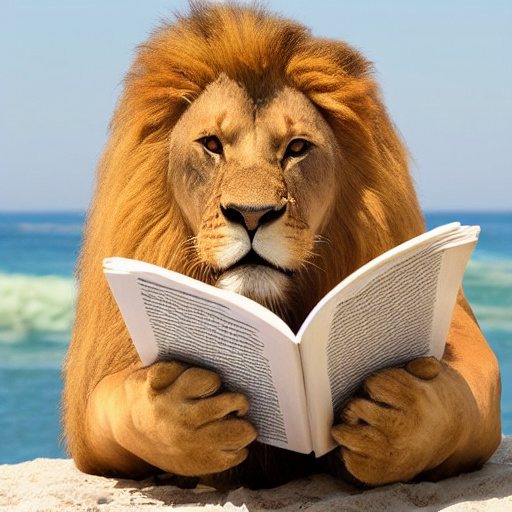}
    \\

    \begin{tabular}{c}
        \maska{\scriptsize{``a canvas with a painting}} \\ 
        \maska{\scriptsize{of a Corgi dog''}} \\
        \maskb{\scriptsize{``a metallic yellow robot''}} \\
        \\
    \end{tabular} &&

    \begin{tabular}{c}
        \maska{\scriptsize{``a mouse''}} \\ 
        \maskb{\scriptsize{``boxing gloves''}} \\
        \maskc{\scriptsize{``a black punching bag''}} \\
        \\
    \end{tabular} &&

    \begin{tabular}{c}
        \maska{\scriptsize{``a lion''}} \\ 
        \maskb{\scriptsize{``a book''}} \\
        \\
        \\
    \end{tabular} 
    \\
\end{tabular}
\vspace{-1em}
\captionof{figure}{Samples of generated images from input text and our proposed spatio-textual representations. Each pair consists of an (i) input global text (top left, black), a spatio-textual representation describing each segment using free-form text prompts (left, colored text and sketches), and (ii) the corresponding generated image (right). As can be seen, \name is able to generate high-quality images that correspond to both the global text and spatio-textual representation content. 
(The colors are for illustration purposes only, and do not affect the actual inputs.)}
\phantom{.}
\label{fig:teaser}

%% file: sections/abstract.tex
\begin{abstract}
    Recent text-to-image diffusion models are able to generate convincing results of unprecedented quality. However, it is nearly impossible to control the shapes of different regions/objects or their layout in a fine-grained fashion. Previous attempts to provide such controls were hindered by their reliance on a fixed set of labels. To this end, we present \name{} --- a new method for text-to-image generation using open-vocabulary scene control. In addition to a global text prompt that describes the entire scene, the user provides a segmentation map where each region of interest is annotated by a free-form natural language description. Due to lack of large-scale datasets that have a detailed textual description for each region in the image, we choose to leverage the current large-scale text-to-image datasets and base our approach on a novel CLIP-based spatio-textual representation, and show its effectiveness on two state-of-the-art diffusion models: pixel-based and latent-based. In addition, we show how to extend the classifier-free guidance method in diffusion models to the multi-conditional case and present an alternative accelerated inference algorithm. Finally, we offer several automatic evaluation metrics and use them, in addition to FID scores and a user study, to evaluate our method and show that it achieves state-of-the-art results on image generation with free-form textual scene control.
\end{abstract}

%% file: sections/introduction.tex
\vspace{-1.5em}
\section{Introduction}
\label{sec:introduction}

Imagine you could generate an image by dipping your digital paintbrush (so to speak) in a ``black horse'' paint, then sketching the specific position and posture of the horse, afterwards, dipping it again in a ``red full moon'' paint and sketching it the desired area. Finally, you want the entire image to be in the style of The Starry Night. Current state-of-the-art text-to-image models \cite{ramesh2022hierarchical, saharia2022photorealistic, yu2022scaling} leave much to be desired in achieving this vision.

The text-to-image interface is extremely powerful --- a single prompt is able to represent an infinite number of possible images. However, it has its cost --- on the one hand, it enables a novice user to explore an endless number of ideas, but, on the other hand, it limits controllability: if the user has a mental image that they wish to generate, with a specific layout of objects or regions in the image and their shapes, it is practically impossible to convey this information with text alone, as demonstrated in \Cref{fig:elaborated_description_failure}. In addition, inferring spatial relations \cite{yu2022scaling} from a single text prompt is one of the current limitations of SoTA models.

Make-A-Scene \cite{gafni2022make} proposed to tackle this problem by adding an additional (optional) input to text-to-image models, a \emph{dense} segmentation map with \emph{fixed} labels. The user can provide two inputs: a text prompt that describes the entire scene and an elaborate segmentation map that includes a label for each segment in the image. This way, the user can easily control the layout of the image. However, it suffers from the following drawbacks: (1) training the model with a fixed set of labels limits the quality for objects that are not in that set at inference time, (2) providing a dense segmentation can be cumbersome for users and undesirable in some cases, e.g., when the user prefers to provide a sketch for only a few main objects they care about, letting the model infer the rest of the layout; and (3) lack of fine-grained control over the specific characteristic of each instance. For example, even if the label set contains the label ``dog'', it is not clear how to generate several instances of dogs of different breeds in a single scene.

In order to tackle these drawbacks, we propose a different approach: (1) rather than using a fixed set of labels to represent each pixel in the segmentation map, we propose to represent it using \emph{spatial free-form text}, and (2) rather than providing a dense segmentation map accounting for each pixel, we propose to use a \emph{sparse} map, that describes only the objects that a user specifies (using spatial free-form text), while the rest of the scene remains unspecified. To summarize, we propose a new problem setting: given a \emph{global text prompt} that describes the entire image, and a spatio-textual scene that specifies for segments of interest their \emph{local text description} as well as their \emph{position and shape}, a corresponding image is generated, as illustrated in \Cref{fig:teaser}. These changes extend expressivity by providing the user with more control over the regions they care about, leaving the rest for the machine to figure out.

Acquiring a large-scale dataset that contains free-form textual descriptions for each segment in an image is prohibitively expensive, and such large-scale datasets do not exist to the best of our knowledge. Hence, we opt to extract the relevant information from existing image-text datasets. To this end, we propose a novel CLIP-based \cite{radford2021learning} spatio-textual representation that enables a user to specify for each segment its description using free-form text and its position and shape. During training, we extract \emph{local regions} using a pre-trained panoptic segmentation model \cite{wu2019detectron2}, and use them as input to a CLIP image encoder to create our representation. Then, at inference time, we use the \emph{text descriptions} provided by the user, embed them using a CLIP text encoder, and translate them to the CLIP image embedding space using a prior model \cite{ramesh2022hierarchical}.

\input{figures/elaborated_description_failure/fig.tex}

In order to assess the effectiveness of our proposed representation \name, we implement it on two state-of-the-art types of text-to-image diffusion models: a pixel-based model (\DALLE~2 \cite{ramesh2022hierarchical}) and a latent-based model (Stable Diffusion \cite{rombach2022high}). Both of these text-to-image models employ classifier-free guidance \cite{ho2022classifier} at inference time, which supports a single conditioning input (text prompt). In order to adapt them to our multi-conditional input (global text as well as the spatio-textual representation), we demonstrate how classifier-free guidance can be extended to any multi-conditional case. To the best of our knowledge, we are the first to demonstrate this. Furthermore, we propose an additional, faster variant of this extension that trades-off controllability for inference time.

Finally, we propose several automatic evaluation metrics for our problem setting and use them along with the FID score to evaluate our method against its baselines. In addition, we conduct a user-study and show that our method is also preferred by human evaluators.

In summary, our contributions are: (1) we address a new scenario of image generation with free-form textual scene control, (2) we propose a novel spatio-textual representation that for each segment represents its semantic properties and structure, and demonstrate its effectiveness on two state-of-the-art diffusion models --- pixel-based and latent-based, (3) we extend the classifier-free guidance in diffusion models to the multi-conditional case and present an alternative accelerated inference algorithm, and (4) we propose several automatic evaluation metrics and use them to compare against baselines we adapted from existing methods. We also evaluate via a user study. We find that our method achieves state-of-the-art results.

%% file: figures/elaborated_description_failure/fig.tex
\begin{figure}[t]
    \centering
    
    \centering
    \setlength{\tabcolsep}{1pt}
    \renewcommand{\arraystretch}{0.5}
    \setlength{\ww}{0.23\columnwidth}
  
    \begin{tabular}{cccc}

        \scriptsize{``at the beach''} &
        \scriptsize{\name} &
        \scriptsize{Stable Diffusion} &
        \scriptsize{\DALLE~2}
        \\

        \includegraphics[width=\ww,frame]{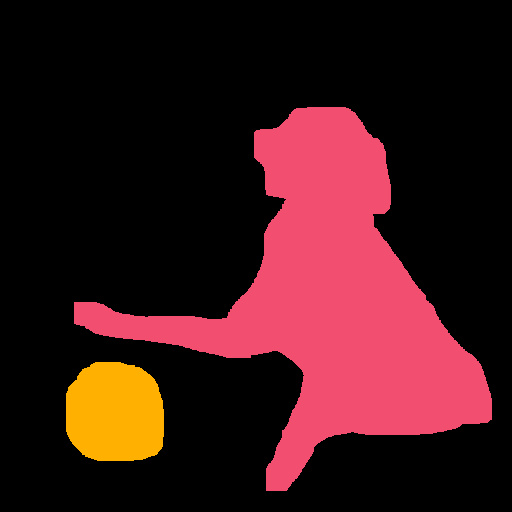} &
        \includegraphics[width=\ww,frame]{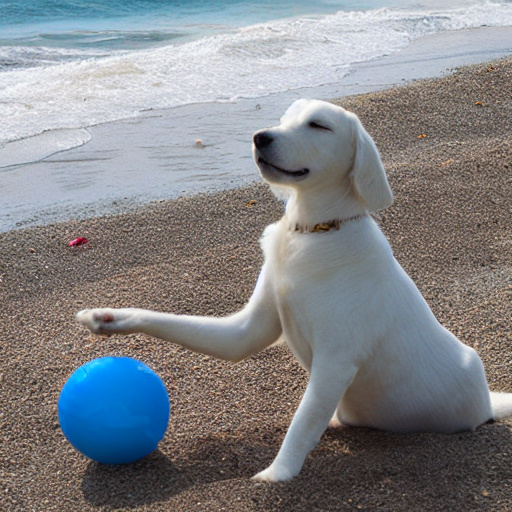}
        \phantom{.}
        &
        \includegraphics[width=\ww,frame]{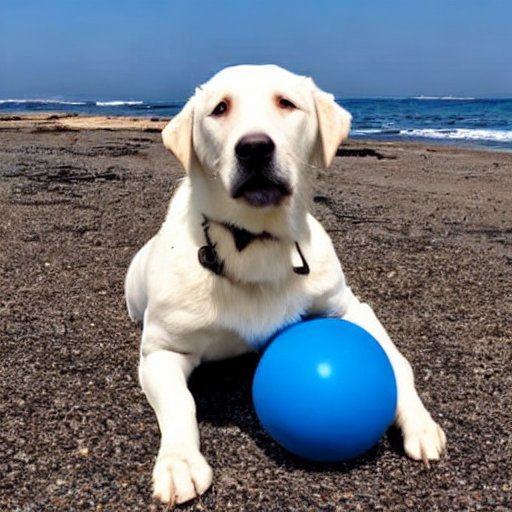} &
        \includegraphics[width=\ww,frame]{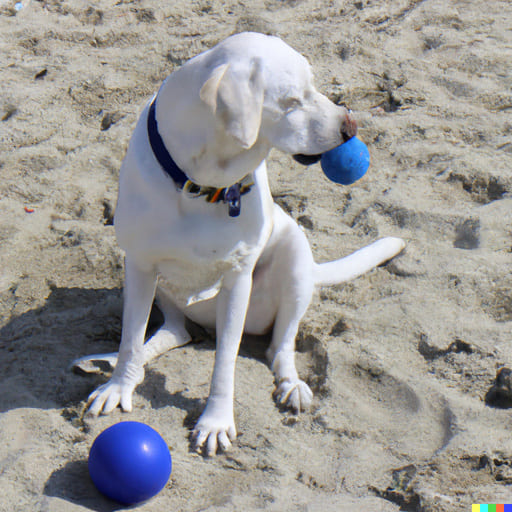}
        \\

        \maska{\scriptsize{``a white Labrador''}} &&
        \multicolumn{2}{c}{\scriptsize{``a white Labrador at the beach puts its}} \\

        \maskb{\scriptsize{``a blue ball''}} &&
        \multicolumn{2}{c}{\scriptsize{right arm above a blue ball without}} \\

        &&\multicolumn{2}{c}{\scriptsize{touching, while sitting in the bottom}} \\

        &&\multicolumn{2}{c}{\scriptsize{right corner of the frame''}} \\
    \end{tabular}
    
    \caption{\textbf{Lack of fine-grained spatial control:} A user with a specific mental image of a Labrador dog holding its paw above a blue ball without touching, can easily generate it with a \name representation (left) but will struggle to do so with traditional text-to-image models (right) \cite{rombach2022high,ramesh2021zero}.}
    \vspace{-1.5em}
    \label{fig:elaborated_description_failure}
\end{figure}

%% file: sections/related_work.tex
\section{Related Work}
\label{sec:related_work}

\textbf{Text-to-image generation.} Recently, we have witnessed great advances in the field of text-to-image generation. The seminal works based on RNNs \cite{hochreiter1997long, cho2014properties} and GANs \cite{goodfellow2014gans} produced promising low-resolution results \cite{reed2016generative, zhang2017stackgan, zhang2018stackgan++, xu2018attngan} in constrained domains (e.g., flowers \cite{nilsback2008automated} and birds \cite{wah2011caltech}). Later, zero-shot open-domain models were achieved using transformer-based \cite{vaswani2017attention} approaches: \DALLE~1 \cite{ramesh2021zero} and VQ-GAN \cite{esser2021taming} propose a two-stage approach by first training a discrete VAE \cite{kingma2013auto, van2017neural, razavi2019generating} to find a rich semantic space, then, at the second stage, they learn to model the joint distribution of text and image tokens autoregressively. CogView \cite{ding2021cogview, ding2022cogview2} and Parti \cite{yu2022scaling} also utilized a transformer model for this task. In parallel, diffusion based \cite{sohl2015deep, ho2020denoising, nichol2021improved, dhariwal2021diffusion, croitoru2022diffusion} text-to-image models were introduced: Latent Diffusion Models (LDMs) \cite{rombach2022high} performed the diffusion process on a lower-dimensional latent space instead on the pixel space. \DALLE~2 \cite{ramesh2022hierarchical} proposed to perform the diffusion process on the $\clipimgd$ space. Finally, Imagen \cite{saharia2022photorealistic} proposed to utilize a pre-trained T5 language model \cite{raffel2020exploring} for conditioning a pixel-based text-to-image diffusion model. Recently, retrieval-based models \cite{ashual2022knn, blattmann2022retrieval, rombach2022text, chen2022re} proposed to augment the text-to-image models using an external database of images. All these methods do not tackle the problem of image generation with free-form textual scene control.

\textbf{Scene-based text-to-image generation.} Image generation with scene control has been studied in the past \cite{reed2016learning, zhao2019image, sun2019image, sylvain2021object, sun2021learning, hong2018inferring, li2019object, hinz2018generating, hinz2020semantic, pavllo2020controlling, frolov2022dt2i}, but not with general masks and free-form text control. No Token Left Behind \cite{paiss2022no} proposed to leverage explainability-based method \cite{chefer2021generic, chefer2021transformer} for image generation with spatial conditioning using VQGAN-CLIP \cite{crowson2022vqgan} optimization. In addition, Make-A-Scene \cite{gafni2022make} proposed to add a \emph{dense} segmentation map using a \emph{fixed} set of labels to allow better controllability. We adapted these two approaches to our problem setting and compared our method against them.

\textbf{Local text-driven image editing.} Recently, various text-driven image editing methods were proposed \cite{bar2022text2live,hertz2022prompt, ruiz2022dreambooth, gal2022image, kawar2022imagic, chefer2021image, gal2022stylegan, patashnik2021styleclip, ashual2022knn, valevski2022unitune, couairon2022diffedit, ackermann2022high, kwon2022diffusion, kong2022leveraging, bau2021paint} that allow editing an existing image. Some of the methods support \emph{localized} image editing: GLIDE \cite{nichol2021glide} and \DALLE~2 \cite{ramesh2022hierarchical} train a designated inpainting model, whereas Blended Diffusion \cite{avrahami2022blended, avrahami2022blended_latent} leverages a pretrained text-to-image model. Combining these localized methods with a text-to-image model may enable scene-based image generation. We compare our method against this approach in the supplementary.

%% file: sections/method.tex
\section{Method}
\label{sec:method}

\input{figures/spatio_textual_representation/fig.tex}

We aim to provide the user with more fine-grained control over the generated image. In addition to a single \emph{global} text prompt, the user will also provide a segmentation map, where the content of each segment of interest is described using a \emph{local} free-form text prompt.

Formally, the input consists of a global text prompt $\tglobal$ that describes the scene in general, and a $H\!\times\!W$ raw spatio-textual matrix $RST$, where each entry $RST[i, j]$ contains the text description of the desired content in pixel $[i, j]$, or $\emptyset$ if the user does not wish to specify the content of this pixel in advance. Our goal is to synthesize an $H\!\times\!W$ image $I$ that complies with both the global text description $\tglobal$ and the raw spatio-textual scene matrix $RST$.

In \Cref{sec:spatio_textual_representation} we present our novel spatio-textual representation, which we use to tackle the problem of text-to-image generation with sparse scene control. Later, in \Cref{sec:sota_models_adaptaion} we explain how to incorporate this representation into two state-of-the-art text-to-image diffusion models. Finally, in \Cref{sec:multi_conditional_cfg} we present two ways for adapting classifier-free guidance to our multi-conditional problem.

\subsection{CLIP-based Spatio-Textual Representation}
\label{sec:spatio_textual_representation}

Over the recent years, large-scale text-to-image datasets were curated by the community, fueling the tremendous progress in this field. Nevertheless, these datasets cannot be \naive{}ly used for our task, because they do not contain \emph{local} text descriptions for each segment in the images. Hence, we need to develop a way to extract the objects in the image along with their textual description. To this end, we opt to use a pre-trained panoptic segmentation model \cite{wu2019detectron2} along with a CLIP \cite{radford2021learning} model.

CLIP was trained to embed images and text prompts into a rich shared latent space by contrastive learning on 400 million image-text pairs. We utilize this shared latent space for our task in the following way: during training we use the image encoder $\clipimgd$ to extract the local embeddings using the \emph{pixels} of the objects that we want to generate (because the local text descriptions are not available), whereas during inference we use the CLIP text encoder $\cliptxtd$ to extract the local embeddings using the \emph{text descriptions} provided by the user.

Hence, we build our spatio-textual representation, as depicted in \Cref{fig:spatio_textual_representation}: for each training image $x$ we first extract its panoptic segments $\{S_i \in [C] \}_{i=1}^{i=N}$ where $C$ is the number of panoptic segmentation classes and N is the number of segments for the current image. Next, we randomly choose $K$ disjoint segments $\{S_i \in [C] \}_{i=1}^{i=K}$. For each segment $S_i$, we crop a tight square around it, black-out the pixels in the square that are not in the segment (to avoid confusing the CLIP model with other content that might fall in the same square), resize it to the CLIP input size, and get the CLIP image embedding of that segment $\clipimg{S_i}$.

Now, for the training image $x$ we define the spatio-textual representation $ST_x$ of shape $(H, W, \clipdim)$ to be:
\begin{equation}
    ST_x[j, k] =
    \begin{cases}
        \clipimg{S_i} & \text{if $[i, k] \in S_i$} \\
        \vv{0} & \text{otherwise} \\
    \end{cases}
\end{equation}
where $\clipdim$ is the dimension of the CLIP shared latent space and $\vv{0}$ is the zero vector of dimension $\clipdim$.

During inference time, to form the raw spatio-textual matrix $RST$, we embed the local prompts using $\cliptxtd$ to the CLIP text embedding space. Next, in order to mitigate the domain gap between train and inference times, we convert these embeddings to $\clipimgd$ using a designated prior model $P$. The prior model $P$ was trained separately to convert CLIP text embeddings to CLIP image embeddings using an image-text paired dataset following \DALLE~2. Finally, as depicted in \Cref{fig:spatio_textual_representation} (right), we construct the spatio-textual representation $ST$ using these embeddings at pixels indicated by the user-supplied spatial map. For more implementation details, please refer to the supplementary.

\subsection{Incorporating Spatio-Textual Representation into SoTA Diffusion Models}
\label{sec:sota_models_adaptaion}

\input{figures/multiscale_control/fig.tex}

The current diffusion-based SoTA text-to-image models are \DALLE~2 \cite{ramesh2022hierarchical}, Imagen \cite{saharia2022photorealistic} and Stable Diffusion \cite{rombach2022high}. At the time of writing this paper, \DALLE~2 \cite{ramesh2022hierarchical} model architecture and weights are unavailable, hence we start by reimplementing \DALLE~2-like text-to-image model that consists of three diffusion-based models: (1) a \emph{prior model $P$} trained to translate the tuples $(\cliptxt{y}, \text{BytePairEncoding}(y))$ into $\clipimg{x}$ where $(x, y)$ is an image-text pair, (2) a \emph{decoder model $D$} that translates $\clipimg{x}$ into a low-resolution version of the image $x_{64\times64}$, and (3) a \emph{super-resolution model $\mathit{SR}$} that upsamples $x_{64\times64}$ into a higher resolution of $x_{256\times256}$. Concatenating the above three models yields a text-to-image model $\mathit{SR} \circ D \circ P$.
 
Now, in order to utilize the vast knowledge it has gathered during the training process, we opt to fine-tune a pre-trained text-to-image model in order to enable localized textual scene control by adapting its decoder component $D$. At each diffusion step, the decoder performs a single denoising step $x_t = D(x_{t-1}, \clipimg{x}, t)$ to get a less noisy version of $x_{t-1}$. In order to keep the spatial correspondence between the spatio-textual representation $ST$ and the noisy image $x_t$ at each stage, we choose to concatenate $x_t$ and $ST$ along the RGB channels dimensions, to get a total input of shape $(H, W, 3 + \clipdim)$. Now, we extend each kernel of the first convolutional layer from shape $(C_{in}, K_H, K_W)$ to $(C_{in} + \clipdim, K_H, K_W)$ by concatenating a tensor of dimension $\clipdim$ that we initialize with He initialization \cite{he2015delving}. Next, we fine-tuned the decoder using the standard simple loss variant of Ho \etal{} \cite{ho2020denoising}
$L_{\text{simple}} = E_{t,x_0,\epsilon}\left[ || \epsilon - \epsilon_{\theta}(x_t, \clipimg{x_0}, ST, t) ||^2 \right]$
where $\epsilon_{\theta}$ is a UNet \cite{long2015fully} model that predicts the added noise at each time step $t$, $x_t$ is the noisy image at time step $t$ and $ST$ is our spatio-textual representation. To this loss, we added the standard variational lower bound (VLB) loss \cite{nichol2021improved}.

Next, we move to handle the second family of SoTA diffusion-based text-to-image models: latent-based models. More specifically, we opt to adapt Stable Diffusion \cite{rombach2022high}, a recent open-source text-to-image model. This model consist of two parts: (1) an autoencoder $(Enc(x), Dec(z))$ that embeds the image $x$ into a lower-dimensional latent space $z$, and, (2) a diffusion model $A$ that performs the following denoising steps on the latent space $z_{t-1} = A(z_t, \cliptxt{t})$. The final denoised latent is fed to the decoder to get the final prediction $Dec(z_0)$.

We leverage the fact that the autoencoder $(Enc(x), Dec(z))$ is fully-convolutional, hence, the latent space $z$ \emph{corresponds spatially} to the generated image $x$, which means that we can concatenate the spatio-textual representation $ST$ the same way we did on the pixel-based model: concatenate the noisy latent $z_t$ and $ST$ along the channels dimensions, to get a total input of shape $(H, W, dim(z_t) + \clipdim)$ where $dim(z_t)$ is the number of feature channels. We initialize the newly-added channels in the kernels of the first convolutional layer using the same method we utilized for the pixel-based variant. Next, we fine-tune the denosing model by $L_{\text{LDM}} = E_{t, y, z_0,\epsilon}\left[ || \epsilon - \epsilon_{\theta}(z_t, \cliptxt{y},ST, t) ||^2 \right]$ where $z_t$ is the noisy latent code at time step $t$ and $y$ is the corresponding text prompt. For more implementation details of both models, please read the supplementary.

\subsection{Multi-Conditional Classifier-Free Guidance}
\label{sec:multi_conditional_cfg}

Classifier-free guidance \cite{ho2022classifier} is an inference method for conditional diffusion models which enables trading-off mode coverage and sample fidelity. It involves training a conditional and unconditional models simultaneously, and combining their predictions during inference. Formally, given a conditional diffusion model $\epsilon_\theta(x_t | c)$ where $c$ is the condition (e.g., a class label or a text prompt) and $x_t$ is the noisy sample, the condition $c$ is replaced by the null condition $\emptyset$ with a fixed probability during training. Then, during inference, we extrapolate towards the direction of the condition $\epsilon_\theta(x_t | c)$ and away from $\epsilon_\theta(x_t | \emptyset)$:
\begin{equation}
    \label{eqn:standard_cfg}
    \hat{\epsilon}_\theta(x_t | c) = \epsilon_\theta(x_t | \emptyset) + s \cdot (\epsilon_\theta(x_t | c) - \epsilon_\theta(x_t | \emptyset))
\end{equation}
where $s \ge 1$ is the guidance scale.

In order to adapt classifier-free guidance to our setting, we need to extend it to support multiple conditions.
Given a conditional diffusion model $\epsilon_\theta(x_t | \{c_i\}_{i=1}^{i=N})$ where $\{c_i\}_{i=1}^{i=N}$ are $N$ condition inputs, during training, we \emph{independently} replace each condition $c_i$ with the null condition $\emptyset$. Then, during inference, we calculate the direction of each condition $\Delta^t_i = \epsilon_\theta(x_t | c_i) - \epsilon_\theta(x_t | \emptyset)$ separately, and linearly combine them using $N$ guidance scales $s_i$ by extending \cref{eqn:standard_cfg}:
\begin{equation}
    \label{eqn:multiscale_cfg}
    \hat{\epsilon}_\theta(x_t | \{c_i\}_{i=1}^{i=N}) = \epsilon_\theta(x_t | \emptyset) + \sum_{i=1}^{i=N} s_i \Delta^t_i
\end{equation}
Using the above formulation, we are able to control each of the conditions separately during inference, as demonstrated in \Cref{fig:multiscale_control}. To the best of our knowledge, we are the first to demonstrate this effect in the multi-conditional case.

The main limitation of the above formulation is that its execution time grows linearly with the number of conditions, i.e., each denoising step requires $(N + 1)$ feed-forward executions: one for the null condition and $N$ for the other conditions. As a remedy, we propose a fast variant of the multi-conditional classifier-free guidance that trades-off the fine-grained controllability of the model with the inference speed: the training regime is identical to the previous variant, but during inference, we calculate only the direction of the joint probability of all the conditions $\Delta^t_{\text{joint}} = \epsilon_\theta(x_t | \{c_i\}_{i=1}^{i=N}) - \epsilon_\theta(x_t | \emptyset)$, and extrapolate along this \emph{single} direction:
\begin{equation}
    \label{eqn:single_scale_cfg}
    \hat{\epsilon}_\theta(x_t | \{c_i\}_{i=1}^{i=N}) = \epsilon_\theta(x_t | \emptyset) + s\cdot \Delta^t_{\text{joint}}
\end{equation}
where $s \ge 1$ is the \emph{single} guidance scale. This formulation requires only two feed-forward executions at each denoising step, however, we can no longer control the magnitude of each direction separately.

We would like to stress that the training regime is identical for both of these formulations. Practically, it means that the user can train the model once, and only during inference decide which variant to choose, based on the preference at the moment. Through the rest of this paper, we used the fast variant with fixed $s=3$. See the ablation study in \Cref{sec:ablation_study} for a comparison between these variants.

In addition, we noticed that the texts in the image-text pairs dataset contain elaborate descriptions of the entire scene, whereas we aim to ease the use for the end-user and remove the need to provide an elaborate global prompt in addition to the local ones, i.e., to not require the user to repeat the same information twice. Hence, in order to reduce the domain gap between the training data and the input at inference time, we perform the following simple trick: we concatenate the local prompts to the global prompt at inference time separated by commas.

%% file: figures/spatio_textual_representation/fig.tex
\begin{figure*}[ht]
    \centering
    
    \includegraphics[width=2\columnwidth]{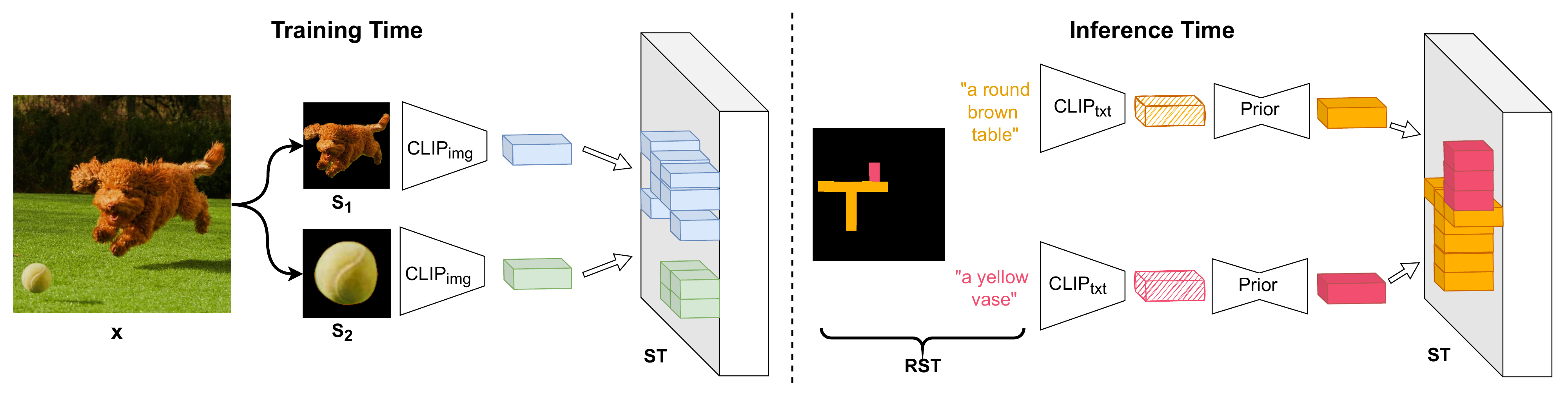}
    
    \caption{\textbf{Spatio-textual representation:} During training (left) --- given a training image $x$, we extract $K$ random segments, pre-process them and extract their CLIP image embeddings. Then we stack these embeddings in the same shapes of the segments to form the spatio-textual representation $ST$. During inference (right) --- we embed the local prompts into the CLIP text embedding space, then convert them using the prior model $P$ to the CLIP image embeddings space, lastly, we stack them in the same shapes of the inputs masks to form the spatio-textual representation $ST$.}
    \label{fig:spatio_textual_representation}
    \vspace{-1.5em}
\end{figure*}

%% file: figures/multiscale_control/fig.tex
\begin{figure*}[ht]
    \centering
    
    \centering
    \setlength{\tabcolsep}{1pt}
    \renewcommand{\arraystretch}{0.5}
    \setlength{\ww}{0.33\columnwidth}
  
    \begin{tabular}{cccccc}

        \scriptsize{``a sunny day at the beach''} &
        \scriptsize{(1)} &
        \scriptsize{(2)} &
        \scriptsize{(3)} &
        \scriptsize{(4)} &
        \scriptsize{(5)}
        \\

        \includegraphics[width=\ww,frame]{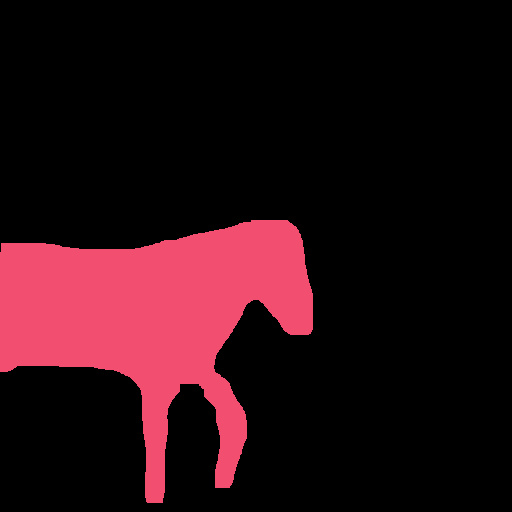}
        \phantom{a}
        &
        \includegraphics[width=\ww,frame]{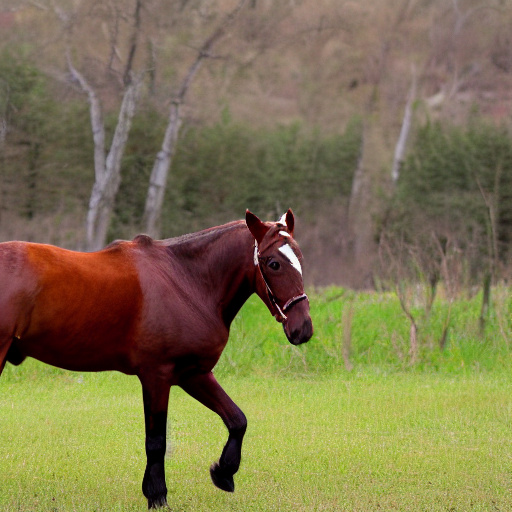} &
        \includegraphics[width=\ww,frame]{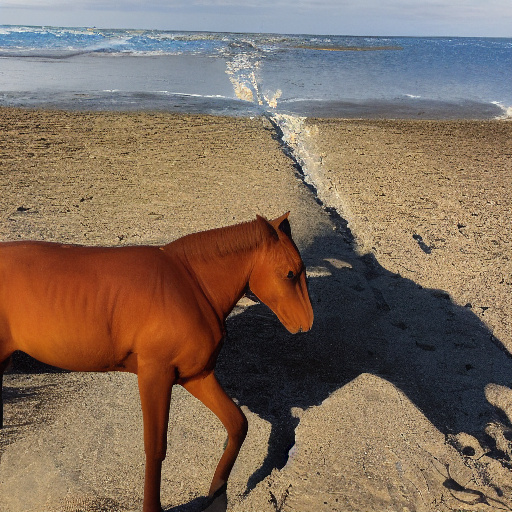} &
        \includegraphics[width=\ww,frame]{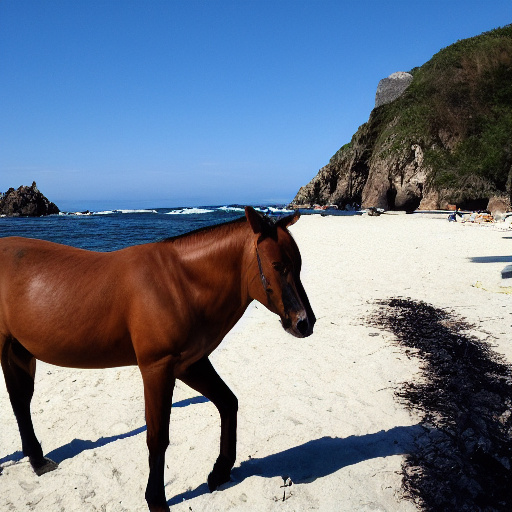} &
        \includegraphics[width=\ww,frame]{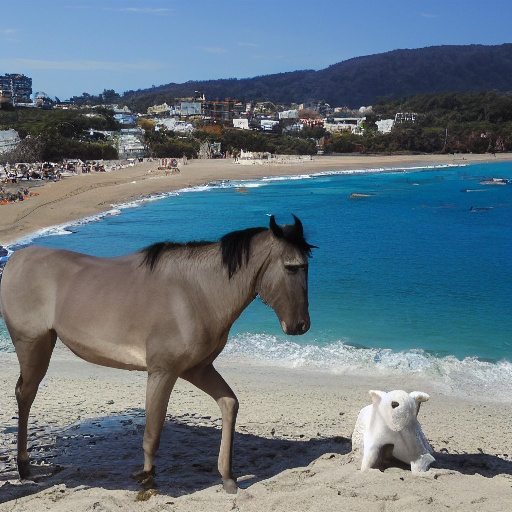} &
        \includegraphics[width=\ww,frame]{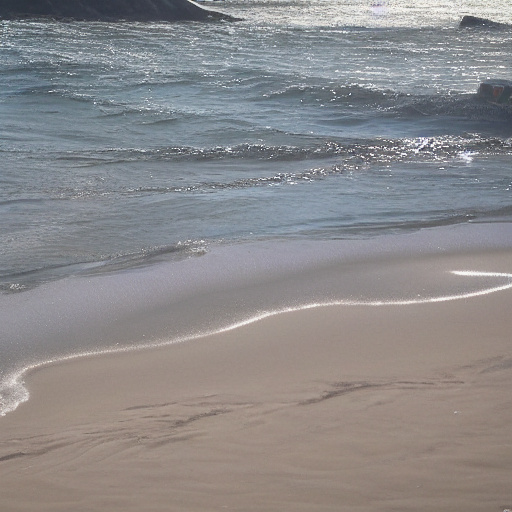}
        \\
        
        \maska{\scriptsize{``a brown horse''}} &
        \scriptsize{$s_{\text{global}} = 0 ; s_{\text{local}} = 3$} &
        \scriptsize{$s_{\text{global}} = 1.5 ; s_{\text{local}} = 3$} &
        \scriptsize{$s_{\text{global}} = 3 ; s_{\text{local}} = 3$} &
        \scriptsize{$s_{\text{global}} = 3 ; s_{\text{local}} = 1.5$} &
        \scriptsize{$s_{\text{global}} = 3 ; s_{\text{local}} = 0$}
        \\
    \end{tabular}
    
    \caption{\textbf{Multi-scale control:} Using the multi-scale inference (\Cref{eqn:multiscale_cfg}) allows fine-grained control over the input conditions. Given the same inputs (left), we can use different scales for each condition. In this example, if we put all the weight on the local scene (1), the generated image contains a horse with the correct color and posture, but not at the beach. Conversely, if we place all the weight on the global text (5), we get an image of a beach with no horse in it. The in-between results correspond to a mix of conditions --- in (4) we get a gray donkey, in (2) the beach contains no water, and in (3) we get a brown horse at the beach on a sunny day.}
    \vspace{-1.5em}
    \label{fig:multiscale_control}
\end{figure*}

%% file: sections/experiments.tex
\section{Experiments}
\label{sec:experimets}

For both the pixel-based and latent-based variants, we fine-tuned pre-trained text-to-image models with 35M image-text pairs, following Make-A-Scene \cite{gafni2022make}, while filtering out image-text pairs containing people.

In \Cref{sec:quantitative_and_qualitative_results} we compare our method against the baselines both qualitatively and quantitatively. Next, in \Cref{sec:user_study} we describe the user study we conducted. Then, in \Cref{sec:mask_sensitivity} we discuss the sensitivity of our method to details in the mask. Finally, in \Cref{sec:ablation_study} we report the ablation study results.

\subsection{Quantitative \& Qualitative Comparison}
\label{sec:quantitative_and_qualitative_results}

\input{tables/metrics_comparison.tex}
\input{figures/qualitative_comparison/fig.tex}

We compare our method against the following baselines: (1) No Token Left Behind (NTLB) \cite{paiss2022no} proposes a method that conditions a text-to-image model on spatial locations using an optimization approach. We adapt their method to our problem setting as follows: the global text prompt $\tglobal{}$ is converted to a full mask (that contains all the pixels in the image), and the raw spatio-textual matrix $RST$ is converted to separate masks. (2) Make-A-Scene (MAS) \cite{gafni2022make} proposes a method that conditions a text-to-image model on a global text $\tglobal{}$ and a \emph{dense} segmentation map with \emph{fixed labels}. We adapt MAS to support \emph{sparse} segmentation maps of \emph{general local prompts} by concatenating the local texts of the raw spatio-textual matrix $RST$ into the global text prompt $\tglobal$ as described in \Cref{sec:multi_conditional_cfg} and provide a label for each segment (if there is no exact label in the fixed list, the user should provide the closest one). Instead of a dense segmentation map, we provided a sparse segmentation map, where the background pixels are marked with an ``unassigned'' label.

In order to evaluate our method effectively, we need an automatic way to generate a large number of coherent inputs (global prompts $\tglobal$ as well as a raw spatio-textual matrix $RST$) for comparison. 
\Naive{}ly taking random inputs is undesirable, because such inputs will typically not represent a meaningful scene and may be impossible to generate. Instead, we propose to derive random inputs from \emph{real images}, thus guaranteeing that there is in fact a possible natural image corresponding to each input. We use 30,000 samples from COCO \cite{lin2014microsoft} validation set that contain global text captions as well as a dense segmentation map for each sample. We convert the segmentation map labels by simply providing the text ``a \{label\}'' for each segment. Then, we randomly choose a subset of those segments to form the sparse input. Notice that for MAS, we additionally provided the ground-truth label for each segment. For more details and generated input examples, see the supplementary document.

In order to evaluate the performance of our method numerically we propose to use the following metrics that test different aspects of the model: (1) \emph{FID score} \cite{heusel2017gans} to assess the overall quality of the results, (2) \emph{Global distance} to assess how well the model's results comply with the global text prompt $\tglobal{}$ --- we use CLIP to calculate the cosine distance between $\cliptxt{\tglobal{}}$ and $\clipimg{I}$, (3) \emph{Local distance} to assess the compliance between the result and the raw spatio-textual matrix $RST$ --- again, using CLIP for each of the segments in $RST$ separately, by cropping a tight area around each segment $c$, feeding it to $\clipimgd$ and calculating the cosine distance with $\cliptxt{\tlocal}$, (4) \emph{Local IOU} to assess the shape matching between the raw spatio-textual matrix $RST$ and the generated image --- for each segment in $RST$, we calculate the IOU between it and the segmentation prediction of a pre-trained segmentation model \cite{xie2021segformer}. As we can see in \Cref{tab:metrics_comparison}(left) our latent-based model outperforms the baselines in all the metrics, except the local IOU, which is better in MAS because our method is somewhat insensitive to the given mask shape (\Cref{sec:mask_sensitivity}) --- we view this as an advantage. In addition, we can see that our latent-based variant outperforms the pixel-based variant in all the metrics, which may be caused by insufficient re-implementation of the \DALLE~2 model. Nevertheless, we noticed that this pixel-based model is also able to take into account both the global text and spatio-textual representation. The rightmost column in \Cref{tab:metrics_comparison} reports the inference times for a single image across the different models computed on a single V100 NVIDIA GPU. The results indicate that our method (especially the latent-based one) outperforms the baselines significantly. For more details, please read the supplementary.

In addition, \Cref{fig:qualitative_comparison} shows a qualitative comparison between the two variants of our method and the baselines. For MAS, we manually choose the closest label from the fixed labels set. As we can see, the \name (latent) outperforms the baselines in terms of compliance with both the global and local prompts, and in overall image quality.

\subsection{User Study}
\label{sec:user_study}
In addition, we conducted a user study on Amazon Mechanical Turk (AMT)~\cite{amt_website} to assess the visual quality, as well as compliance with global and local prompts. For each task, the raters were asked to choose between two images generated by different models along the following dimensions: (1) overall image quality, (2) text-matching to the global prompt $\tglobal$ and (3) text-matching to the local prompts of the raw spatio-textual matrix $RST$. For more details, please read the supplementary.

In \Cref{tab:metrics_comparison} (middle) we present the evaluation results against the baselines, as the percentage of majority rates that preferred our method (based on the latent model) over the baseline. As we can see, our method is preferred by human evaluators in all these aspects vs. all the baselines. In addition, NTLB \cite{paiss2022no} achieved overoptimistic scores in the CLIP-based automatic metrics --- it achieved lower human evaluation ratings than Make-A-Scene \cite{gafni2022make} in the global and local text-matching aspects, even though it got better scores in the corresponding automatic metrics. This might be because NTLB is an optimization-based solution that uses CLIP for generation, hence is susceptible to adversarial attacks \cite{szegedy2013intriguing, goodfellow2014explaining}.

\subsection{Mask Sensitivity}
\label{sec:mask_sensitivity}
\input{figures/mask_sensitivity/fig.tex}
During our experiments, we noticed that the model generates images that correspond to the implicit masks in the spatio-textual representation $\mathit{ST}$, but not \emph{perfectly}. This is also evident in the local IOU scores in \Cref{tab:metrics_comparison}. Nevertheless, we argue that this characteristic can be beneficial, especially when the input mask is not realistic. As demonstrated in \Cref{fig:mask_sensitivity}, given an inaccurate, hand drawn, general animal shape (left) the model is able to generate different objects guided by the local text prompt, even when it does not perfectly match the mask. For example, the model is able to add ears (in the cat and dog examples) and horns (in the goat example), which are not presented in the input mask, or to change the body type (as in the tortoise example). However, all the results share the same pose as the input mask. One reason for this behavior might be the downsampling of the input mask, so during training some fine-grained details are lost, hence the model is incentivized to fill in the missing gaps according to the prompts.

\subsection{Ablation Study}
\label{sec:ablation_study}

\input{tables/ablation_study.tex}

We conducted an ablation study for the following cases: (1) \emph{Binary representation} --- in \Cref{sec:spatio_textual_representation} we used the CLIP model for the spatio-textual representation $ST$. Alternatively, we could use a simpler binary representation that converts the raw spatio-textual matrix $RST$ into a binary mask $\mathit{B}$ of shape $(H, W)$ by:
\begin{equation}
    B[j, k] =
    \begin{cases}
        1 & \text{if $RST[j, k] \ne \emptyset$} \\
        0 & \text{otherwise} \\
    \end{cases}
\end{equation}
and concatenate the local text prompts into the global prompt.
(2) \emph{CLIP text embedding} --- as described in \Cref{sec:sota_models_adaptaion}, we mitigate the domain gap between $\clipimgd$ and $\cliptxtd$ we employing a prior model $P$. Alternatively, we could use the $\cliptxtd$ directly by removing the prior model. (3) \emph{Multi-scale inference} --- as described in \Cref{sec:multi_conditional_cfg} we used the single scale variant (\Cref{eqn:single_scale_cfg}). Alternatively, we could use the multi-scale variant (\Cref{eqn:multiscale_cfg}).

As can be seen in \Cref{tab:ablation_study} our method outperforms the ablated cases in terms of FID score, human visual quality and human global text-match. When compared to the simple representation (1) our method is able to achieve better local text-match determined by the user study but smaller local IOU, one possible reason is that it is easier for the model to fit the shape of a simple mask (as in the binary case), but associating the relevant portion of the global text prompt to the corresponding segment is harder. When compared to the version with CLIP text embedding (2) our model achieves slightly less local IOU score and human local text-match while achieving better FID and overall visual quality. Lastly, the single scale manages to achieve better results than the multi-scale one (3) while only slightly less in the local CLIP distance.

%% file: tables/metrics_comparison.tex
\begin{table}[t]
    \begin{center}
        \setlength{\tabcolsep}{2pt}
        \begin{adjustbox}{width=1.0\columnwidth}
            \begin{tabular}{lcccc|ccc|c}
                
                &
                \multicolumn{4}{c}{Automatic Metrics} &
                \multicolumn{3}{c}{User Study} &
                Performance
                \\

                \toprule
     
                \textbf{Method} & 
                Global $\downarrow$ & 
                Local $\downarrow$ &
                Local $\uparrow$ &
                FID  $\downarrow$&
                Visual &
                Global &
                Local &
                Inference $\downarrow$
                \\

                &
                distance & 
                distance & 
                IOU &
                & 
                quality & 
                match  &
                match &
                time (sec)
                \\
                
                \midrule
                NTLB \cite{paiss2022no} &
                0.7547 &
                0.7814 &
                0.1914 &
                36.004 &
                91.4\% &
                85.54\% &
                79.29\% &
                326
                \\

                MAS \cite{gafni2022make} &
                0.7591 &
                0.7835 &
                \textbf{0.2984} &
                21.367 &
                81.25\% &
                70.61\% &
                57.81\% &
                76
                \\

                \midrule
                \name (pixel) &
                0.7661 &
                0.7862 &
                0.2029 &
                23.128 &
                87.11\% &
                80.96\% &
                71.09\% &
                52
                \\

                \name (latent) &
                \textbf{0.7436} &
                \textbf{0.7795} &
                0.2842 &
                \textbf{6.7721} &
                - &
                - &
                - &
                \textbf{7}
                \\

                \bottomrule
            \end{tabular}
        \end{adjustbox}
    \caption{\textbf{Metrics comparison:} We evaluated our method against the baselines using automatic metrics (left) and human ratings (middle). The results for the human ratings (middle) are reported as the percentage of the majority vote raters that preferred our latent-based variant of our method over the baseline, i.e., any value above 50\% means our method was favored. The inference time reported (right) are for a single image.}
    \vspace{-2em}
    \label{tab:metrics_comparison}
    \end{center}
\end{table}
    

%% file: figures/qualitative_comparison/fig.tex
\begin{figure*}[ht]
    \centering
    
    \centering
    \setlength{\tabcolsep}{1pt}
    \renewcommand{\arraystretch}{0.5}
    \setlength{\ww}{0.282\columnwidth}
  
    \begin{tabular}{cccccccc}
        
        &&&
        \scriptsize{``a sunny day after} &
        &
        \scriptsize{``a bathroom with} &
        \\

        &
        \scriptsize{``near a lake''} &
        \scriptsize{``a painting''} &
        \scriptsize{the snow''} &
        \scriptsize{``on a table''} &
        \scriptsize{an artificial light''} &
        \scriptsize{``near a river''} &
        \scriptsize{``at the desert''}
        \\

        \rotatebox{90}{\scriptsize\phantom{AAAAa} Inputs} &
        \includegraphics[width=\ww,frame]{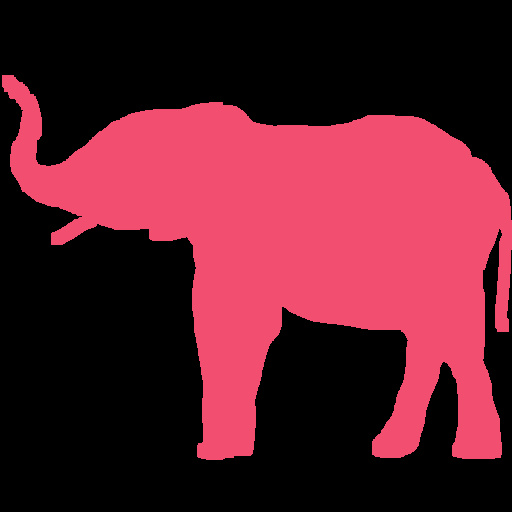} &
        \includegraphics[width=\ww,frame]{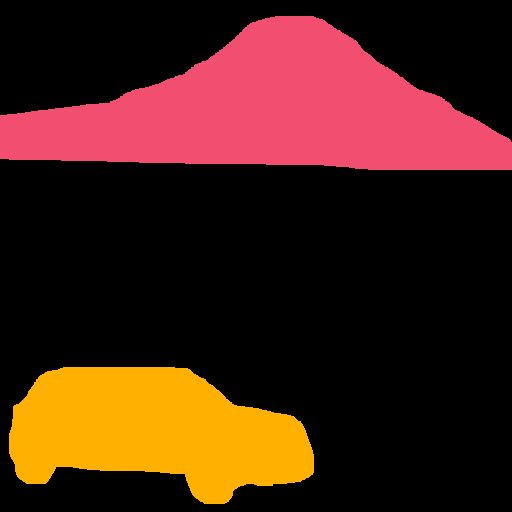} &
        \includegraphics[width=\ww,frame]{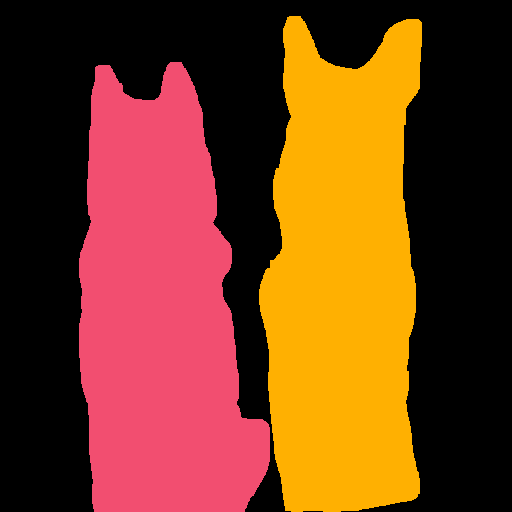} &
        \includegraphics[width=\ww,frame]{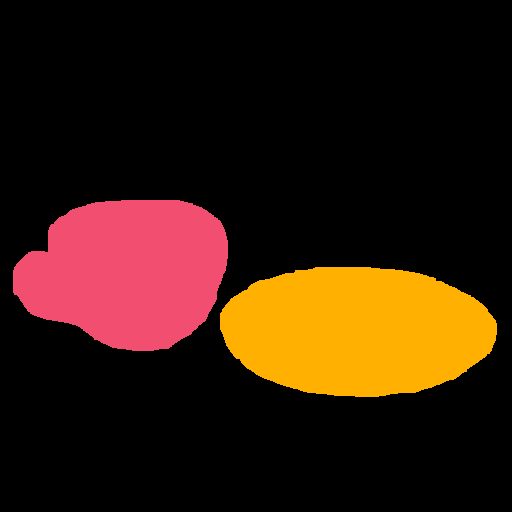} &
        \includegraphics[width=\ww,frame]{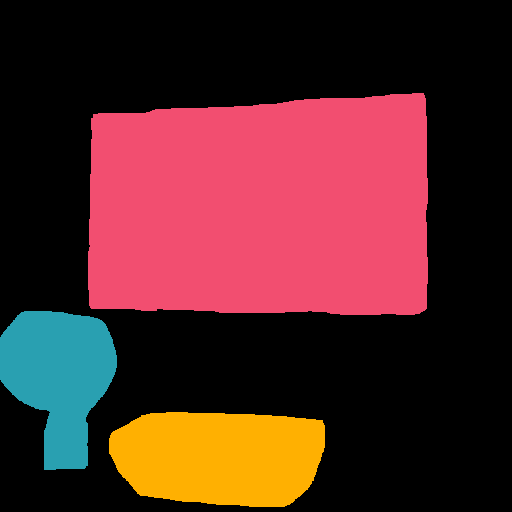} &
        \includegraphics[width=\ww,frame]{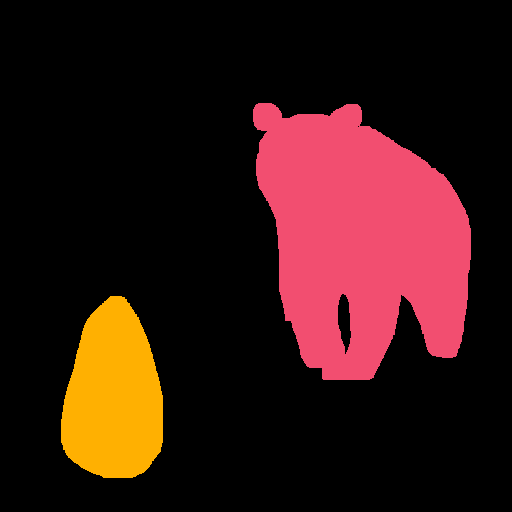} &
        \includegraphics[width=\ww,frame]{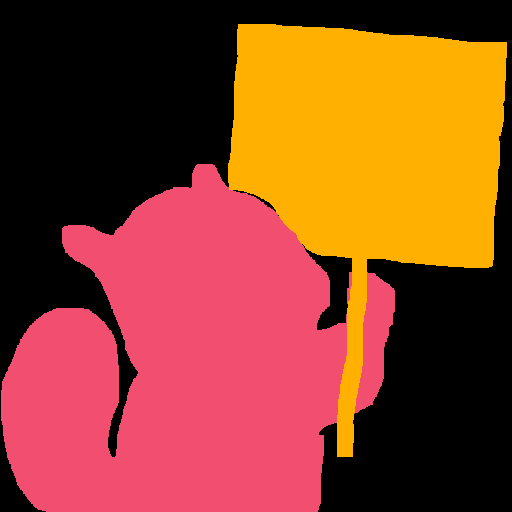}
        \\
        
        &
        \begin{tabular}{c}
            \maska{\scriptsize{``a black elephant''}} \\ 
            \\
            \\
            \\
            \\
        \end{tabular} &

        \begin{tabular}{c}
            \maska{\scriptsize{``a snowy mountain''}} \\ 
            \maskb{\scriptsize{``a red car''}} \\
            \\
            \\
            \\
        \end{tabular} &

        \begin{tabular}{c}
            \maska{\scriptsize{``a Husky dog''}} \\
            \maskb{\scriptsize{``a German }} \\
            \maskb{\scriptsize{Shepherd dog''}} \\
            \\
            \\
        \end{tabular} &

        \begin{tabular}{c}
            \maska{\scriptsize{``a mug''}} \\
            \maskb{\scriptsize{``a white plate }} \\
            \maskb{\scriptsize{with cookies''}} \\
            \\
            \\
        \end{tabular} &

        \begin{tabular}{c}
            \maska{\scriptsize{``a mirror''}} \\
            \maskb{\scriptsize{``a white sink''}} \\
            \maskc{\scriptsize{``a vase with}} \\
            \maskc{\scriptsize{red flowers''}} \\
            \\
        \end{tabular} &

        \begin{tabular}{c}
            \maska{\scriptsize{``a grizzly bear''}} \\
            \maskb{\scriptsize{``a huge avocado''}} \\
            \\
            \\
            \\
        \end{tabular} &

        \begin{tabular}{c}
            \maska{\scriptsize{``a squirrel''}} \\
            \maskb{\scriptsize{``a sign with an }} \\
            \maskb{\scriptsize{apple painting''}} \\
            \\
            \\
        \end{tabular}

        \\

        \rotatebox{90}{\scriptsize\phantom{AAAa} NTLB \cite{paiss2022no}} &
        \includegraphics[width=\ww,frame]{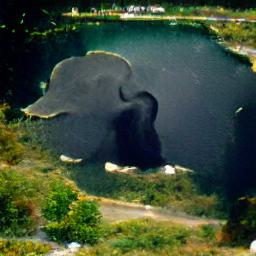} &
        \includegraphics[width=\ww,frame]{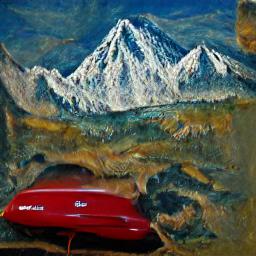} &
        \includegraphics[width=\ww,frame]{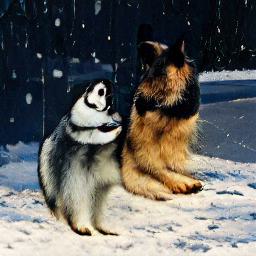} &
        \includegraphics[width=\ww,frame]{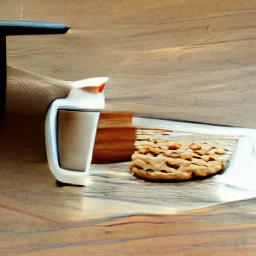} &
        \includegraphics[width=\ww,frame]{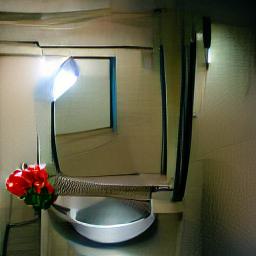} &
        \includegraphics[width=\ww,frame]{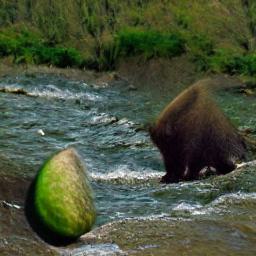} &
        \includegraphics[width=\ww,frame]{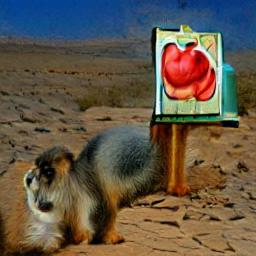}
        \\

        \rotatebox{90}{\scriptsize\phantom{AAA} MAS \cite{gafni2022make}} &
        \includegraphics[width=\ww,frame]{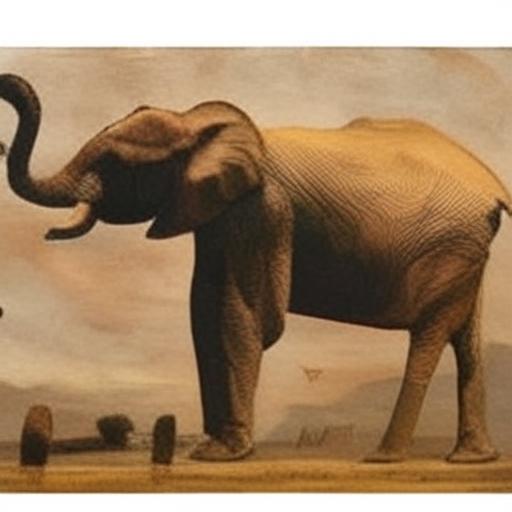} &
        \includegraphics[width=\ww,frame]{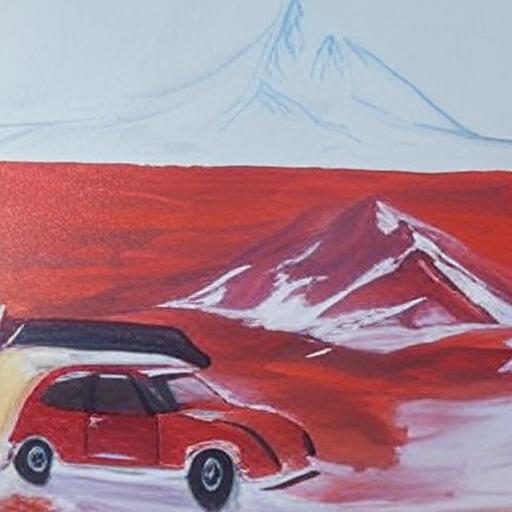} &
        \includegraphics[width=\ww,frame]{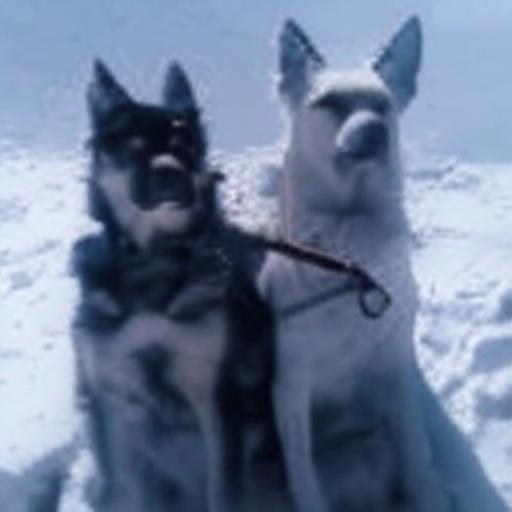} &
        \includegraphics[width=\ww,frame]{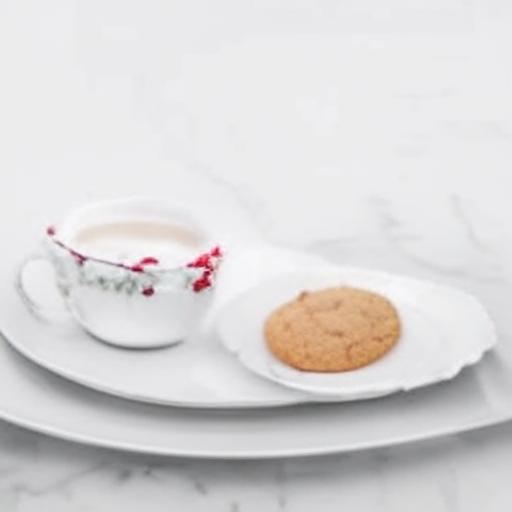} &
        \includegraphics[width=\ww,frame]{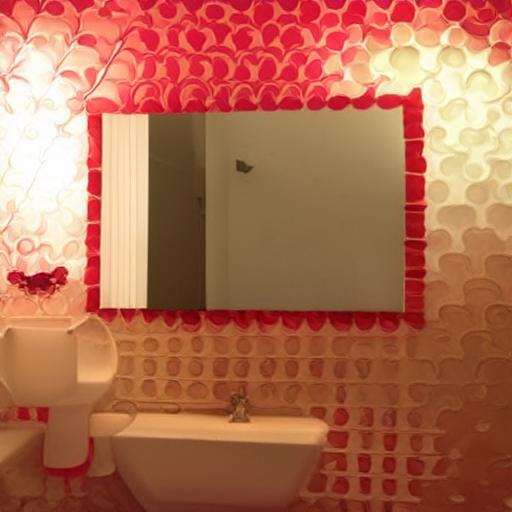} &
        \includegraphics[width=\ww,frame]{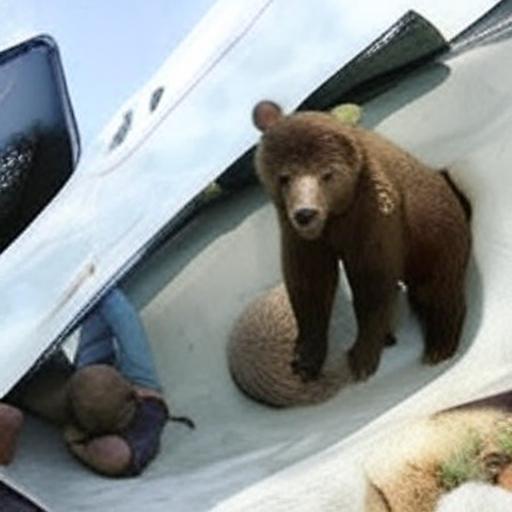} &
        \includegraphics[width=\ww,frame]{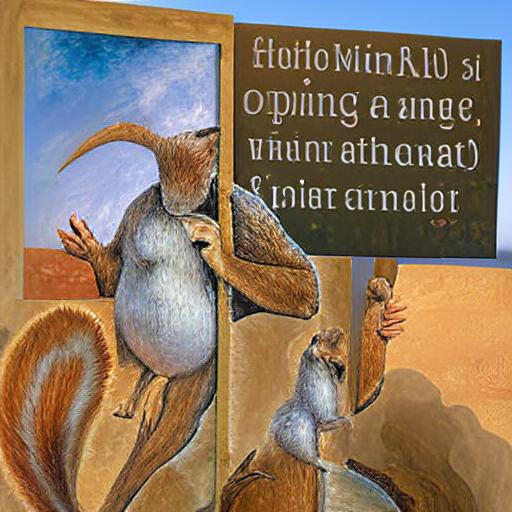}
        \\

        \rotatebox{90}{\scriptsize\phantom{AA} \name (pixel)} &
        \includegraphics[width=\ww,frame]{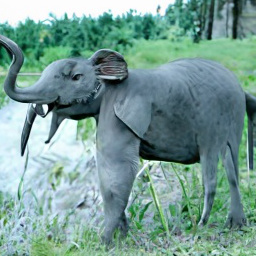} &
        \includegraphics[width=\ww,frame]{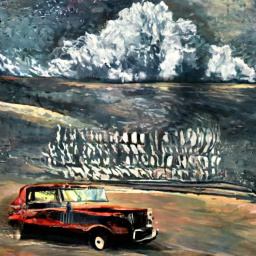} &
        \includegraphics[width=\ww,frame]{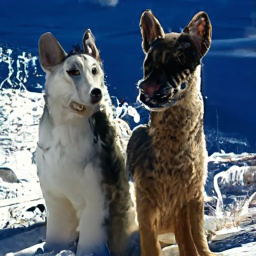} &
        \includegraphics[width=\ww,frame]{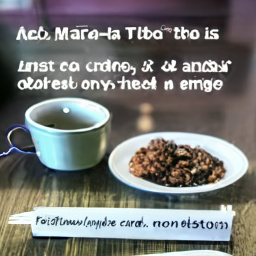} &
        \includegraphics[width=\ww,frame]{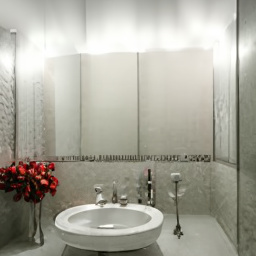} &
        \includegraphics[width=\ww,frame]{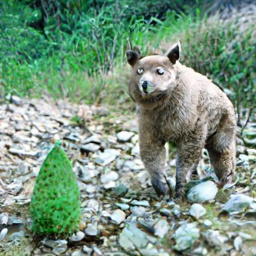} &
        \includegraphics[width=\ww,frame]{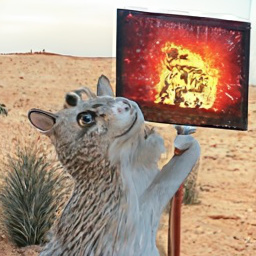}
        \\

        \rotatebox{90}{\scriptsize\phantom{AA} \name (pixel)} &
        \includegraphics[width=\ww,frame]{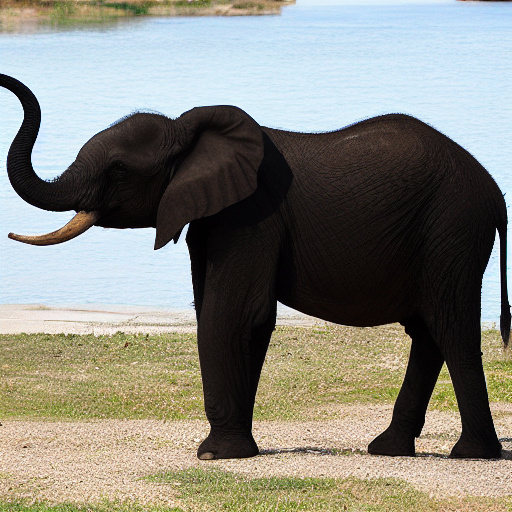} &
        \includegraphics[width=\ww,frame]{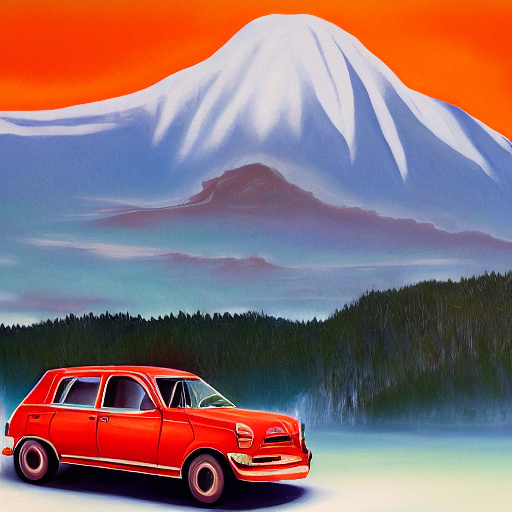} &
        \includegraphics[width=\ww,frame]{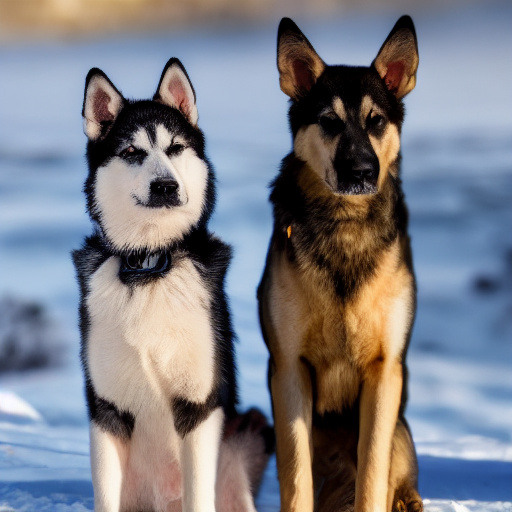} &
        \includegraphics[width=\ww,frame]{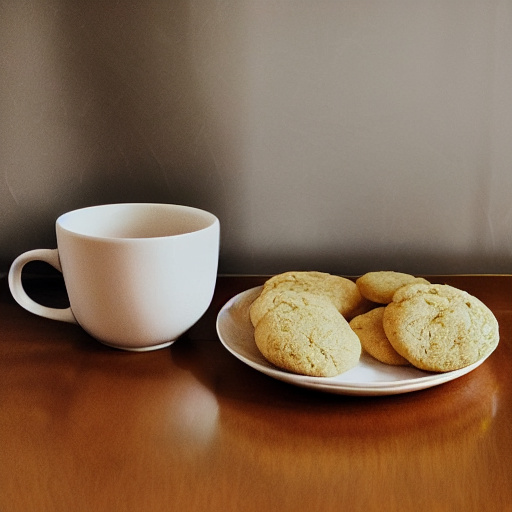} &
        \includegraphics[width=\ww,frame]{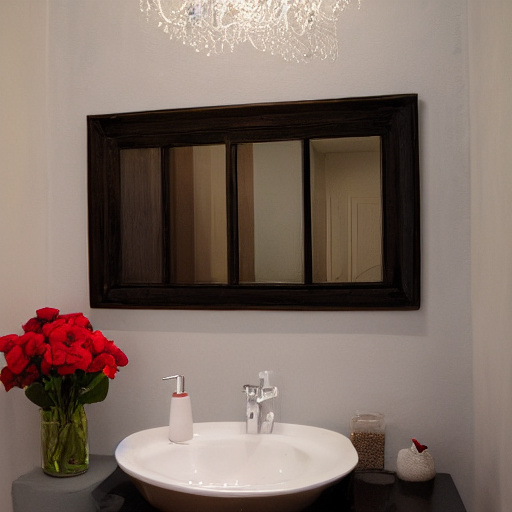} &
        \includegraphics[width=\ww,frame]{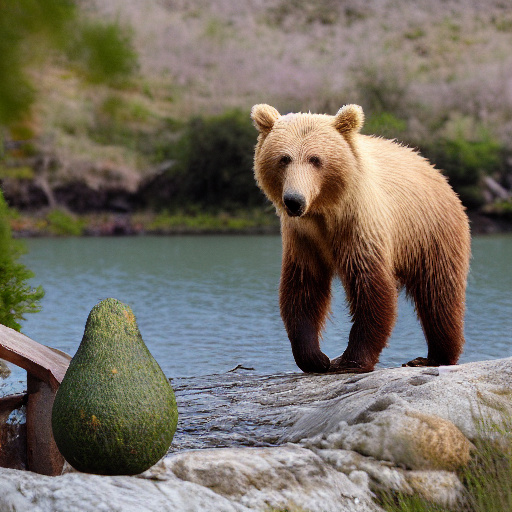} &
        \includegraphics[width=\ww,frame]{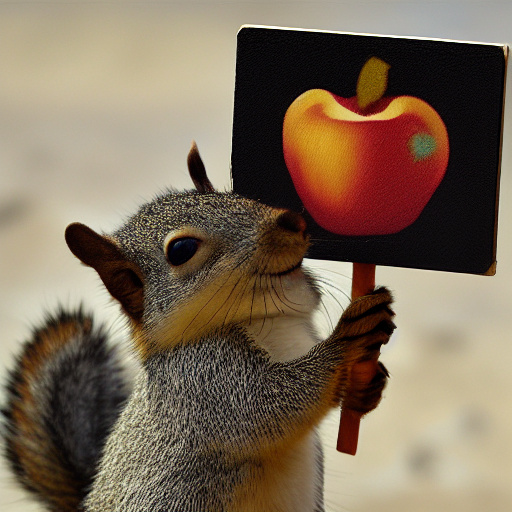}
        \\
    \end{tabular}
    
    \caption{\textbf{Qualitative comparison:} Given the inputs (top row), we generate images using the baselines (adapted to our task as described in \Cref{sec:quantitative_and_qualitative_results}) and the two variants of our method. As we can see, \name (latent) outperforms the baselines in terms of compliance with both the global and local texts, and in overall image quality.}
    \vspace{-0.25em}
    \label{fig:qualitative_comparison}
\end{figure*}

%% file: figures/mask_sensitivity/fig.tex
\begin{figure*}[ht]
    \centering
    
    \centering
    \setlength{\tabcolsep}{1pt}
    \renewcommand{\arraystretch}{0.5}
    \setlength{\ww}{0.28\columnwidth}
  
    \begin{tabular}{ccccccc}

        \scriptsize{``on a pile of dry leaves''} &
        \scriptsize{(1)} &
        \scriptsize{(2)} &
        \scriptsize{(3)} &
        \scriptsize{(4)} &
        \scriptsize{(5)} &
        \scriptsize{(6)}
        \\

        \includegraphics[width=\ww,frame]{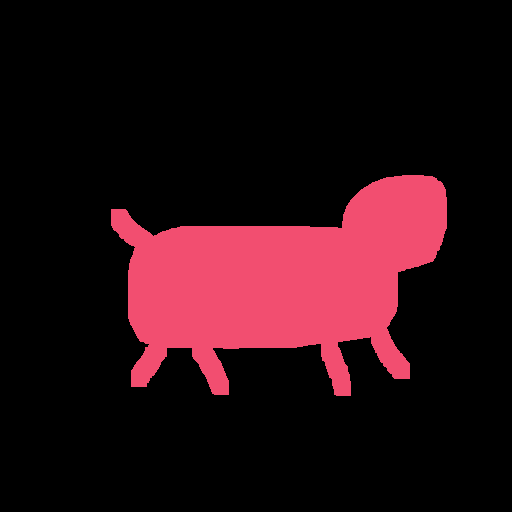}
        \phantom{.}
        &
        \includegraphics[width=\ww,frame]{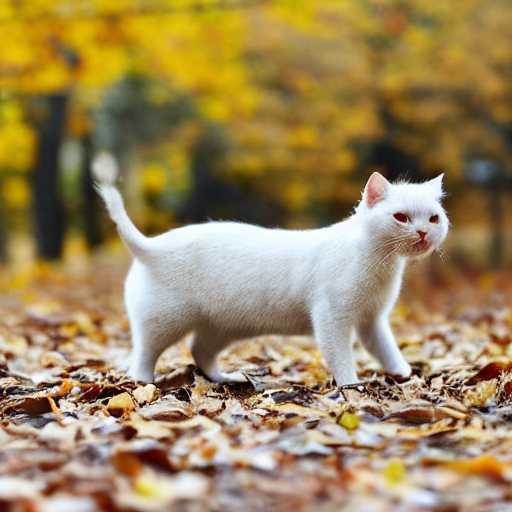} &
        \includegraphics[width=\ww,frame]{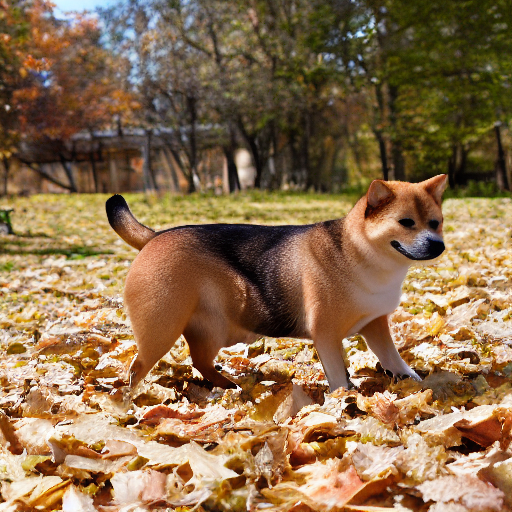} &
        \includegraphics[width=\ww,frame]{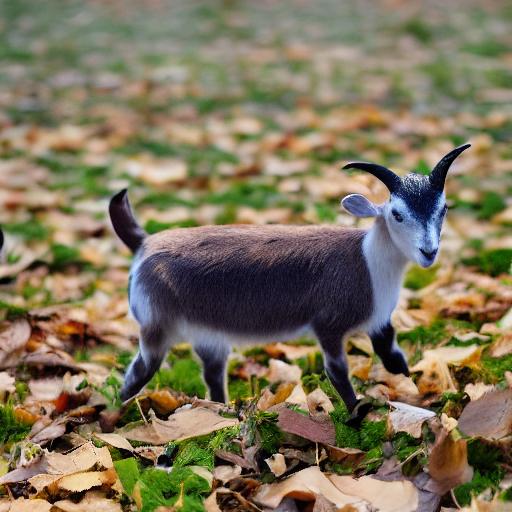} &
        \includegraphics[width=\ww,frame]{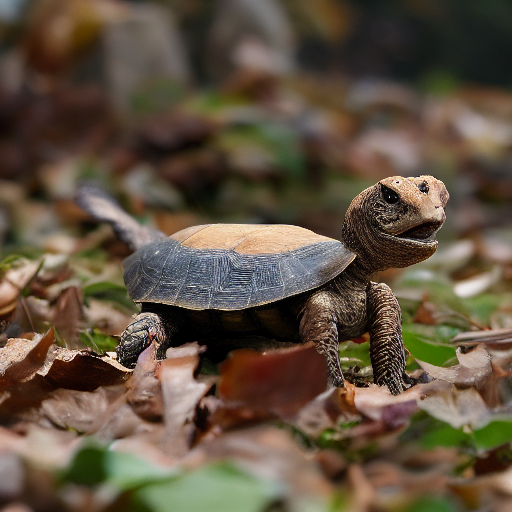} &
        \includegraphics[width=\ww,frame]{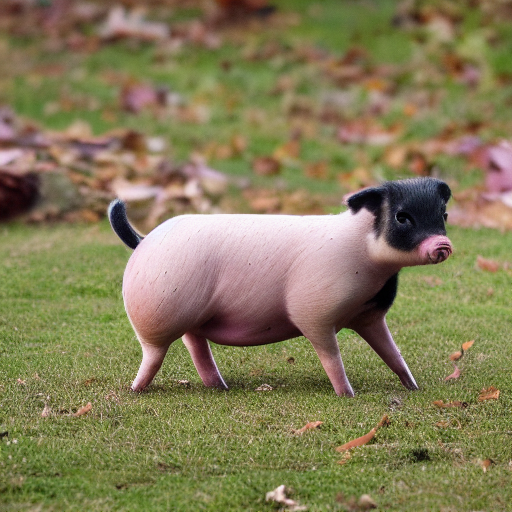} &
        \includegraphics[width=\ww,frame]{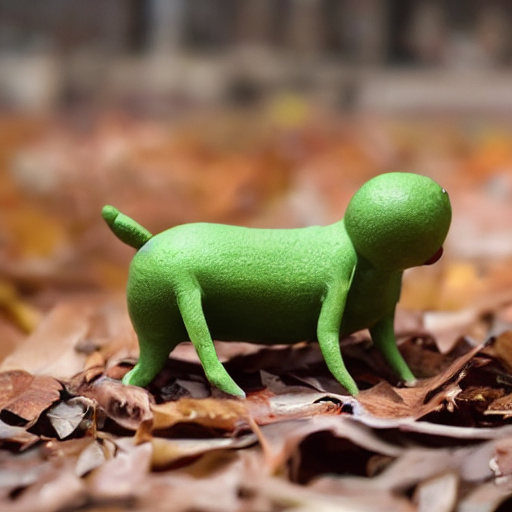}
        \\
        
        &
        \maska{\scriptsize{``a white cat''}} &
        \maska{\scriptsize{``a Shiba Inu dog''}} &
        \maska{\scriptsize{``a goat''}} &
        \maska{\scriptsize{``a tortoise''}} &
        \maska{\scriptsize{``a pig''}} &
        \maska{\scriptsize{``an avocado''}}
        \\
    \end{tabular}
    
    \caption{\textbf{Mask insensitivity:} We found that the model is relatively insensitive to errors in the input mask. Given a general animal shape mask (left), the model is able to generate a diverse set of results driven by the different local prompts. It can add ears/horns, as in the cat, dog and goat examples or change the body type, as in the tortoise example. However, all the results share the same pose as the input mask.}
    \vspace{-1em}
    \label{fig:mask_sensitivity}
\end{figure*}

%% file: tables/ablation_study.tex
\begin{table}[t]
    \begin{center}
        \setlength{\tabcolsep}{2pt}
        \begin{adjustbox}{width=1.0\columnwidth}
            \begin{tabular}{lcccc|cccc}
                
                &
                \multicolumn{4}{c}{Automatic Metrics} &
                \multicolumn{3}{c}{User Study}
                \\

                \toprule
     
                \textbf{Scenario} & 
                Global $\downarrow$ & 
                Local $\downarrow$ &
                Local $\uparrow$ &
                FID  $\downarrow$&
                Visual &
                Global &
                Local
                \\

                &
                distance & 
                distance & 
                IOU &
                & 
                quality & 
                text-match  &
                text-match
                \\
                
                \midrule
                (1) Binary &
                0.7457 &
                0.7797 &
                0.2973 &
                7.6085 &
                53.13\% &
                50.3\% &
                54.98\%
                \\

                (2) $\cliptxtd$ &
                0.7447 &
                0.7795 &
                \textbf{0.3092} &
                7.025 &
                58.6\% &
                56.74\% &
                48.53\%
                \\

                (3) Multiscale &
                0.7566 &
                \textbf{0.7794} &
                0.2767 &
                10.5896 &
                53.61\% &
                58.59\% &
                55.57\%
                \\

                \midrule
                \name (latent) &
                \textbf{0.7436} &
                0.7795 &
                0.2842 &
                \textbf{6.7721} &
                - &
                - &
                -
                \\

                \bottomrule
            \end{tabular}
        \end{adjustbox}
    \caption{\textbf{Ablation study:} The baseline method that we used in this paper achieves better FID score and visual quality than the ablated cases. It is outperformed in terms of local IOU in (1) and (2), and in terms of local text-match in (2). The results for the human ratings (right) are reported as the percentage of the majority vote raters that preferred our \name (latent).}
    \vspace{-2em}
    \label{tab:ablation_study}
    \end{center}
\end{table}

%% file: sections/limitations.tex
\section{Limitations and Conclusions}
\label{sec:limitations}

\input{figures/limitations/fig.tex}
We found that in some cases, especially when there are more than a few segments, the model might miss some of the segments or propagate their characteristics. For example, instead of generating a blue bowl in \Cref{fig:limitations}(left), the model generates a beige vase, matching the appearance of the table. Improving the accuracy of the model in these cases is an appealing research direction.

In addition, the model struggles to handle tiny segments. For example, as demonstrated in \Cref{fig:limitations}(right), the model ignores the golden coin masks altogether. This might be caused by our fine-tuning procedure: when we fine-tune the model, we choose a random number of segments that are above a size threshold because CLIP embeddings are meaningless for very low-resolution images. For additional examples, please read the supplementary. 

In conclusion, in this paper we addressed the scenario of text-to-image generation with sparse scene control. 
We believe that our method has the potential to accelerate the democratization of content creation by enabling greater control over the content generation process, supporting professional artists and novices alike.

%% file: figures/limitations/fig.tex
\begin{figure}[th]
    \centering
    
    \centering
    \setlength{\tabcolsep}{1pt}
    \renewcommand{\arraystretch}{0.5}
    \setlength{\ww}{0.22\columnwidth}
  
    \begin{tabular}{cccc}

        \scriptsize{``a room with} &&
        \\
        \scriptsize{sunlight''} &&
        \scriptsize{``on the grass''}
        \\

        \includegraphics[width=\ww,frame]{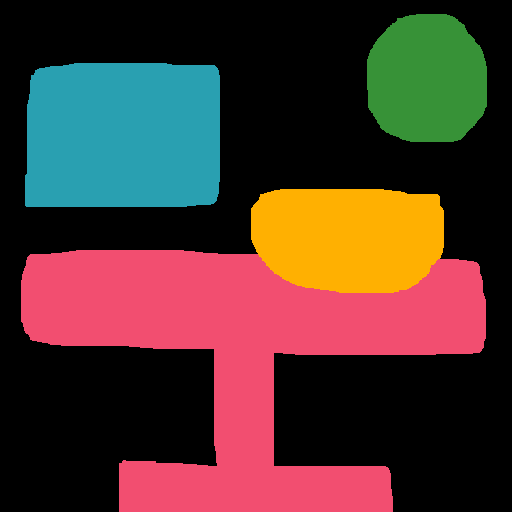} &
        \includegraphics[width=\ww,frame]{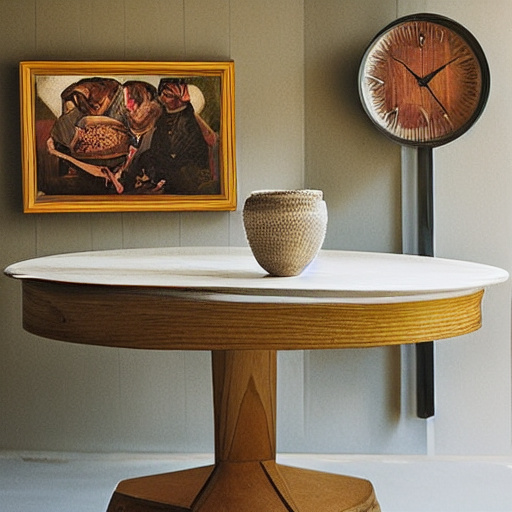} 
        \phantom{.}
        &
        \includegraphics[width=\ww,frame]{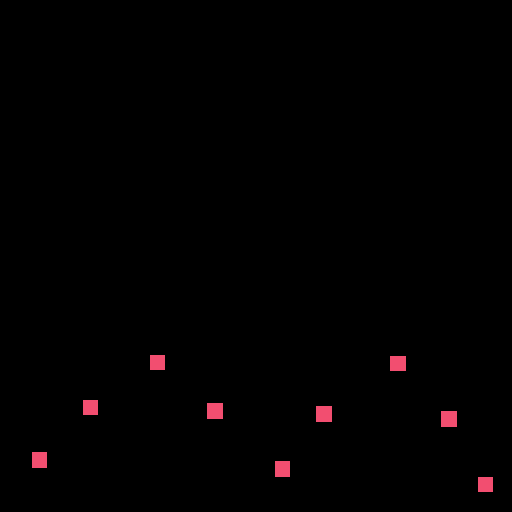} &
        \includegraphics[width=\ww,frame]{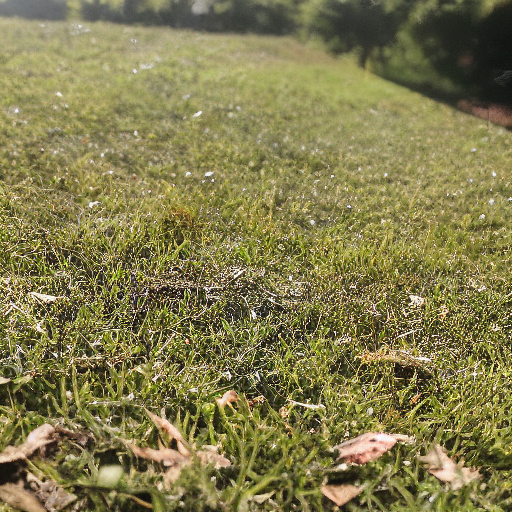}
        \\
        
        \begin{tabular}{c}
            \maska{\scriptsize{``a wooden table''}} \\ 
            \maskb{\scriptsize{``a blue bowl''}} \\
            \maskc{\scriptsize{``a picture on}} \\
            \maskc{\scriptsize{the wall''}} \\
            \maskd{\scriptsize{``a clock''}} \\
        \end{tabular} &&

        \begin{tabular}{c}
            \maska{\scriptsize{``golden coins''}} \\ 
            \\
            \\
            \\
            \\
        \end{tabular}
        \\
    \end{tabular}
    
    \vspace{-0.5em}
    \caption{\textbf{Limitations:} In some cases, characteristics propagate to adjacent segments, e.g. (left), instead of a blue bowl the model generated a vase with a wooden color. In addition, the model tends to ignore tiny masks (right).}
    \vspace{-0.5em}
    \label{fig:limitations}
\end{figure}

%% file: sections/appendix/additional_examples.tex
\section{Additional Examples}
\label{sec:additional_examples}

\input{figures/method_examples/additional1.tex}
\input{figures/method_examples/additional2.tex}
\input{figures/method_examples/additional3.tex}
\input{figures/method_examples/additional4.tex}
\input{figures/method_examples/additional5.tex}

\input{figures/additional_mask_sensitivity1/fig.tex}
\input{figures/additional_mask_sensitivity2/fig.tex}

\input{figures/additional_multiscale_control1/fig.tex}
\input{figures/additional_multiscale_control2/fig.tex}

\input{figures/additional_limitations/fig.tex}

In Figures \ref{fig:method_additional_examples1}, \ref{fig:method_additional_examples2}, \ref{fig:method_additional_examples3}, \ref{fig:method_additional_examples4} and \ref{fig:method_additional_examples5} we provide additional results from our model. In \Cref{fig:additional_mask_sensitivity1,fig:additional_mask_sensitivity2} we provide additional examples for the mask insensitivity of our method. In \Cref{fig:additional_multiscale_control1,fig:additional_multiscale_control2} we show the fine-grained control that is achievable via the multi-scale version of our method. In \Cref{fig:additonal_limitations} we provide additional limitations of our method.

%% file: figures/method_examples/additional1.tex
\begin{figure*}[ht]
    \centering
    \setlength{\tabcolsep}{1pt}
    \renewcommand{\arraystretch}{0.5}
    \setlength{\ww}{0.49\columnwidth}
    \begin{tabular}{cccc}

        ``an oil painting'' &&
        ``in the snow''
        \\

        \includegraphics[width=\ww,frame]{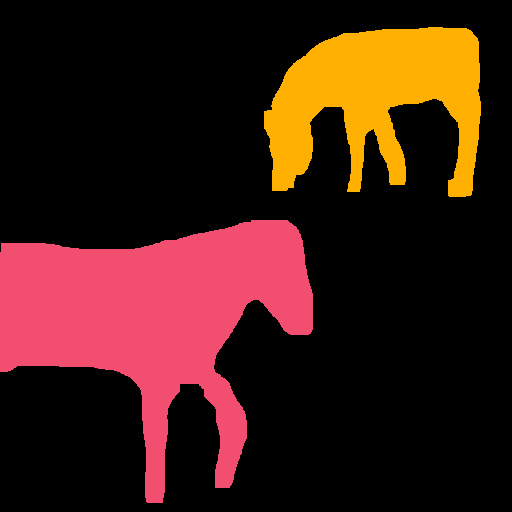} & 
        \includegraphics[width=\ww,frame]{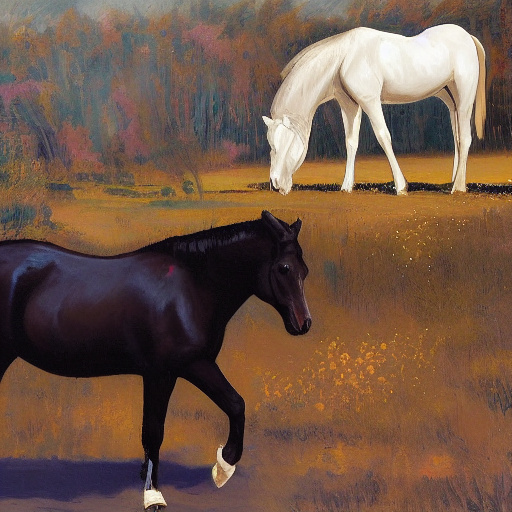} 
        \phantom{a}
        &
        
        \includegraphics[width=\ww,frame]{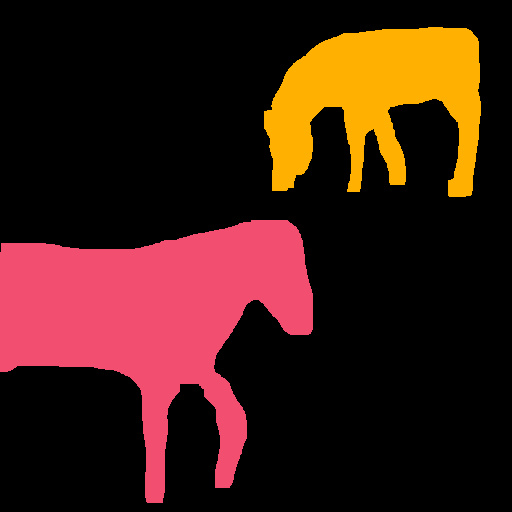} & 
        \includegraphics[width=\ww,frame]{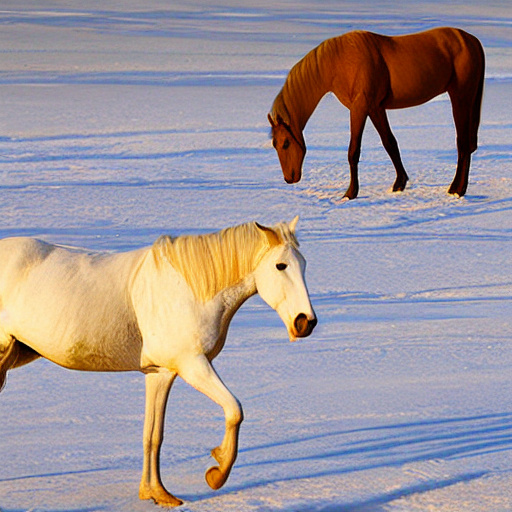}
        \phantom{a}
        \\

        \begin{tabular}{c}
            \maska{``a black horse''} \\ 
            \maskb{``a white horse''} \\
            \\
            \\
        \end{tabular} &&

        \begin{tabular}{c}
            \maska{``a white horse''} \\ 
            \maskb{``a brown horse''} \\
            \\
            \\
        \end{tabular}
        \\

        ``a photo taken during &&
        \\
        the golden hour'' &&
        ``a sunny day at the beach''
        \\

        \includegraphics[width=\ww,frame]{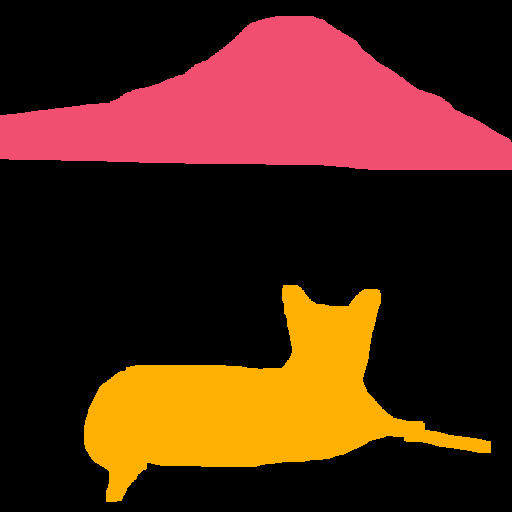} & 
        \includegraphics[width=\ww,frame]{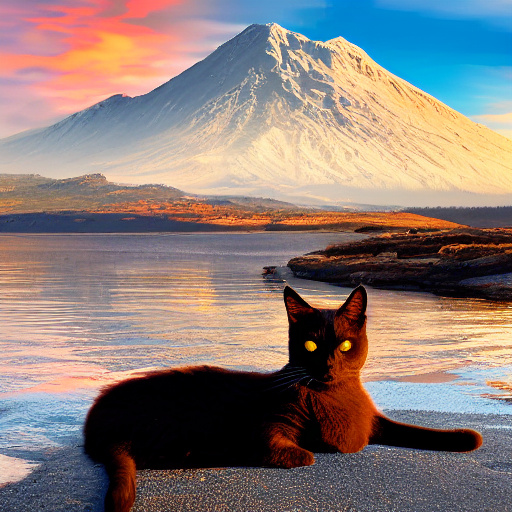} 
        \phantom{a}
        &
        
        \includegraphics[width=\ww,frame]{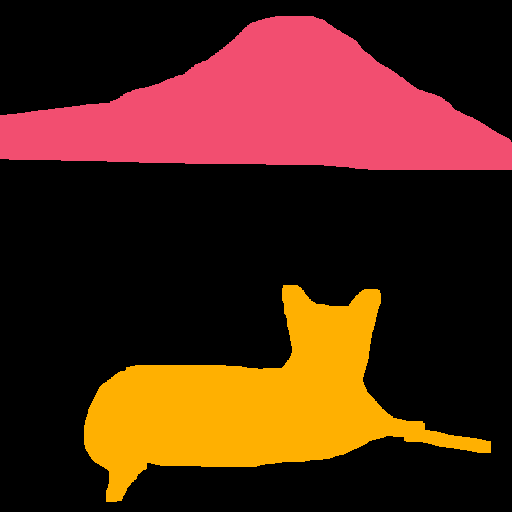} & 
        \includegraphics[width=\ww,frame]{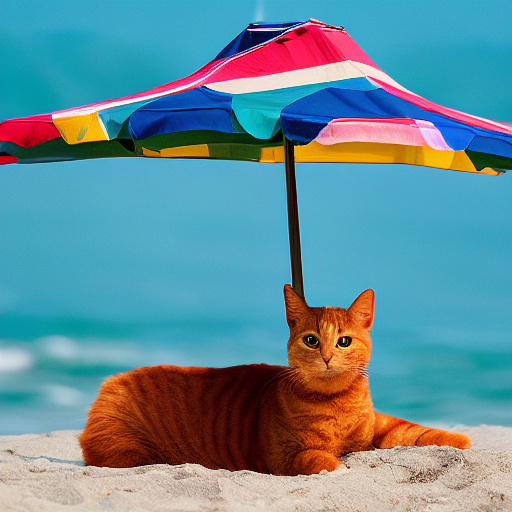}
        \phantom{a}
        \\

        \begin{tabular}{c}
            \maska{``a snowy mountain''} \\ 
            \maskb{``a black cat''} \\
            \\
            \\
        \end{tabular} &&

        \begin{tabular}{c}
            \maska{``a colorful beach umbrella''} \\
            \maskb{``a ginger cat''} \\
            \\
            \\
        \end{tabular}
        \\

        ``near some ruins'' &&
        ``at the desert''
        \\

        \includegraphics[width=\ww,frame]{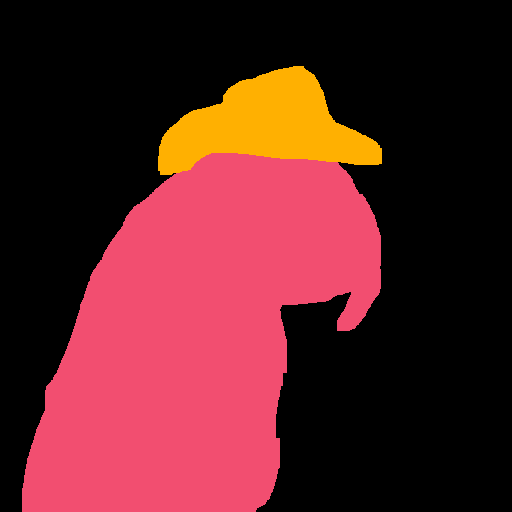} & 
        \includegraphics[width=\ww,frame]{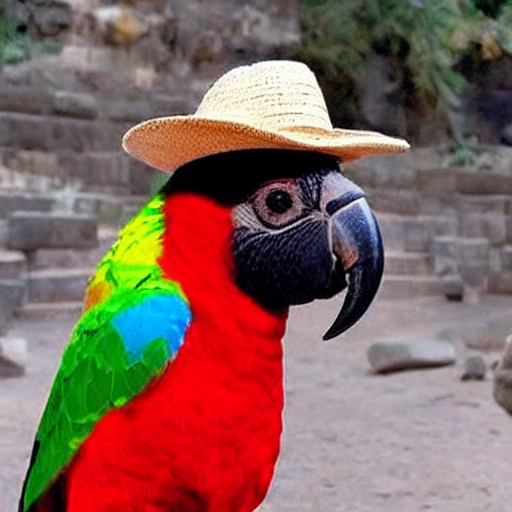} 
        \phantom{a}
        &
        
        \includegraphics[width=\ww,frame]{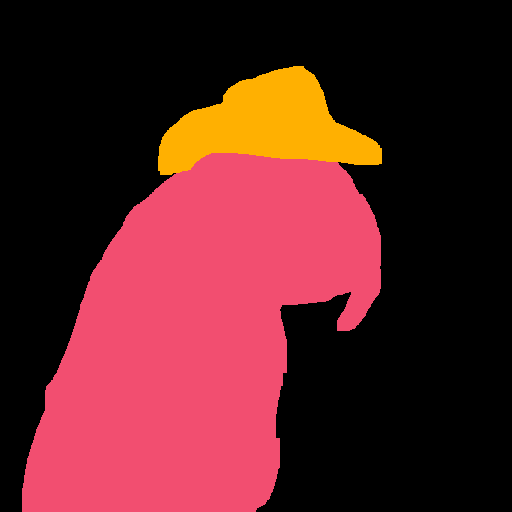} & 
        \includegraphics[width=\ww,frame]{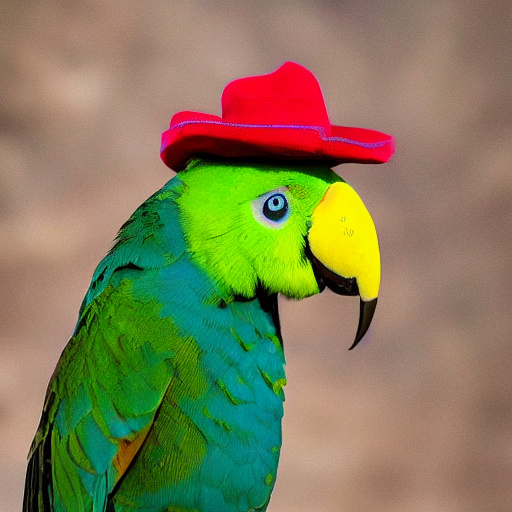} 
        \phantom{a}
        \\

        \begin{tabular}{c}
            \maska{``a colorful parrot''} \\ 
            \maskb{``a straw hat''} \\
        \end{tabular} &&

        \begin{tabular}{c}
            \maska{``a green parrot''} \\ 
            \maskb{``a red hat''} \\
        \end{tabular}

    \end{tabular}
    \caption{\textbf{Additional examples of our method:} Each pair consists of an (i) input global text (top left, black), a spatio-textual representation describing each segment using free-form text prompts (left, colored text and sketches), and (ii) the corresponding generated image (right). As can be seen, \name is able to generate high-quality images that correspond to both the global text and spatio-textual representation content. Please note that the colors are for illustration purposes only, and do not affect the actual inputs.}
    \label{fig:method_additional_examples1}
    \vspace{-1em}
\end{figure*}

%% file: figures/method_examples/additional2.tex
\begin{figure*}[ht]
    \centering
    \setlength{\tabcolsep}{1pt}
    \renewcommand{\arraystretch}{0.5}
    \setlength{\ww}{0.49\columnwidth}
    \begin{tabular}{cccc}

        ``sitting on a wooden floor'' &&
        ``in the street''
        \\

        \includegraphics[width=\ww,frame]{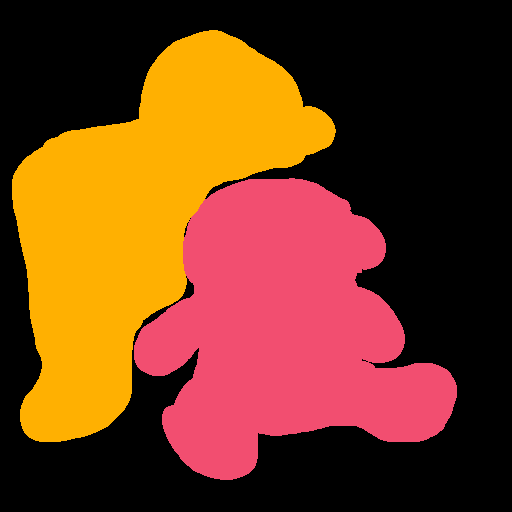} & 
        \includegraphics[width=\ww,frame]{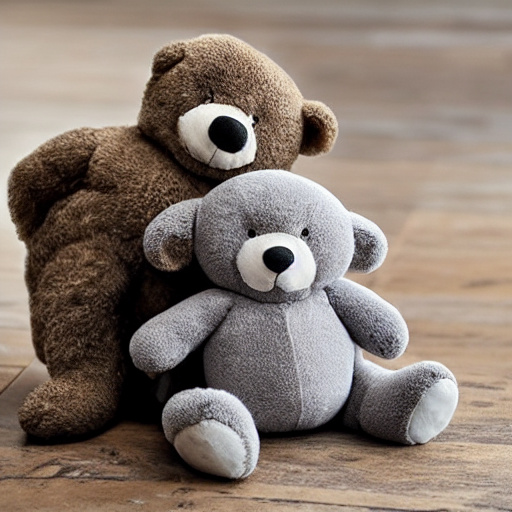} 
        \phantom{a}
        &
        
        \includegraphics[width=\ww,frame]{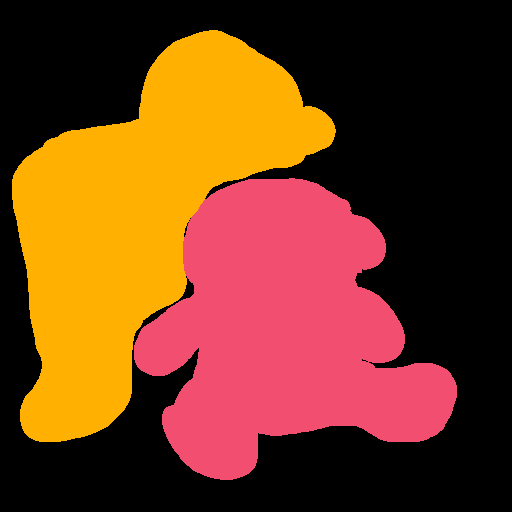} & 
        \includegraphics[width=\ww,frame]{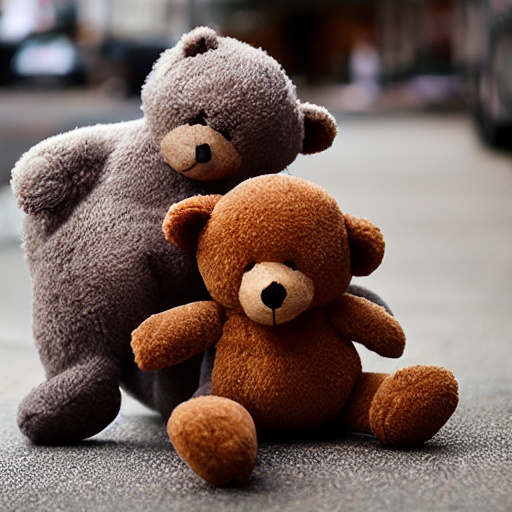}
        \phantom{a}
        \\

        \begin{tabular}{c}
            \maska{``a gray teddy bear''} \\ 
            \maskb{``a brown teddy bear''} \\
            \\
        \end{tabular} &&

        \begin{tabular}{c}
            \maska{``a brown teddy bear''} \\ 
            \maskb{``a gray teddy bear''} \\
            \\
        \end{tabular}
        \\
        \\

        ``a night with the city &&
        ``in a sunny day near
        \\
        in the background'' &&
        near the forest''
        \\

        \includegraphics[width=\ww,frame]{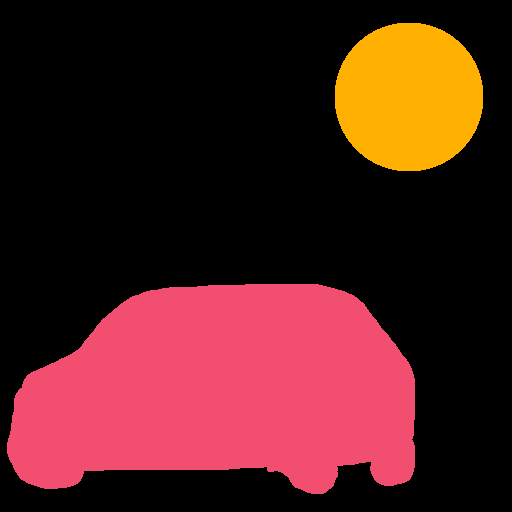} & 
        \includegraphics[width=\ww,frame]{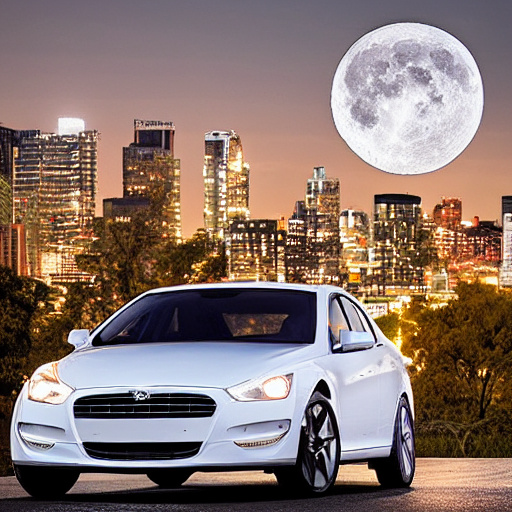}
        \phantom{a}
        &
        
        \includegraphics[width=\ww,frame]{figures/method_examples/assets/car_moon/vis.png} & 
        \includegraphics[width=\ww,frame]{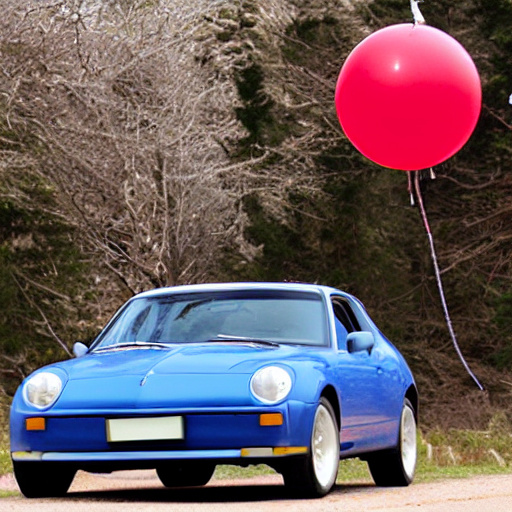}
        \phantom{a}
        \\

        \begin{tabular}{c}
            \maska{``a white car''} \\ 
            \maskb{``a big full moon''} \\
            \\
        \end{tabular} &&

        \begin{tabular}{c}
            \maska{``a blue car''} \\ 
            \maskb{``a red balloon''} \\
            \\
        \end{tabular}
        \\
        \\

        ``in an empty room'' &&
        ``day outdoors''
        \\

        \includegraphics[width=\ww,frame]{figures/method_examples/assets/robot_painting/vis.png} & 
        \includegraphics[width=\ww,frame]{figures/method_examples/assets/robot_painting/pred.jpg}
        \phantom{a}
        &
        
        \includegraphics[width=\ww,frame]{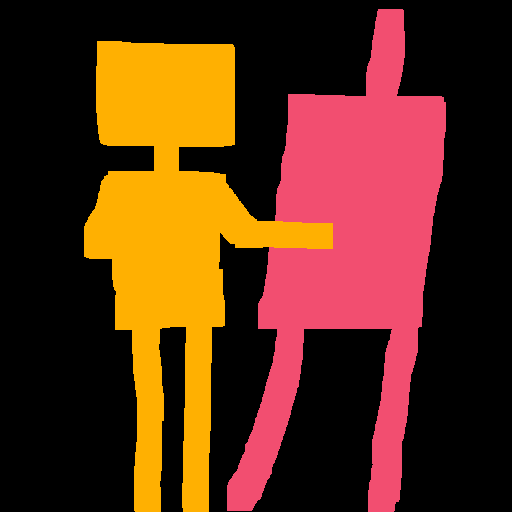} & 
        \includegraphics[width=\ww,frame]{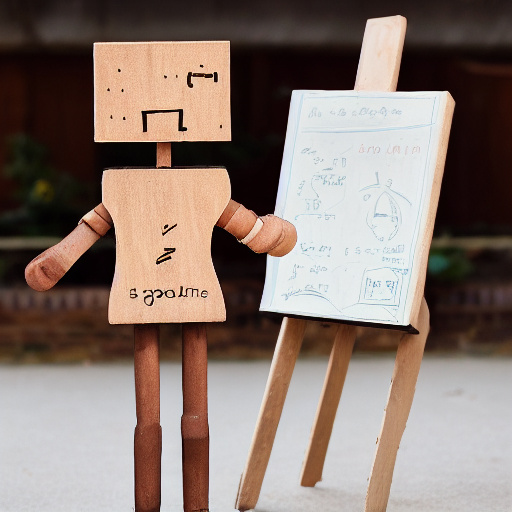}
        \phantom{a}
        \\

        \begin{tabular}{c}
            \maska{``a canvas with a painting} \\
            \maska{of a Corgi dog''} \\
            \maskb{``a metallic yellow robot''} \\
        \end{tabular} &&

        \begin{tabular}{c}
            \maska{``a canvas with} \\
            \maska{math equations''} \\
            \maskb{``a wooden robot''} \\
        \end{tabular}

    \end{tabular}
    \caption{\textbf{Additional examples of our method:} Each pair consists of an (i) input global text (top left, black), a spatio-textual representation describing each segment using free-form text prompts (left, colored text and sketches), and (ii) the corresponding generated image (right). As can be seen, \name is able to generate high-quality images that correspond to both the global text and spatio-textual representation content. Please note that the colors are for illustration purposes only, and do not affect the actual inputs.}
    \label{fig:method_additional_examples2}
\end{figure*}

%% file: figures/method_examples/additional3.tex
\begin{figure*}[ht]
    \centering
    \setlength{\tabcolsep}{1pt}
    \renewcommand{\arraystretch}{0.5}
    \setlength{\ww}{0.49\columnwidth}
    \begin{tabular}{cccc}

        ``on a wooden table outdoors'' &&
        ``on a concrete floor''
        \\

        \includegraphics[width=\ww,frame]{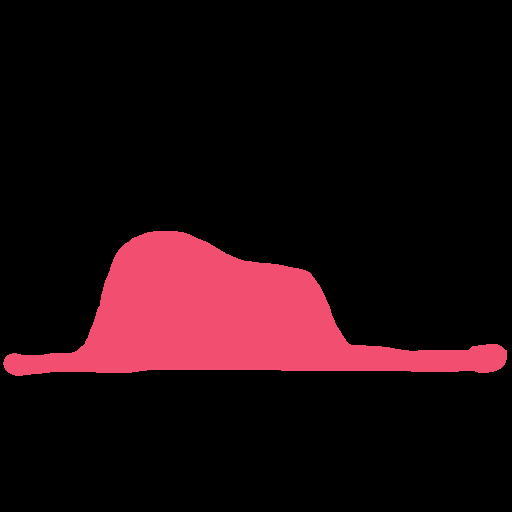} & 
        \includegraphics[width=\ww,frame]{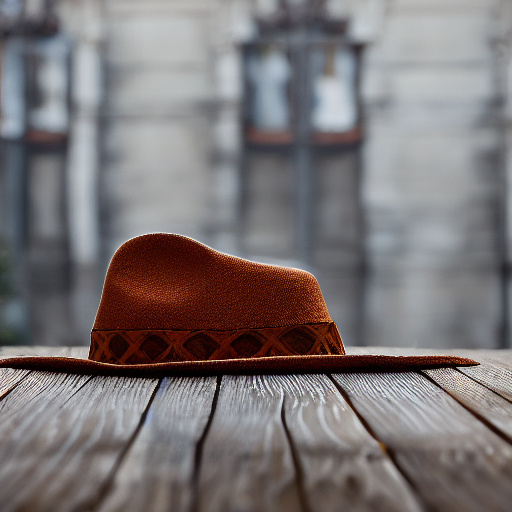} 
        \phantom{a}
        &
        
        \includegraphics[width=\ww,frame]{figures/method_examples/assets/little_prince_hat/vis.png} & 
        \includegraphics[width=\ww,frame]{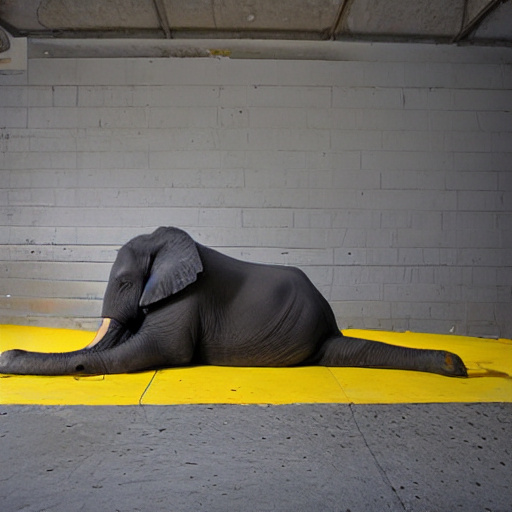}
        \phantom{a}
        \\

        \begin{tabular}{c}
            \maska{``a brown hat''} \\ 
            \\
        \end{tabular} &&

        \begin{tabular}{c}
            \maska{``an elephant''} \\ 
            \\
        \end{tabular}
        \\
        \\

        ``next to a wooden house'' &&
        ``indoors''
        \\

        \includegraphics[width=\ww,frame]{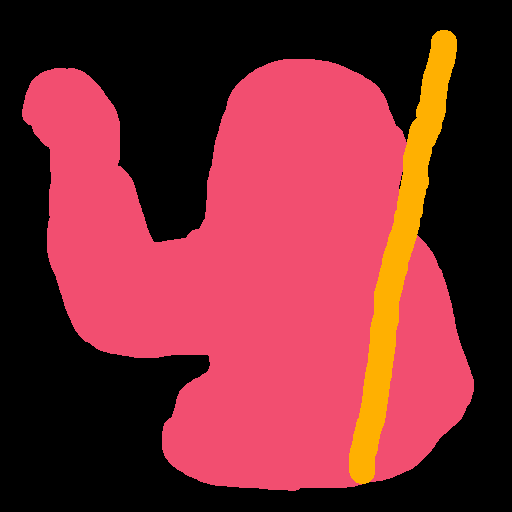} & 
        \includegraphics[width=\ww,frame]{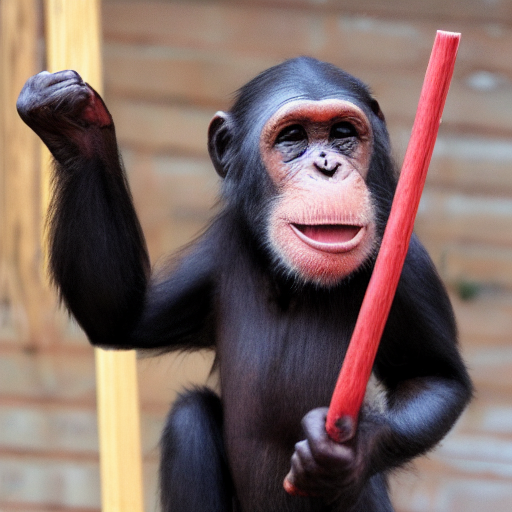}
        \phantom{a}
        &
        
        \includegraphics[width=\ww,frame]{figures/method_examples/assets/monkey_stick/vis.png} & 
        \includegraphics[width=\ww,frame]{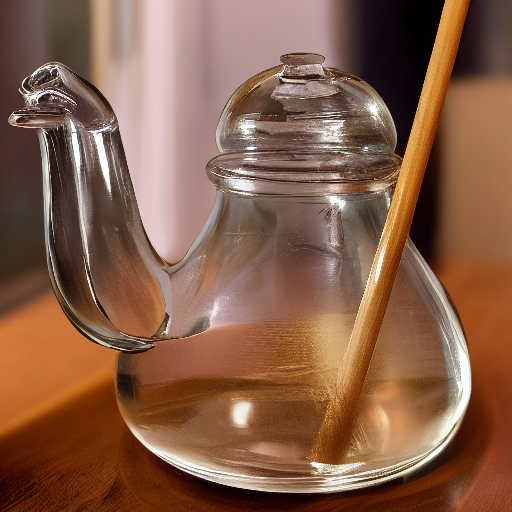}
        \phantom{a}
        \\

        \begin{tabular}{c}
            \maska{``a chimpanzee''} \\ 
            \maskb{``a red wooden stick''} \\ 
            \\
        \end{tabular} &&

        \begin{tabular}{c}
            \maska{``a glass tea pot''} \\ 
            \maskb{``a golden straw''} \\ 
            \\
        \end{tabular}
        \\
        \\

        &&
        ``a black and white photo
        \\

        ``on the grass'' &&
        in the desert''
        \\

        \includegraphics[width=\ww,frame]{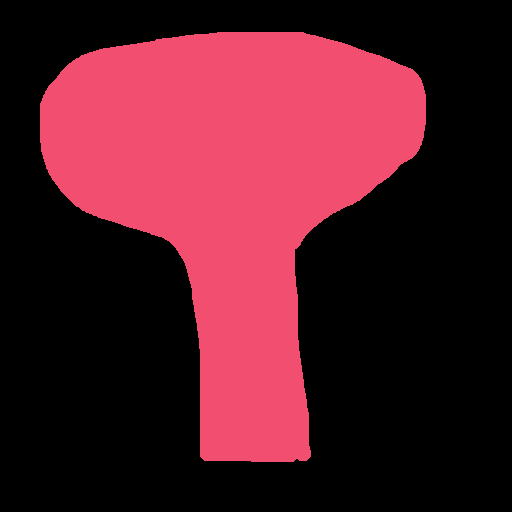} & 
        \includegraphics[width=\ww,frame]{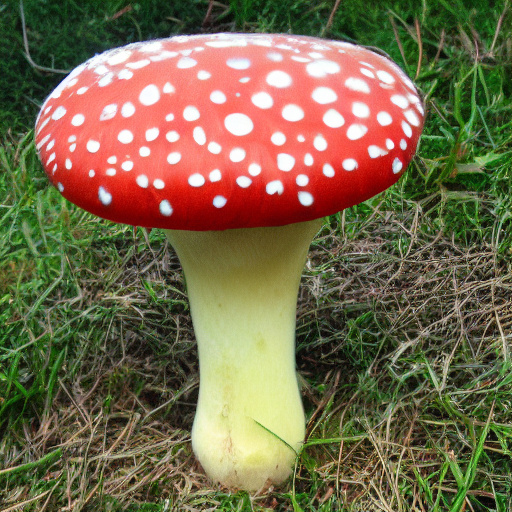}
        \phantom{a}
        &
        
        \includegraphics[width=\ww,frame]{figures/method_examples/assets/mushroom/vis.png} & 
        \includegraphics[width=\ww,frame]{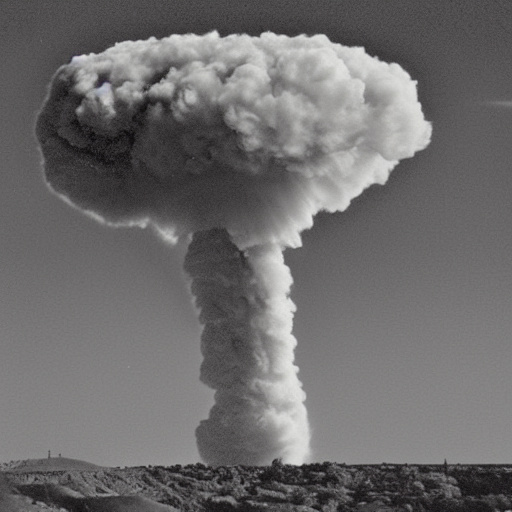}
        \phantom{a}
        \\

        \begin{tabular}{c}
            \maska{``an Amanita mushroom''}
        \end{tabular} &&

        \begin{tabular}{c}
            \maska{``a nuclear explosion''}
        \end{tabular}

    \end{tabular}
    \caption{\textbf{Additional examples of our method:} Each pair consists of an (i) input global text (top left, black), a spatio-textual representation describing each segment using free-form text prompts (left, colored text and sketches), and (ii) the corresponding generated image (right). As can be seen, \name is able to generate high-quality images that correspond to both the global text and spatio-textual representation content. Please note that the colors are for illustration purposes only, and do not affect the actual inputs.}
    \label{fig:method_additional_examples3}
\end{figure*}

%% file: figures/method_examples/additional4.tex
\begin{figure*}[ht]
    \centering
    \setlength{\tabcolsep}{1pt}
    \renewcommand{\arraystretch}{0.5}
    \setlength{\ww}{0.49\columnwidth}
    \begin{tabular}{cccc}

        ``a portrait photo'' &&
        ``a portrait photo''
        \\

        \includegraphics[width=\ww,frame]{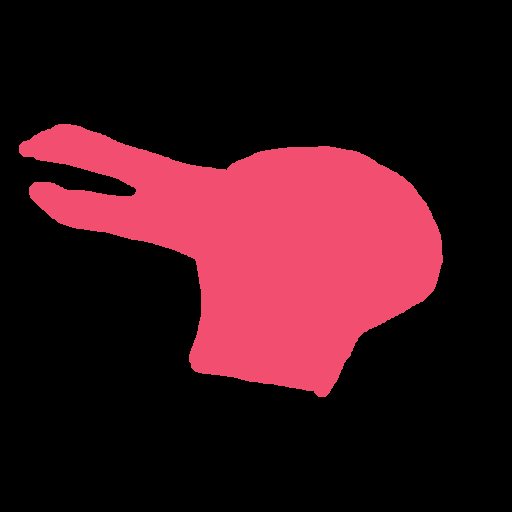} & 
        \includegraphics[width=\ww,frame]{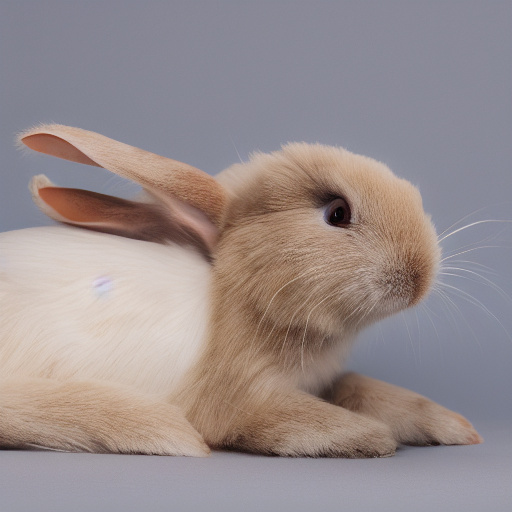} 
        \phantom{a}
        &
        
        \includegraphics[width=\ww,frame]{figures/method_examples/assets/rabbit_head/vis.png} & 
        \includegraphics[width=\ww,frame]{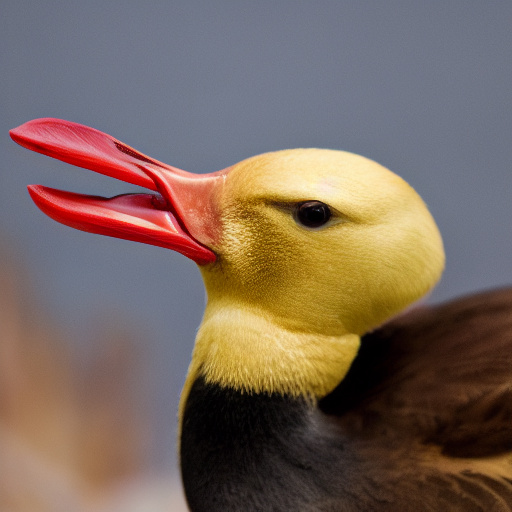}
        \phantom{a}
        \\

        \begin{tabular}{c}
            \maska{``a rabbit''} \\ 
            \\
        \end{tabular} &&

        \begin{tabular}{c}
            \maska{``a duck''} \\ 
            \\
        \end{tabular}
        \\
        \\

        ``under the sun'' &&
        ``inside a lake''
        \\

        \includegraphics[width=\ww,frame]{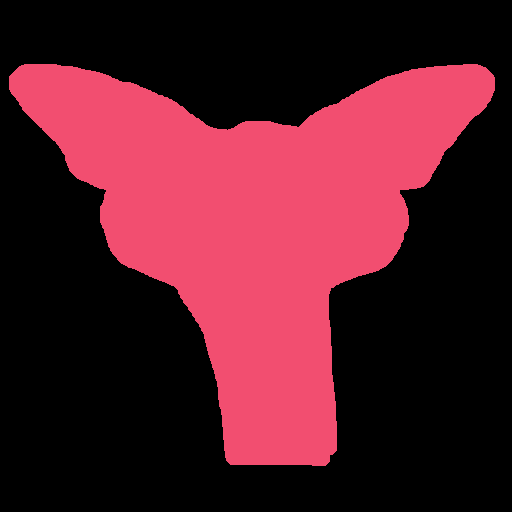} & 
        \includegraphics[width=\ww,frame]{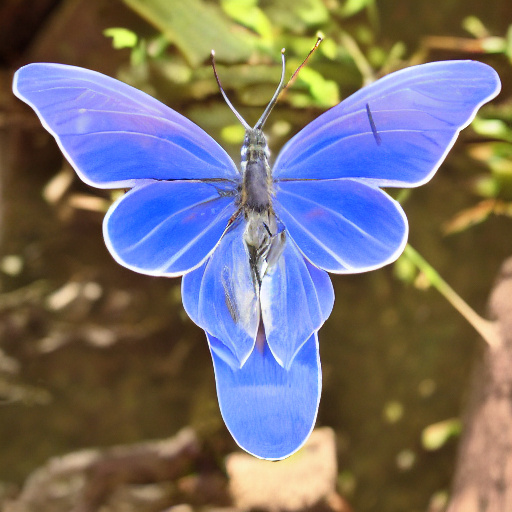} 
        \phantom{a}
        &
        
        \includegraphics[width=\ww,frame]{figures/method_examples/assets/butterfly/vis.png} & 
        \includegraphics[width=\ww,frame]{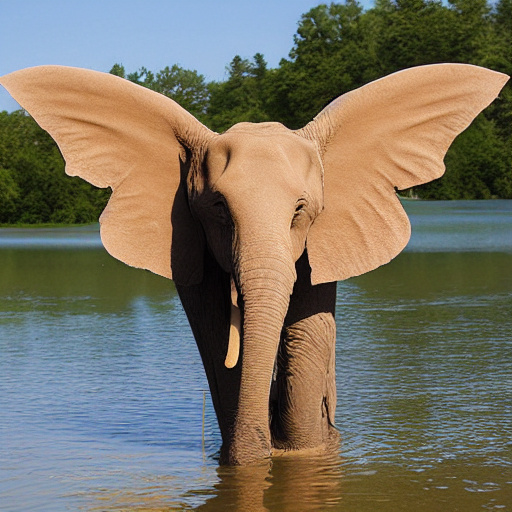}
        \phantom{a}
        \\

        \begin{tabular}{c}
            \maska{``a blue butterfly''} \\ 
            \\
        \end{tabular} &&

        \begin{tabular}{c}
            \maska{``an elephant''} \\ 
            \\
        \end{tabular}
        \\
        \\

        ``on a snowy day'' &&
        ``a sunny day at the street''
        \\

        \includegraphics[width=\ww,frame]{figures/method_examples/assets/mouse_boxing/vis.png} & 
        \includegraphics[width=\ww,frame]{figures/method_examples/assets/mouse_boxing/pred_mouse2.jpg} 
        \phantom{a}
        &
        
        \includegraphics[width=\ww,frame]{figures/method_examples/assets/mouse_boxing/vis.png} & 
        \includegraphics[width=\ww,frame]{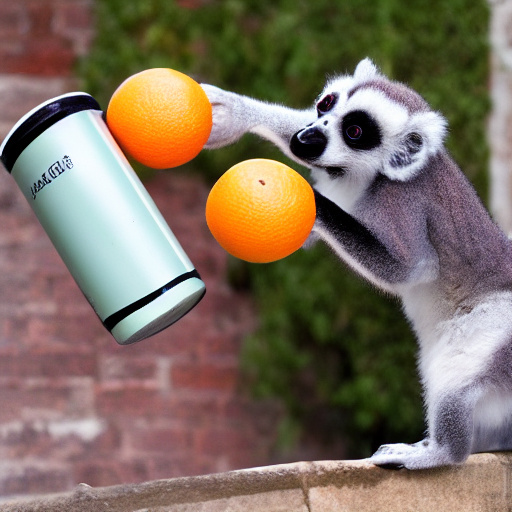}
        \phantom{a}
        \\

        \begin{tabular}{c}
            \maska{``a mouse''} \\
            \maskb{``boxing gloves''} \\
            \maskc{``a black punching bag''} \\
        \end{tabular} &&

        \begin{tabular}{c}
            \maska{``a lemur''} \\
            \maskb{``oranges''} \\
            \maskc{``a soda can''} \\
        \end{tabular}

    \end{tabular}
    \caption{\textbf{Additional examples of our method:} Each pair consists of an (i) input global text (top left, black), a spatio-textual representation describing each segment using free-form text prompts (left, colored text and sketches), and (ii) the corresponding generated image (right). As can be seen, \name is able to generate high-quality images that correspond to both the global text and spatio-textual representation content. Please note that the colors are for illustration purposes only, and do not affect the actual inputs.}
    \label{fig:method_additional_examples4}
\end{figure*}

%% file: figures/method_examples/additional5.tex
\begin{figure*}[ht]
    \centering
    \setlength{\tabcolsep}{1pt}
    \renewcommand{\arraystretch}{0.5}
    \setlength{\ww}{0.49\columnwidth}
    \begin{tabular}{cccc}

        ``a sunny day near
        \\
        the Eiffel tower'' &&
        ``room with sunlight''
        \\

        \includegraphics[width=\ww,frame]{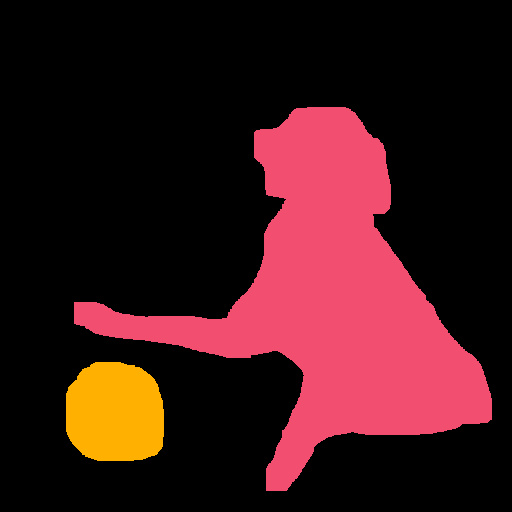} & 
        \includegraphics[width=\ww,frame]{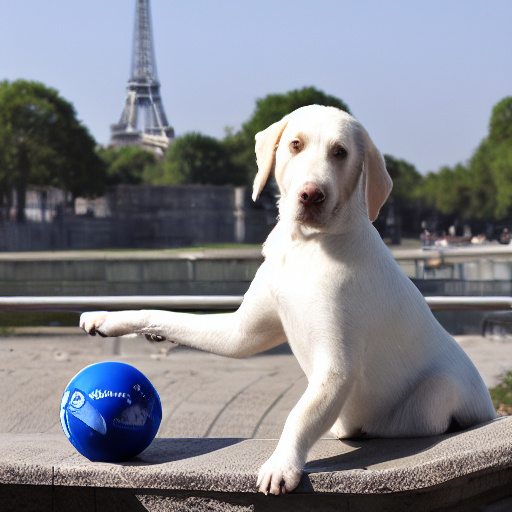} 
        \phantom{a}
        &
        
        \includegraphics[width=\ww,frame]{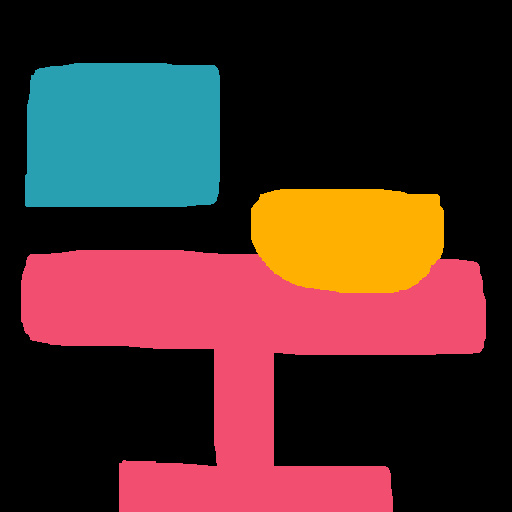} & 
        \includegraphics[width=\ww,frame]{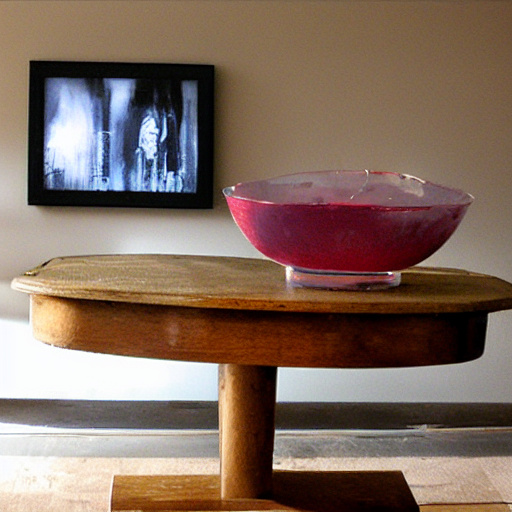}
        \phantom{a}
        \\

        \begin{tabular}{c}
            \maska{``a white Labrador''} \\
            \maskb{``a blue ball''} \\ 
            \\
        \end{tabular} &&

        \begin{tabular}{c}
            \maska{``a wooden table''} \\
            \maskb{``a red bowl''} \\ 
            \maskc{``a picture on the wall''} \\ 
        \end{tabular}

    \end{tabular}
    \caption{\textbf{Additional examples of our method:} Each pair consists of an (i) input global text (top left, black), a spatio-textual representation describing each segment using free-form text prompts (left, colored text and sketches), and (ii) the corresponding generated image (right). As can be seen, \name is able to generate high-quality images that correspond to both the global text and spatio-textual representation content. Please note that the colors are for illustration purposes only, and do not affect the actual inputs.}
    \label{fig:method_additional_examples5}
\end{figure*}

%% file: figures/additional_mask_sensitivity1/fig.tex
\begin{figure*}[t]
    \centering
    \setlength{\tabcolsep}{1pt}
    \renewcommand{\arraystretch}{0.5}
    \setlength{\ww}{0.5\columnwidth}
  
    \begin{tabular}{cccc}

        ``a sunny day outdoors''
        \\

        \includegraphics[width=\ww,frame]{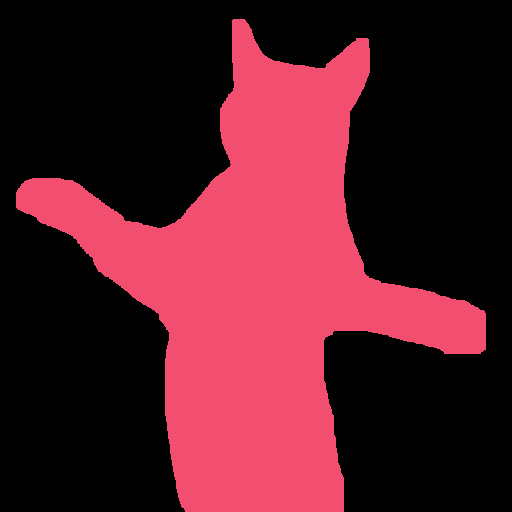} 
        &
        \includegraphics[width=\ww,frame]{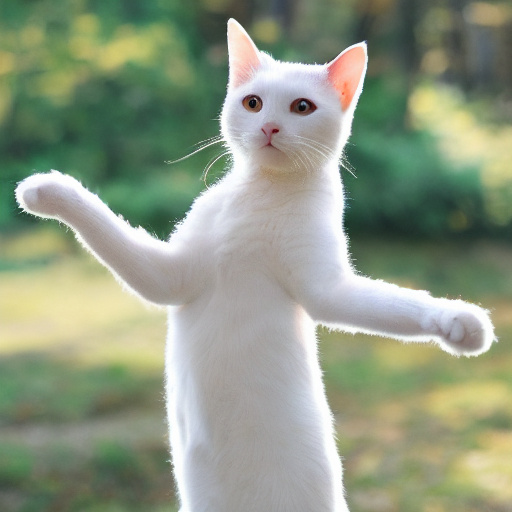} &
        \includegraphics[width=\ww,frame]{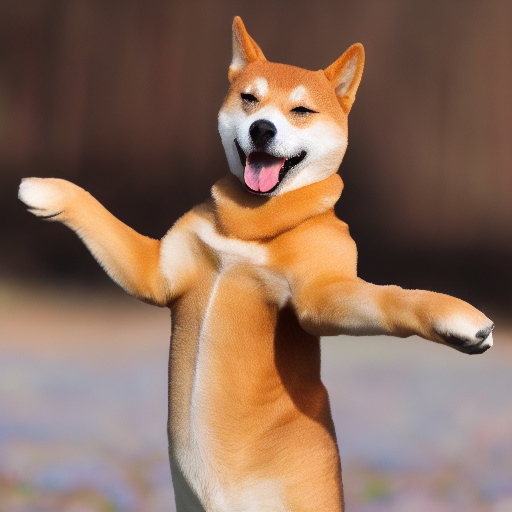} &
        \includegraphics[width=\ww,frame]{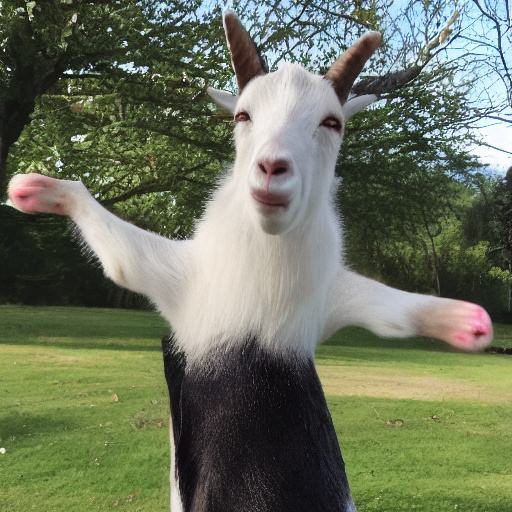}
        \\

        &
        \maska{``a white cat''} &
        \maska{``a Shiba Inu dog''} &
        \maska{``a goat''}
        \\
        \\
        \\

        \includegraphics[width=\ww,frame]{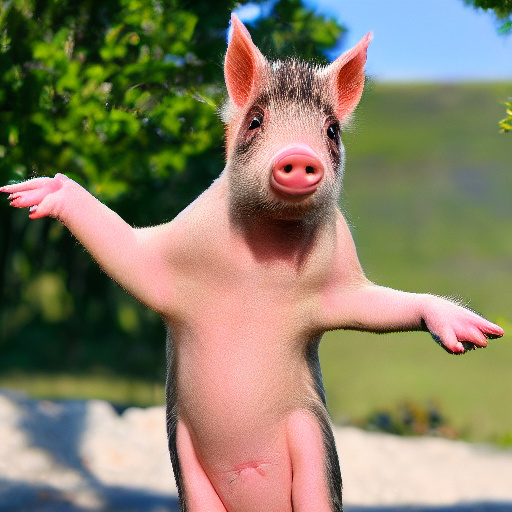} &
        \includegraphics[width=\ww,frame]{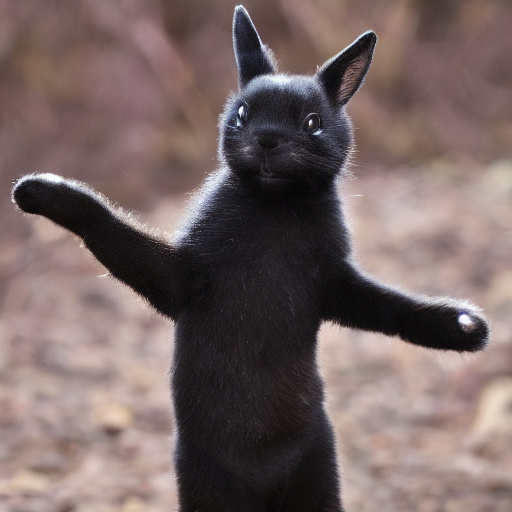} &
        \includegraphics[width=\ww,frame]{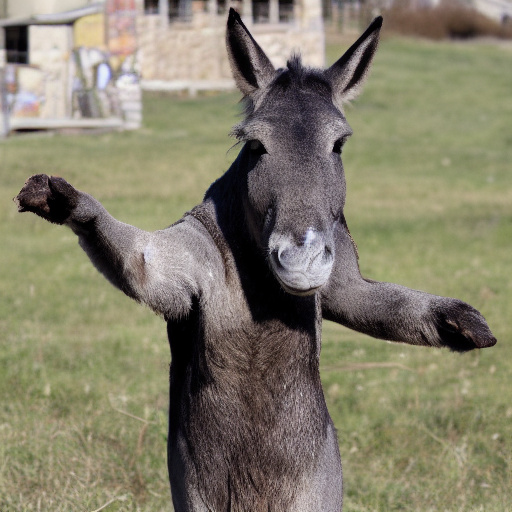} &
        \includegraphics[width=\ww,frame]{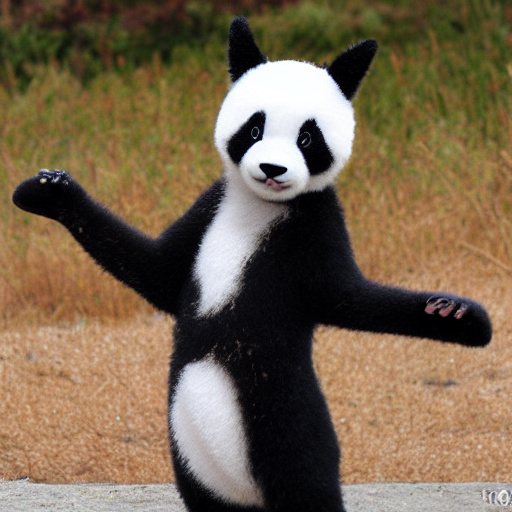}
        \\

        \maska{``a pig''} &
        \maska{``a black rabbit''} &
        \maska{``a gray donkey''} &
        \maska{``a panda bear''}
        \\
        \\
        \\

        \includegraphics[width=\ww,frame]{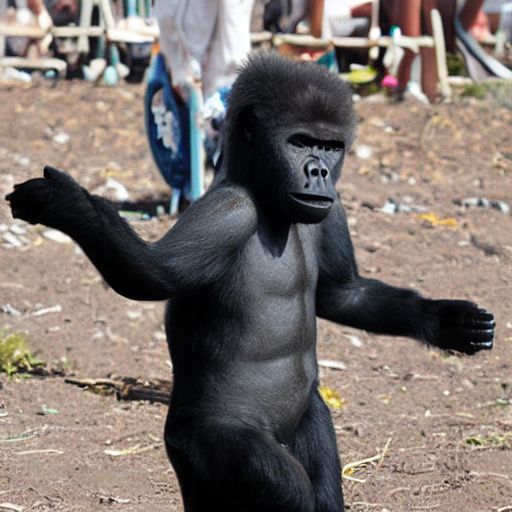} &
        \includegraphics[width=\ww,frame]{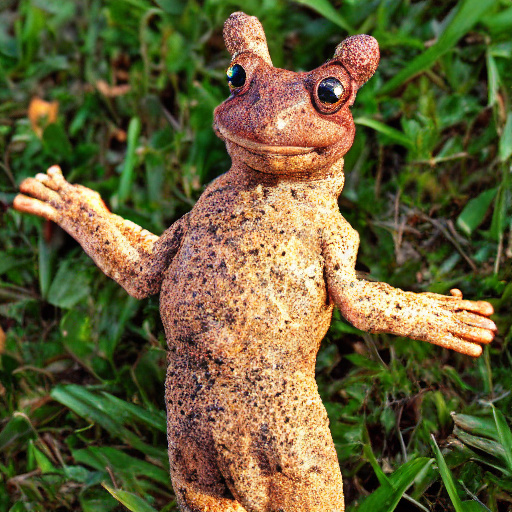} &
        \includegraphics[width=\ww,frame]{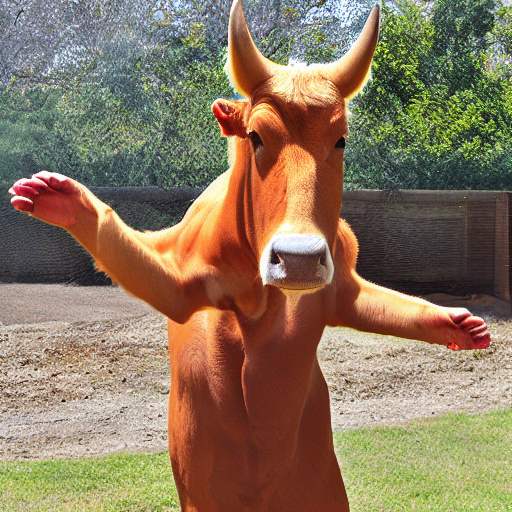} &
        \includegraphics[width=\ww,frame]{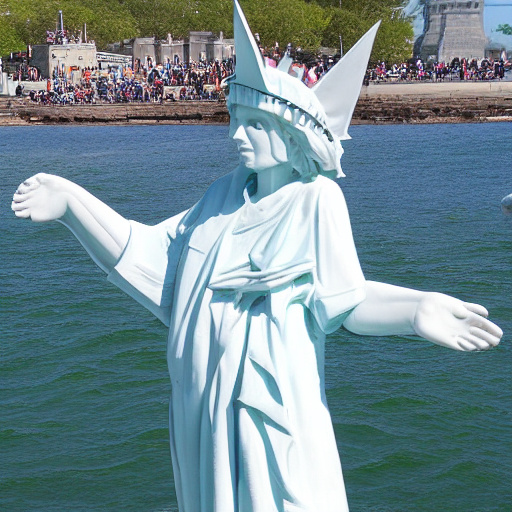}
        \\

        \maska{``a gorilla''} &
        \maska{``a toad''} &
        \maska{``a cow''} &
        \maska{``The Statue of Liberty''}
        \\
        \\
        \\

        \includegraphics[width=\ww,frame]{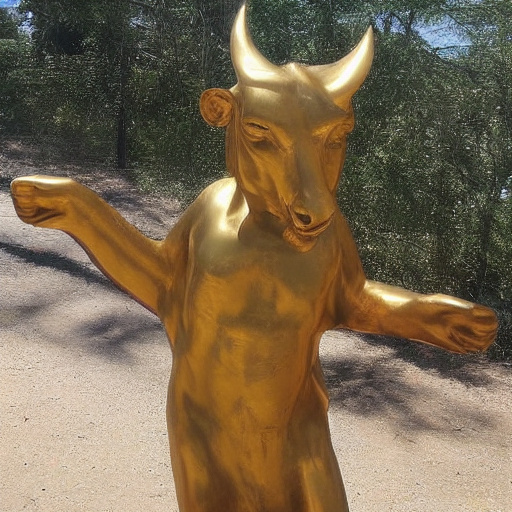} &
        \includegraphics[width=\ww,frame]{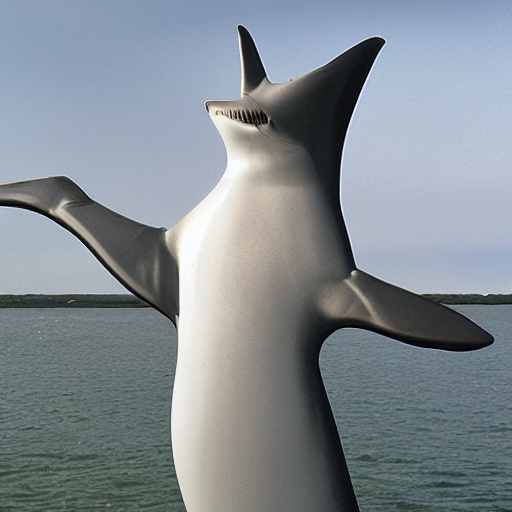} &
        \includegraphics[width=\ww,frame]{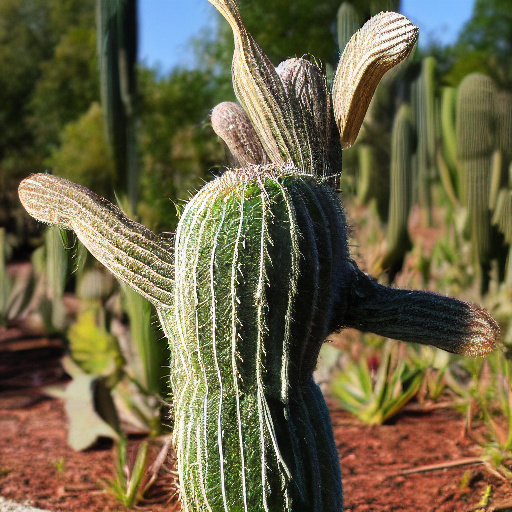} &
        \includegraphics[width=\ww,frame]{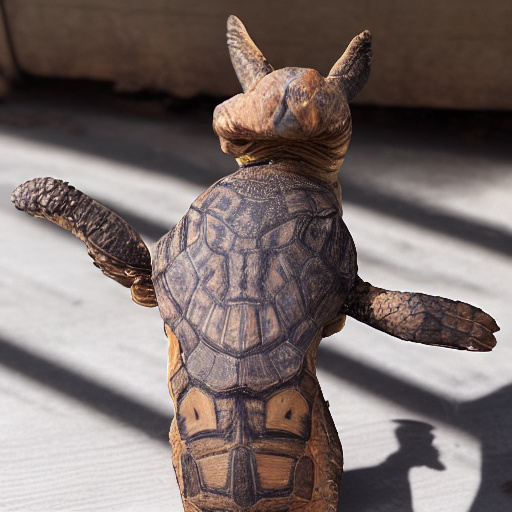}
        \\

        \maska{``a golden calf''} &
        \maska{``a shark''} &
        \maska{``a cactus''} &
        \maska{``a tortoise''}
        \\
        \\
    \end{tabular}
    
    \caption{\textbf{Mask insensitivity:} We found that the model is relatively insensitive to inaccuracies in the input mask. Given a general animal shape mask (top left), the model is able to generate a diverse set of results driven by the different local prompts. It changes the body type according to the local prompt, while leaving the overall posture of the character intact.}
    \label{fig:additional_mask_sensitivity1}
\end{figure*}

%% file: figures/additional_mask_sensitivity2/fig.tex
\begin{figure*}[t]
    \centering
    \setlength{\tabcolsep}{1pt}
    \renewcommand{\arraystretch}{0.5}
    \setlength{\ww}{0.5\columnwidth}
  
    \begin{tabular}{cccc}

        ``a painting''
        \\

        \includegraphics[width=\ww,frame]{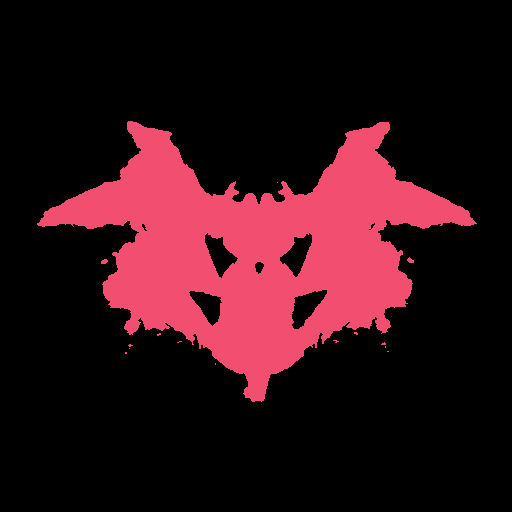} 
        &
        \includegraphics[width=\ww,frame]{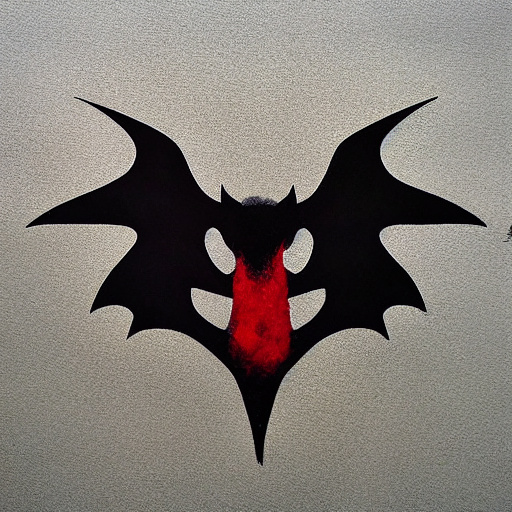} &
        \includegraphics[width=\ww,frame]{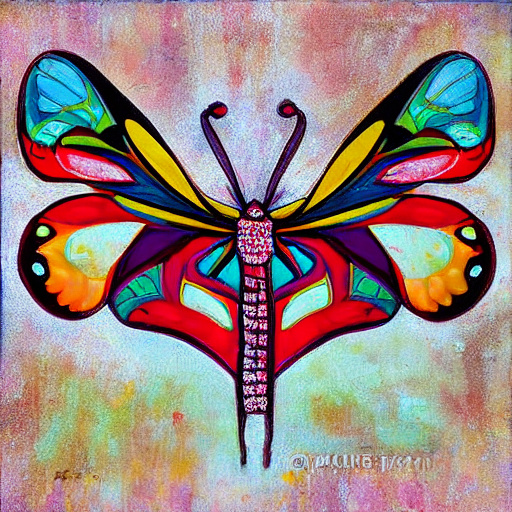} &
        \includegraphics[width=\ww,frame]{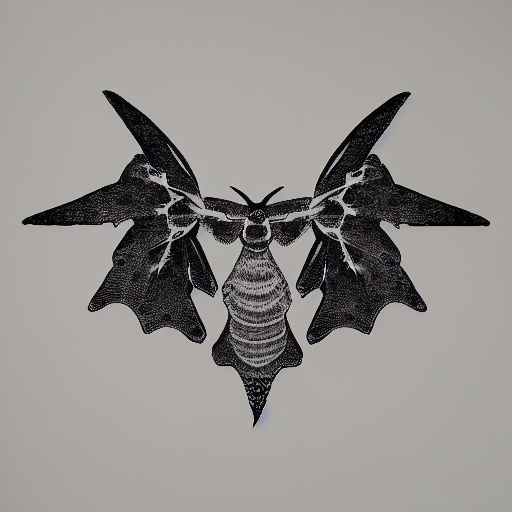}
        \\

        &
        \maska{``a bat''} &
        \maska{``a colorful butterfly''} &
        \maska{``a moth''}
        \\
        \\
        \\

        \includegraphics[width=\ww,frame]{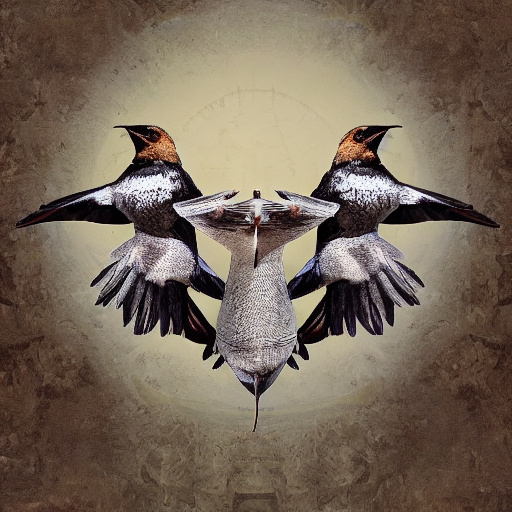} 
        &
        \includegraphics[width=\ww,frame]{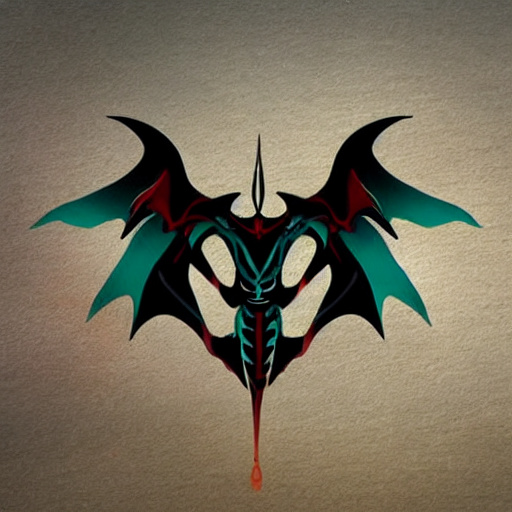} &
        \includegraphics[width=\ww,frame]{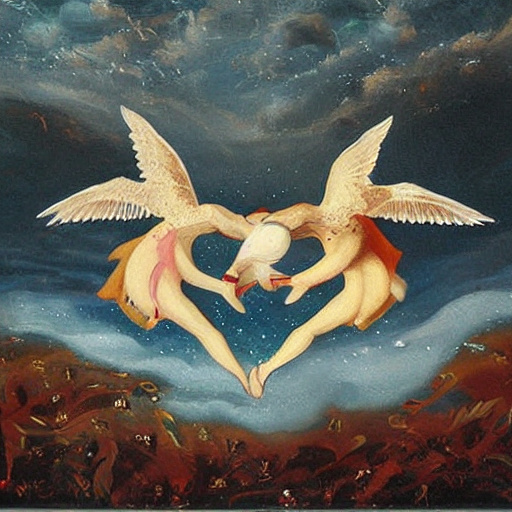} &
        \includegraphics[width=\ww,frame]{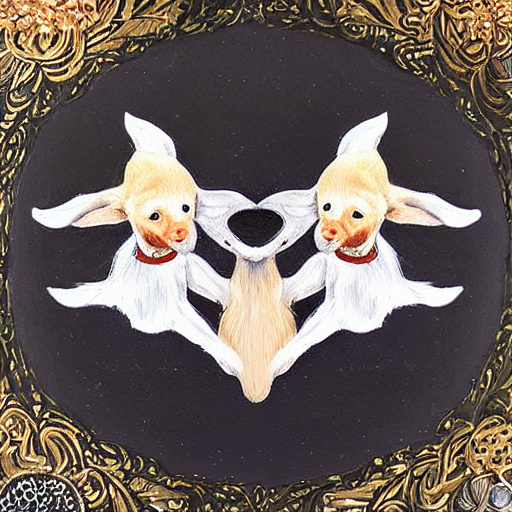}
        \\

        \maska{``two birds facing away} &
        \maska{``a dragon''} &
        \maska{``mythical creatures''} &
        \maska{``two dogs''}
        \\
        \maska{from each other''} &
        \\
        \\

        \includegraphics[width=\ww,frame]{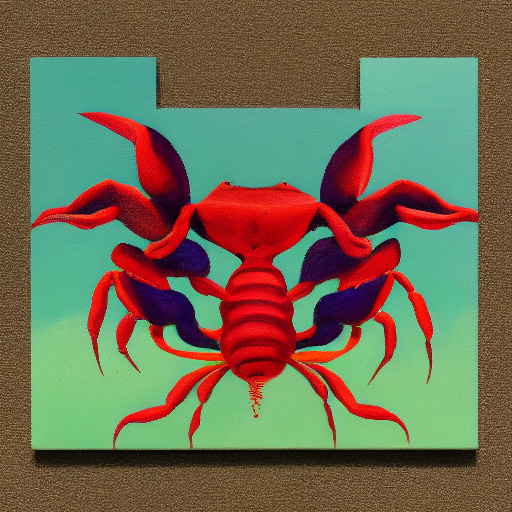} 
        &
        \includegraphics[width=\ww,frame]{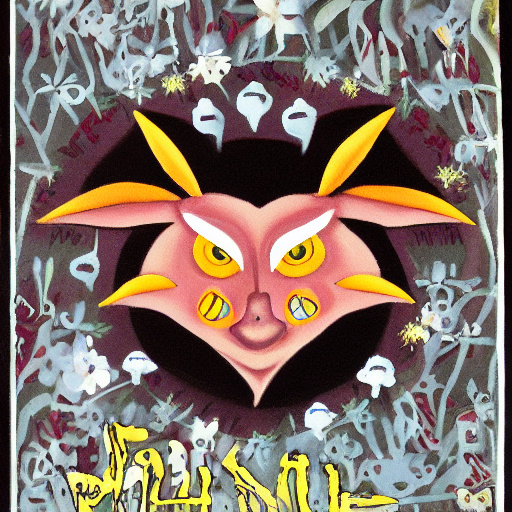} &
        \includegraphics[width=\ww,frame]{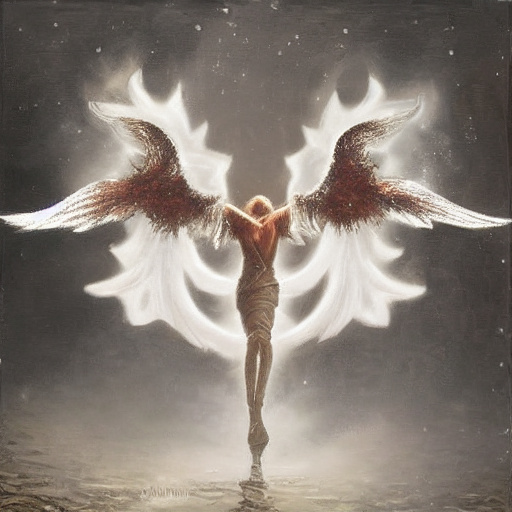} &
        \includegraphics[width=\ww,frame]{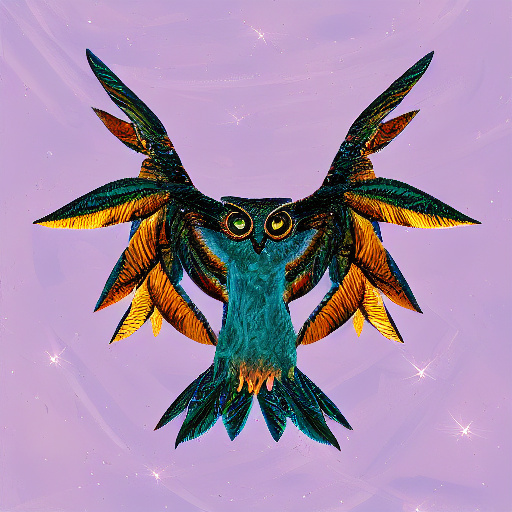}
        \\

        \maska{``a crab''} &
        \maska{``an evil pig''} &
        \maska{``a flying angel''} &
        \maska{``an owl''}
        \\
        \\

        \includegraphics[width=\ww,frame]{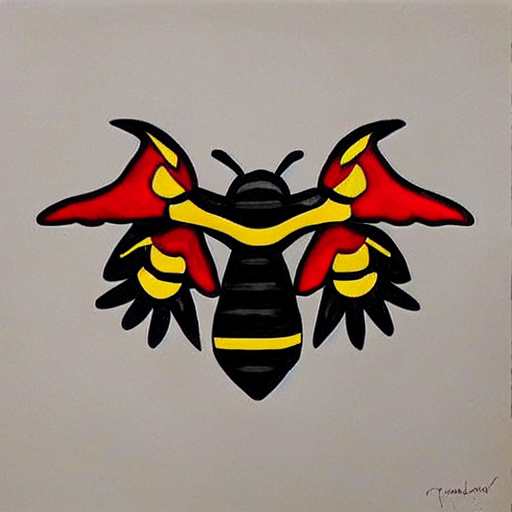} 
        &
        \includegraphics[width=\ww,frame]{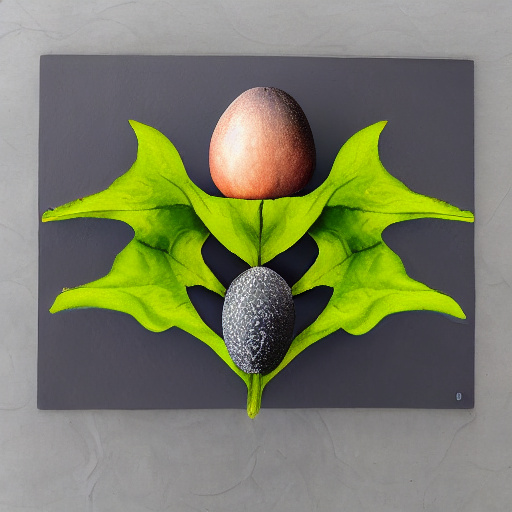} &
        \includegraphics[width=\ww,frame]{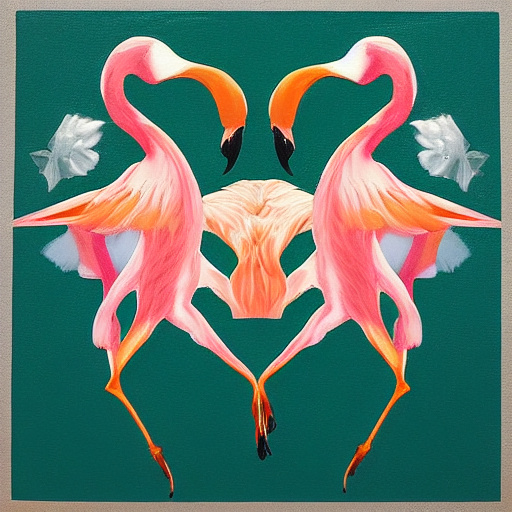} &
        \includegraphics[width=\ww,frame]{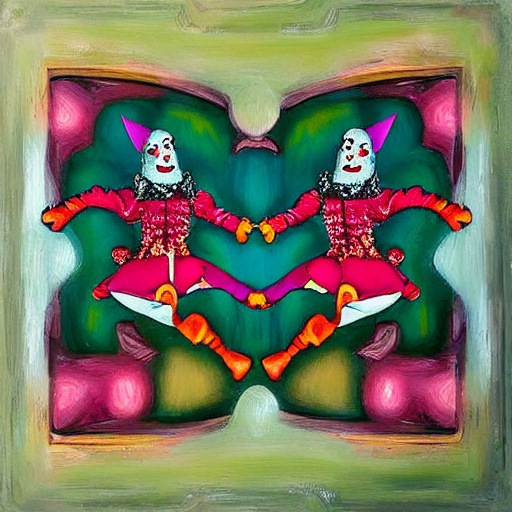}
        \\

        \maska{``a bee''} &
        \maska{``an avocado''} &
        \maska{``two flamingos''} &
        \maska{``two clowns''}
        \\
        \\
    \end{tabular}
    
    \caption{\textbf{Mask insensitivity:} We found that the model is relatively insensitive to inaccuracies in the input mask. Given a general Rorschach \cite{klopfer1942rorschach} mask (top left), the model is able to generate a diverse set of results driven by the different local prompts. It changes fine-details according to the local prompt, while leaving the overall general shape intact.}
    \label{fig:additional_mask_sensitivity2}
\end{figure*}

%% file: figures/additional_multiscale_control1/fig.tex
\begin{figure*}[ht]
    \centering
    
    \centering
    \setlength{\tabcolsep}{1pt}
    \renewcommand{\arraystretch}{0.5}
    \setlength{\ww}{0.33\columnwidth}
  
    \begin{tabular}{cccccc}

        \scriptsize{``at the desert''} &
        \scriptsize{(1)} &
        \scriptsize{(2)} &
        \scriptsize{(3)} &
        \scriptsize{(4)} &
        \scriptsize{(5)}
        \\

        \includegraphics[width=\ww,frame]{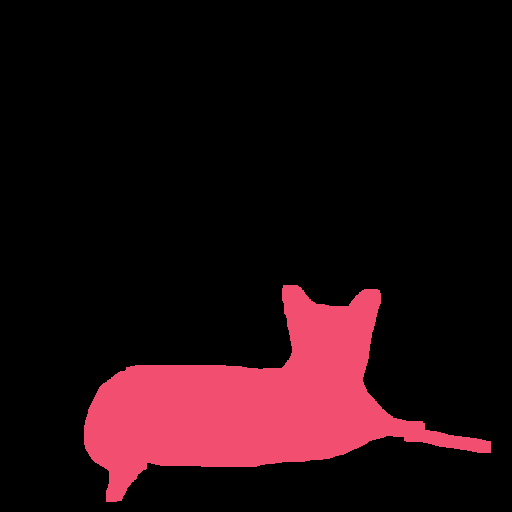}
        \phantom{a}
        &
        \includegraphics[width=\ww,frame]{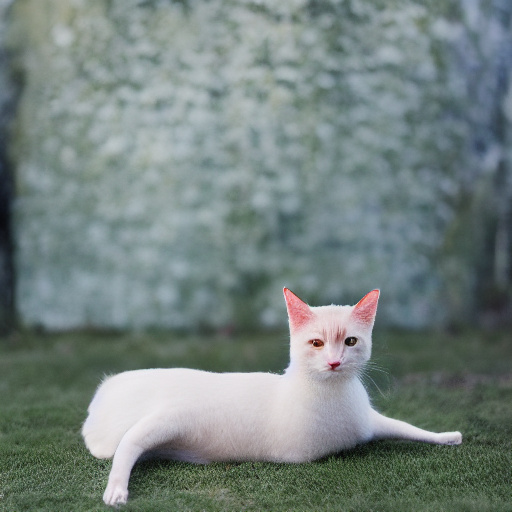} &
        \includegraphics[width=\ww,frame]{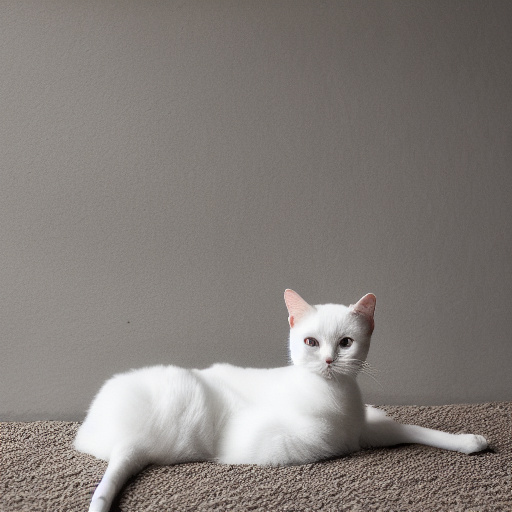} &
        \includegraphics[width=\ww,frame]{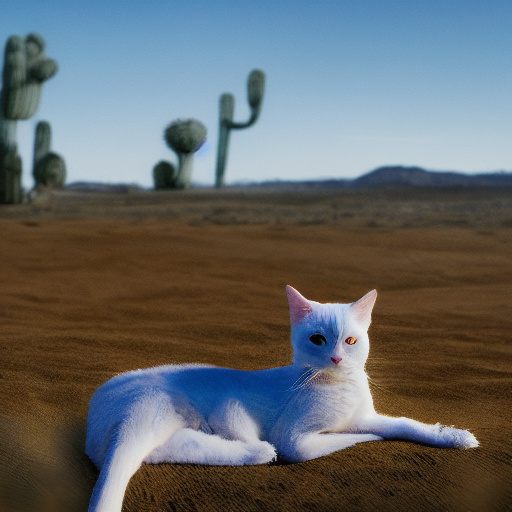} &
        \includegraphics[width=\ww,frame]{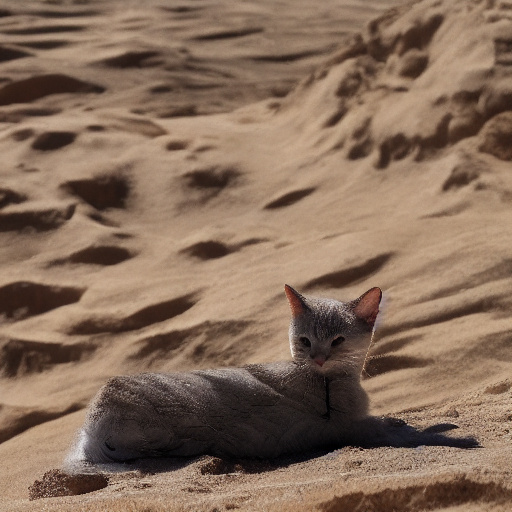} &
        \includegraphics[width=\ww,frame]{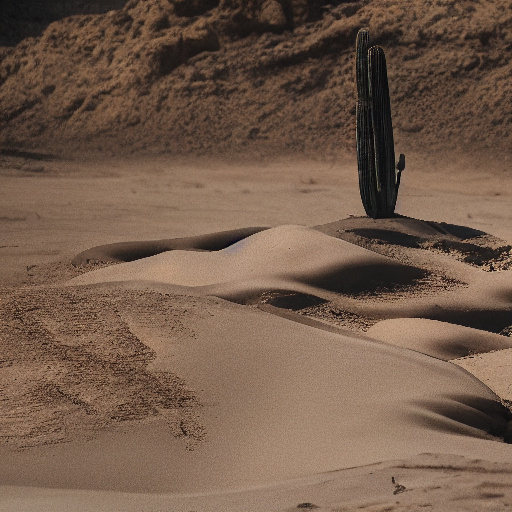}
        \\
        
        \maska{\scriptsize{``a white cat''}} &
        \scriptsize{$s_{\text{global}} = 0 ; s_{\text{local}} = 3$} &
        \scriptsize{$s_{\text{global}} = 1.5 ; s_{\text{local}} = 3$} &
        \scriptsize{$s_{\text{global}} = 3 ; s_{\text{local}} = 3$} &
        \scriptsize{$s_{\text{global}} = 3 ; s_{\text{local}} = 1.5$} &
        \scriptsize{$s_{\text{global}} = 3 ; s_{\text{local}} = 0$}
        \\
    \end{tabular}
    
    \caption{\textbf{Multi-scale control:} Using the multi-scale inference allows fine-grained control over the input conditions. Given the same inputs (left), we can use different scales for each condition. In this example, if we put all the weight on the local scene (1), the generated image contains a cat with the correct color and posture, but not at the desert. Conversely, if we place all the weight on the global text (5), we get an image of a desert with no cat in it. The in-between results correspond to a mix of conditions --- in (4) we get a gray cat with slightly different posture, in (2) the cat sits on dirt, but not in the desert, and in (3) we get a white cat at the desert.}
    \label{fig:additional_multiscale_control1}
\end{figure*}

%% file: figures/additional_multiscale_control2/fig.tex
\begin{figure*}[ht]
    \centering
    
    \centering
    \setlength{\tabcolsep}{1pt}
    \renewcommand{\arraystretch}{0.5}
    \setlength{\ww}{0.33\columnwidth}
  
    \begin{tabular}{cccccc}

        \scriptsize{``at the park''} &
        \scriptsize{(1)} &
        \scriptsize{(2)} &
        \scriptsize{(3)} &
        \scriptsize{(4)} &
        \scriptsize{(5)}
        \\

        \includegraphics[width=\ww,frame]{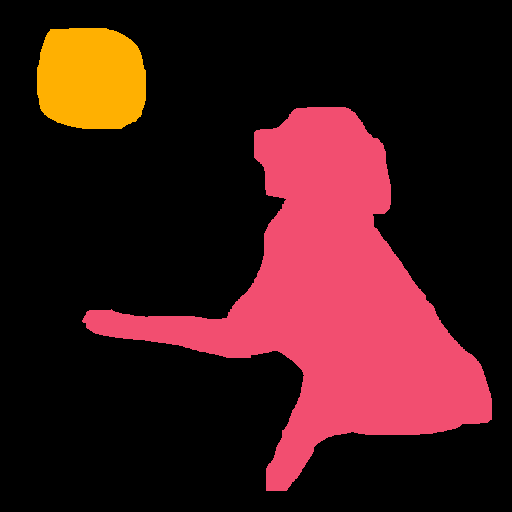}
        \phantom{a}
        &
        \includegraphics[width=\ww,frame]{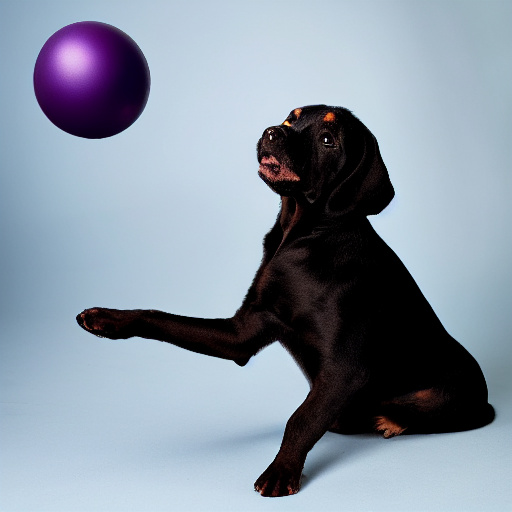} &
        \includegraphics[width=\ww,frame]{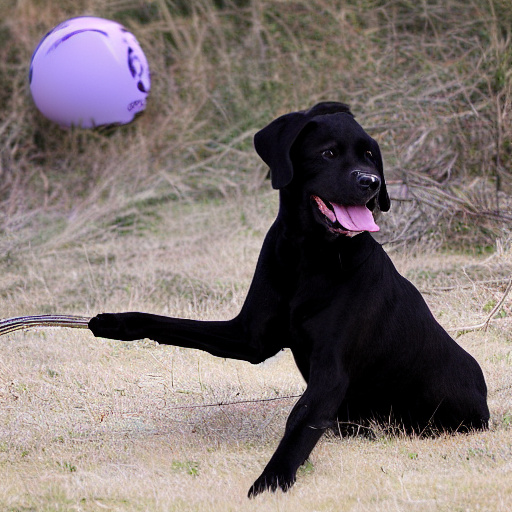} &
        \includegraphics[width=\ww,frame]{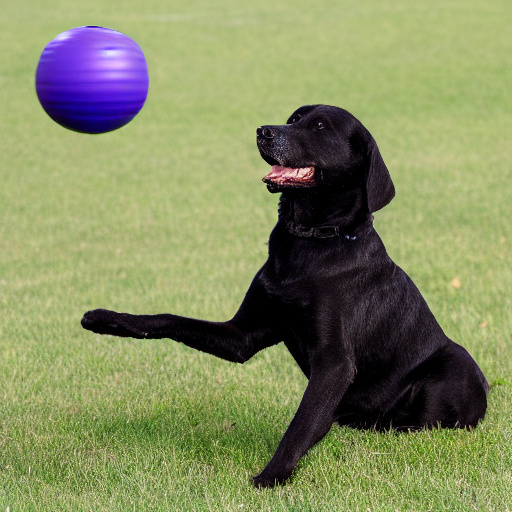} &
        \includegraphics[width=\ww,frame]{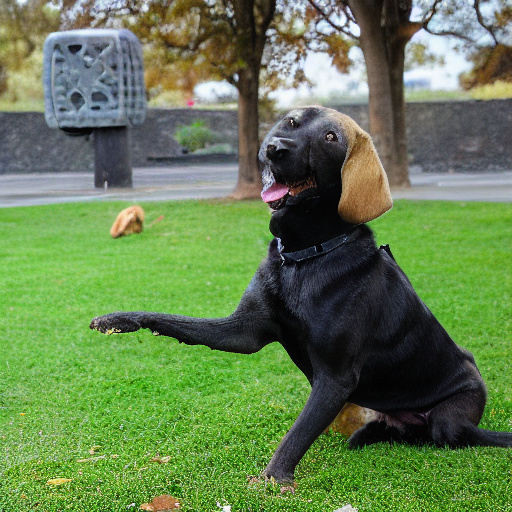} &
        \includegraphics[width=\ww,frame]{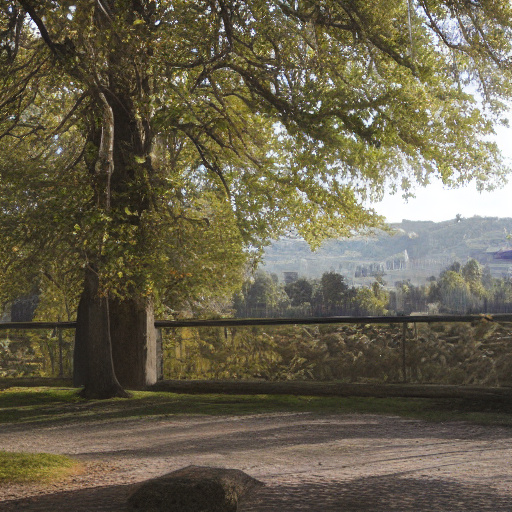}
        \\
        
        \maska{\scriptsize{``a black Labrador dog''}} &
        \scriptsize{$s_{\text{global}} = 0 ; s_{\text{local}} = 3$} &
        \scriptsize{$s_{\text{global}} = 1.5 ; s_{\text{local}} = 3$} &
        \scriptsize{$s_{\text{global}} = 3 ; s_{\text{local}} = 3$} &
        \scriptsize{$s_{\text{global}} = 3 ; s_{\text{local}} = 1.5$} &
        \scriptsize{$s_{\text{global}} = 3 ; s_{\text{local}} = 0$}
        \\

        \maskb{\scriptsize{``a purple ball''}}
        \\
    \end{tabular}
    
    \caption{\textbf{Multi-scale control:} Using the multi-scale inference allows fine-grained control over the input conditions. Given the same inputs (left), we can use different scales for each condition. In this example, if we put all the weight on the local scene (1), the generated image contains a Labrador dog and a purple ball with the correct color and posture, but not at the park. Conversely, if we place all the weight on the global text (5), we get an image of a park with no dog or ball in it. The in-between results correspond to a mix of conditions --- in (4) we get a gray brick instead of a purple ball, in (2) the dog is outside but not in the park, and in (3) we get a black Labrador dog and a purple ball in the park.}
    \label{fig:additional_multiscale_control2}
\end{figure*}

%% file: figures/additional_limitations/fig.tex
\begin{figure*}[ht]
    \centering
    \setlength{\tabcolsep}{1pt}
    \renewcommand{\arraystretch}{0.5}
    \setlength{\ww}{0.47\columnwidth}
    \begin{tabular}{cccccc}

        &
        ``on an asphalt road'' &&&
        ``on a blue table''
        \\

        \rotatebox[origin=c]{90}{(1)} &
        \includegraphics[valign=c, width=\ww,frame]{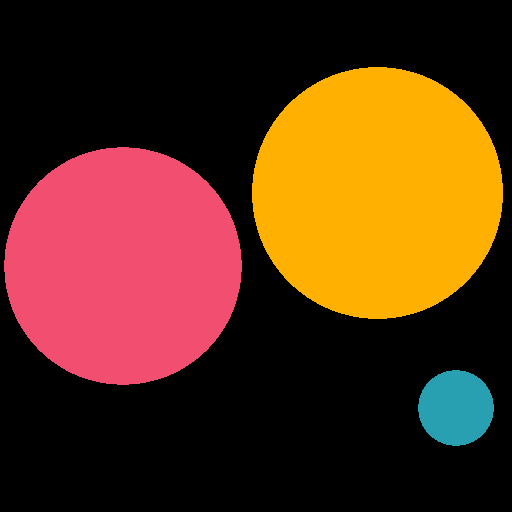} & 
        \includegraphics[valign=c, width=\ww,frame]{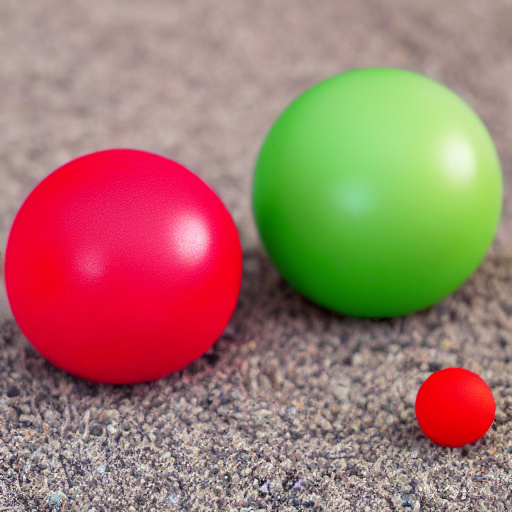}
        \phantom{a}
        &
        
        \rotatebox[origin=c]{90}{(2)} &
        \includegraphics[valign=c, width=\ww,frame]{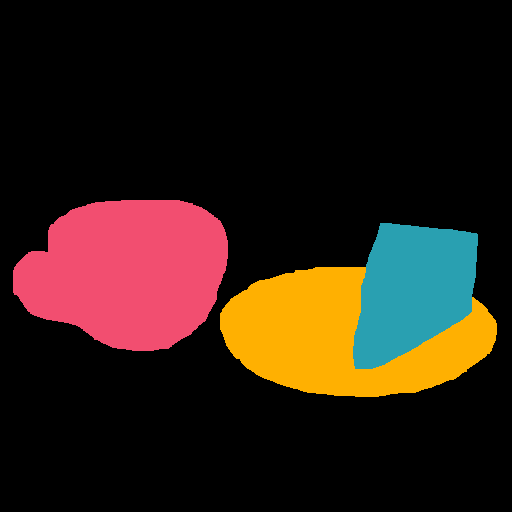} & 
        \includegraphics[valign=c, width=\ww,frame]{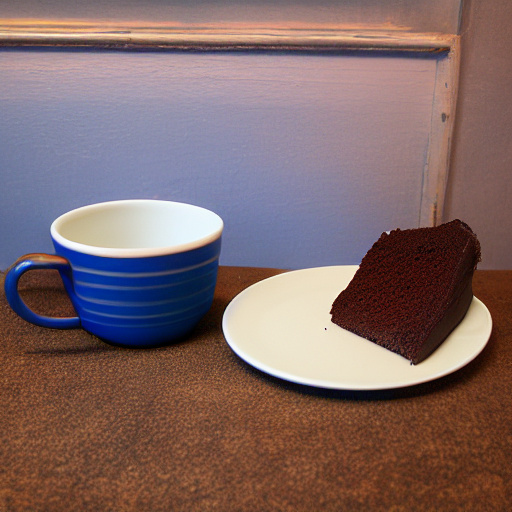}
        \phantom{a}
        \\

        &
        \begin{tabular}{c}
            \maska{``a red ball''} \\ 
            \maskb{``a green ball''} \\
            \maskc{``a blue ball''} \\
            \\
        \end{tabular} &&&

        \begin{tabular}{c}
            \maska{``a red mug''} \\ 
            \maskb{``a white plate''} \\
            \maskc{``a chocolate cake''} \\
            \\
        \end{tabular}
        \\
        \\

        &
        ``in the forest'' &&&
        ``above the desert''
        \\

        \rotatebox[origin=c]{90}{(3)} &
        \includegraphics[valign=c,width=\ww,frame]{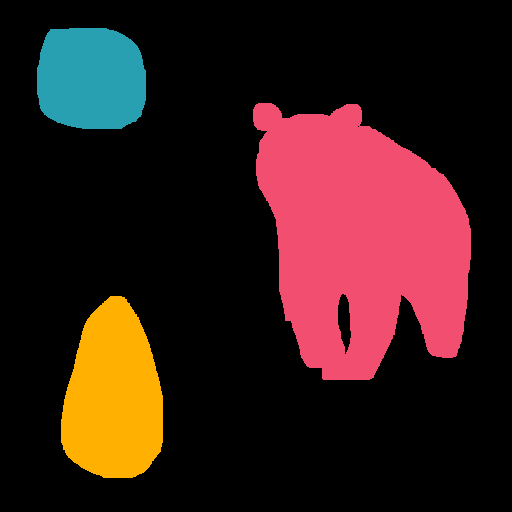} & 
        \includegraphics[valign=c,width=\ww,frame]{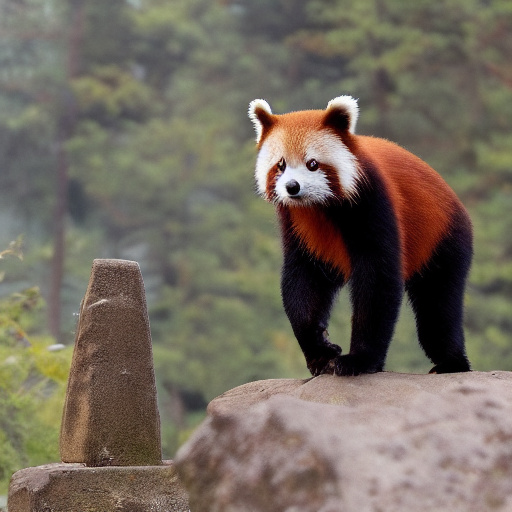}
        \phantom{a}
        &
        
        \rotatebox[origin=c]{90}{(4)} &
        \includegraphics[valign=c, width=\ww,frame]{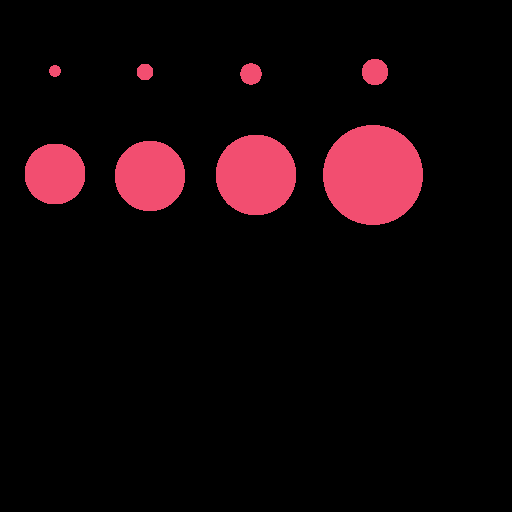} & 
        \includegraphics[valign=c, width=\ww,frame]{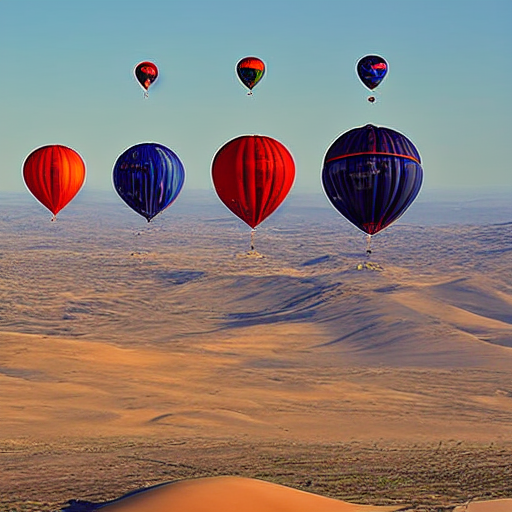}
        \phantom{a}
        \\

        &
        \begin{tabular}{c}
            \maska{``a red panda''} \\ 
            \maskb{``a big rock''} \\
            \maskc{``a sunrise''} \\
            \\
        \end{tabular} &&&

        \begin{tabular}{c}
            \maska{``hot air balloons''} \\ 
            \\
            \\
            \\
        \end{tabular}
        \\
    \end{tabular}
    \caption{\textbf{Limitations:} In some cases there is a ``characteristics leakage'' between segments, as in example (1) where instead of a blue ball we get another red ball, or a leakage between the global text and some segments, as in example (2) where the mug is generated in a blue color originated in the global text. In other cases, the model ignores some of the objects, as the sun in example (3) and the smallest hot air balloon in example (4).}
    \label{fig:additonal_limitations}
\end{figure*}

%% file: sections/appendix/implementation_details.tex
\section{Implementation Details}
\label{sec:implementation_details}

In the following section, we describe the implementation details that were omitted from the main paper. In \Cref{sec:diffusion_models_implementation_details} we start by describing the diffusion models implementation details. Then, in \Cref{sec:spatio_textual_implementation_detials} we describe the implementation details of our spatio-textual representation. Later, in \Cref{sec:baselines_implementation_details} we describe the implementation details of the baselines and how we adapt them to our problem setting. Afterwards, in \Cref{sec:dataset_implementation_details} we describe the implementation details of the automatic input creation process that we used to compute our automatic metrics. Finally, in \Cref{sec:user_study_details} we describe the details of the user study.

\subsection{Diffusion Models Implementation Details}
\label{sec:diffusion_models_implementation_details}
We based our approach on two state-of-the-art diffusion-based text-to-image models: \DALLE~2 \cite{ramesh2022hierarchical} and Stable Diffusion \cite{rombach2022high}. We trained these models on a custom-made dataset of 35M image-text pairs, following Make-A-Scene~\cite{gafni2022make}.

\subsubsection{\DALLE~2 Implementation Details}
\label{sec:dalle2_implementation_deails}
Since the implementation of \DALLE~2 is not available to the public, we re-implemented it following the details included in their paper~\cite{ramesh2022hierarchical}. This model consists of the following submodules, given an $(x, y)$ image-text pair:
\begin{itemize}
    \item \textbf{A decoder model D}: that is trained to translate $\clipimg{x}$ into a $64\times64$ resolution image $x$.
    \item \textbf{A super-resolution model SR}: that is trained to upsample the $64\times64$ resolution image $x$ into $256\times256$.
    \item \textbf{A prior model P}: that is trained to translate the tuples $(\cliptxt{y}, \text{BytePairEncoding}(y))$ into $\clipimg{x}$.
\end{itemize}
Concatenating the above three models yields a text-to-image model $SR \circ D \circ P$.

In order to adapt the model to the task of text-to-image generation with sparse scene control, we chose to fine-tune the decoder $D$. For the fine-tuning we used the standard simple loss variant of Ho \etal{} \cite{ho2020denoising}:
\begin{equation}
    L_{\text{simple}} = E_{t,x_0,\epsilon}\left[ || \epsilon - \epsilon_{\theta}(x_t, \clipimg{x_0}, ST, t) ||^2 \right]
\end{equation}
where $\epsilon_{\theta}$ is a UNet \cite{long2015fully} model that predicts the added noise at each time step $t$, $x_t$ is the noisy image at time step $t$ and $ST$ is our spatio-textual representation. To this loss, we added the same variational lower bound (VLB) loss as in \cite{nichol2021improved} to get the total loss of:
\begin{equation}
    L_{hybrid} = L_{simple} + \lambda L_{VLB}
\end{equation}
we set $\lambda = 0.001$ in our experiments.
We used Adam optimizer \cite{kingma2014adam} with $\beta_1=0.9$ and $\beta_2=0.999$ with learning rate $6\times10^{-5}$ for $64{,}000$ iterations.

During inference, we utilize composition of the CLIP text encoder $\cliptxtd$ and the prior model $P$ to infer the CLIP image embedding for both the spatio-textural representation $ST$ and for the global text prompt $P \circ \cliptxt{\tglobal}$. We used the DDIM \cite{song2020denoising} inference method with a different number of inference steps for each component: $50$ steps for the prior model, $250$ for the decoder, and $100$ for the super resolution model. 

\subsubsection{Stable Diffusion Implementation Details}
\label{sec:stable_diffusion_deails}
For Stable Diffusion \cite{rombach2022high} we used the official implementation \cite{stable_diffusion_implementation} and the official pre-trained v1.3 weights from Hugging Face \cite{stable_diffusion_weights}.

We followed the same training procedure as the original implementation, and adapted the latent denosing model to get as an additional input the spatio-textual representation $ST$ with the following training loss:
\begin{equation}
    L_{\text{LDM}} = E_{t, y, z_0,\epsilon}\left[ || \epsilon - \epsilon_{\theta}(z_t, \cliptxt{y},ST, t) ||^2 \right]
\end{equation}
where $z_t$ is the noisy latent code at time step $t$ and $y$ is the corresponding text prompt. We fine-tuned only the denoising model while keeping the autoencoder and $\cliptxtd$ frozen. We used Adam optimizer \cite{kingma2014adam} with $\beta_1=0.9$ and $\beta_2=0.999$ with learning rate $1\times10^{-4}$ for $100{,}000$ iterations.

During inference, we used the DDIM \cite{song2020denoising} inference method with $50$ sampling steps.

\subsection{Spatio-Textual Representation Details}
\label{sec:spatio_textual_implementation_detials}
In order to create the spatio-textual CLIP-based representation, we used the following models:
\begin{itemize}
    \item A pre-trained ViT-L/14 \cite{dosovitskiy2020image} variant of CLIP \cite{radford2021learning} model released by OpenAI \cite{clip_implementation}.
    \item A pre-trained panoptic segmentation model R101-FPN from Detectron2 \cite{wu2019detectron2}.
\end{itemize}

During the training phase, we extracted candidate segments using R101-FPN model from the Detectron2 \cite{wu2019detectron2} codebase model and filtered the small segments that accounted for less than $5\%$ of the image area because their CLIP image embeddings are less meaningful for low-res images. Then, we randomly used $1 \le K \le 3$ segments for the formation of the spatio-textual representation.

In addition, in order to enable multi-conditional classifier-free guidance, as explained in
\Cref{sec:multi_conditional_cfg},
we dropped each of the input conditions (the global text and the spatio-textual representation) during training $10\%$ of the time (i.e., the model was trained totally unconditionally about $1\%$ of the time).

\subsection{Baselines Implementation Details}
\label{sec:baselines_implementation_details}
For the No Token Left Behind (NTLB) baseline~\cite{paiss2022no} we used the official PyTorch \cite{paszke2019pytorch} implementation \cite{ntlb_implementation}. The original model did not support global text and was mainly demonstrated on rectangular masks. In order to adapt it to our problem setting, we added a degenerate mask of all ones for the global text. Then, we used the rest of the segmentation maps as-is, along with their corresponding text prompt. For Make-A-Scene (MAS)~\cite{gafni2022make}, we followed the exact implementation details from the paper.

In addition, we used the official \DALLE~2 and Stable Diffusion online demos \cite{dalle2_demo, stable_diffusion_demo} to generate the assets for some of the figures in this paper:
\Cref{fig:elaborated_description_failure}
and \Cref{fig:additional_elaborated_description_failure} below.

\subsection{Evaluation Dataset Details}
\label{sec:dataset_implementation_details}

\input{figures/coco_qualitative_comparison/fig.tex}

As explained in
\Cref{sec:quantitative_and_qualitative_results},
we proposed to evaluate our method automatically by generating a large number of coherent inputs based on natural images. To this end, we used the COCO \cite{lin2014microsoft} validation set that contains global text captions as well as a dense segmentation map for each image. We convert the segmentation map labels by simply providing the text ``a \{label\}'' for each segment. Then, we randomly choose a subset of size $1 \le K \le 3$ segments to form the sparse input. This way, we generated $30,000$ input samples for comparison. \Cref{fig:coco_qualitative_comparison} (top row) shows a random number of generated input samples.

In addition, we provide in \Cref{fig:coco_qualitative_comparison} an additional qualitative comparison of our method against the baselines. As we can see, the latent-based variant of our method outperforms the baselines in terms of compliance with both the global and local texts, and in terms of overall image quality.

\subsection{User Study}
\label{sec:user_study_details}
As explained in
\Cref{sec:user_study},
we conducted a user study using the Amazon Mechanical Turk (AMT) platform. In each question the evaluators were asked to choose between two images in terms of (1) overall image quality, (2) text-matching to the global prompt $\tglobal$ and (3) text-matching to the local prompts of the raw spatio-textual representation $RST$. For each one of those metrics, we created 512 coherent inputs automatically from COCO validation set \cite{lin2014microsoft} as described in 
\Cref{sec:quantitative_and_qualitative_results}
and presented a pair of generated results to five raters, yielding a total of 2,560 ratings per task. For each question, the raters were asked to choose the better result of the two (according to the given criterion). We reported the majority vote percentage per question. In addition, the raters were also given the option to indicate that both models are equal, in a case which the majority vote indicated equal, or in a tie case, we divided the points equally between the evaluated models.

The questions we asked per comparison are:
\begin{itemize}
    \item For the overall quality test --- ``Which image has a better visual quality?''
    \item For the global text correspondence test --- ``Which image best matches the text: \{GLOBAL TEXT\}'', where \{GLOBAL TEXT\} is $\tglobal$.
    \item For the local text correspondence test --- we provided in addition one mask from the raw spatio-textual representation $RST$ and asked ``Which image best matches the text and the shape of the mask?''
\end{itemize}

\subsection{Inference Time and Parameters Comparison}
\input{tables/inferece_time_and_parameters_comparison.tex}

In \Cref{tab:inference_time_and_paramters_comparison} we compare the number of parameters and the inference time of the baselines and the different variants of our method. For each method, we describe its submodules and their corresponding number of parameters and inference times for a single image. As we can see, our latent-based variant is significantly faster than the rest of the baselines. In addition, it has fewer parameters than Make-A-Scene \cite{gafni2022make} and the pixel-based variant of our method.

%% file: figures/coco_qualitative_comparison/fig.tex
\begin{figure*}[ht]
    \centering
    
    \centering
    \setlength{\tabcolsep}{1pt}
    \renewcommand{\arraystretch}{0.5}
    \setlength{\ww}{0.32\columnwidth}
  
    \begin{tabular}{ccccccc}

        \rotatebox{90}{\scriptsize\phantom{A} Underlying real images} &
        \includegraphics[width=\ww,frame]{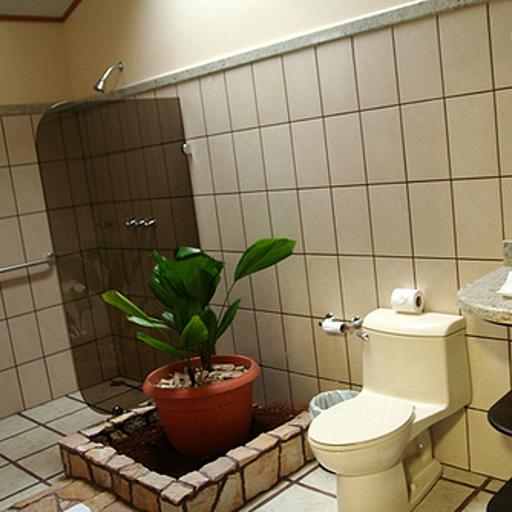} &
        \includegraphics[width=\ww,frame]{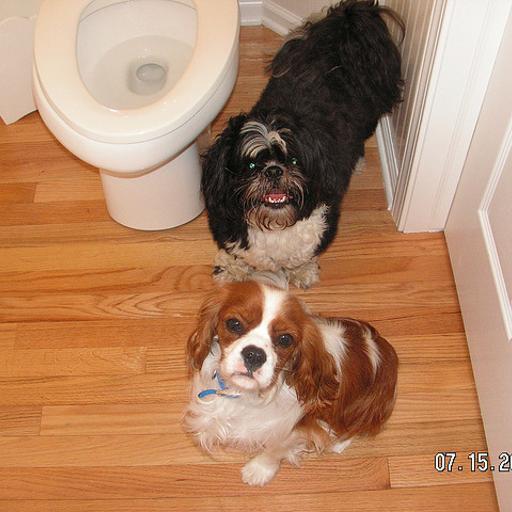} &
        \includegraphics[width=\ww,frame]{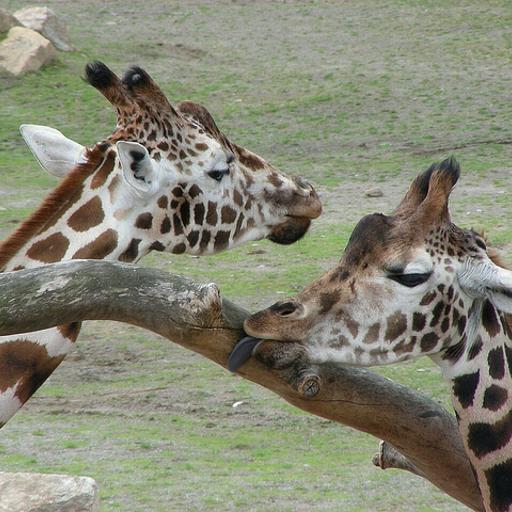} &
        \includegraphics[width=\ww,frame]{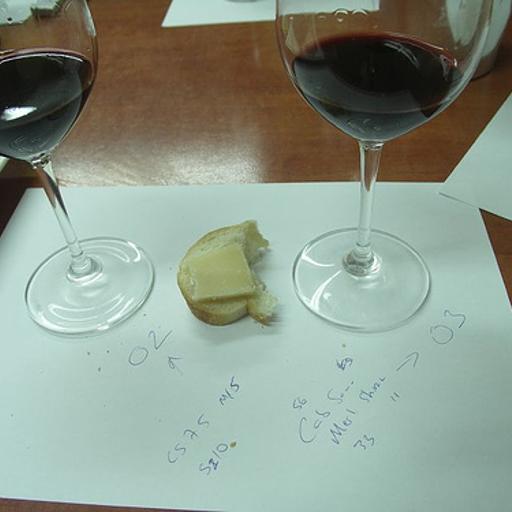} &
        \includegraphics[width=\ww,frame]{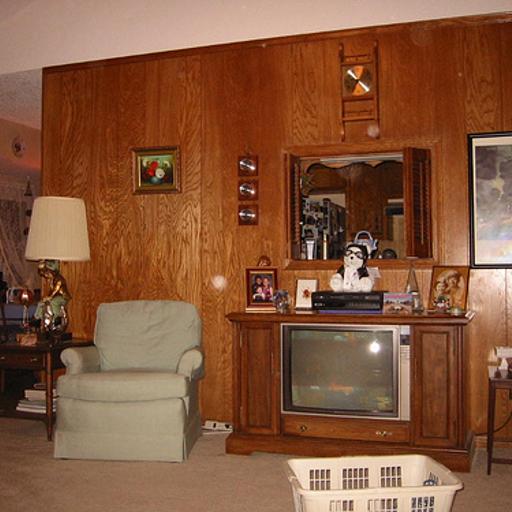} &
        \includegraphics[width=\ww,frame]{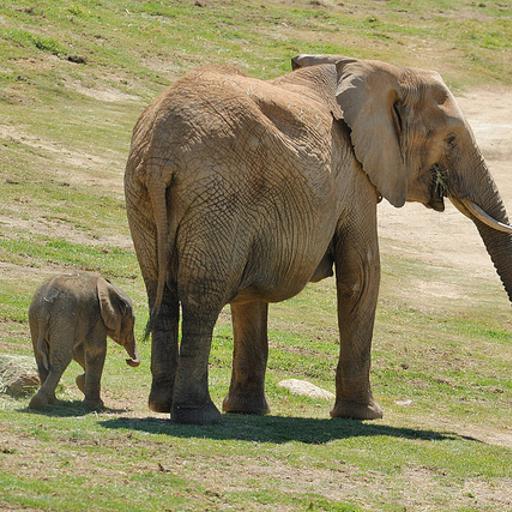}
        \\
        \\

        &
        \scriptsize{``Interior bathroom scene} &
        &
        &
        \scriptsize{``a close up of a piece} &
        \scriptsize{``A picture of a old} &
        \\

        &
        \scriptsize{with modern furnishings} &
        \scriptsize{``Two small dogs stand} &
        &
        \scriptsize{of bread near two} &
        \scriptsize{fashioned looking} &
        \scriptsize{``A big and a small}
        \\

        &
        \scriptsize{including a plant''} &
        \scriptsize{together in a bathroom''} &
        \scriptsize{``Heads of two giraffes''} &
        \scriptsize{glasses of wine''} &
        \scriptsize{living room''} &
        \scriptsize{elephant out in the sun''}
        \\

        \rotatebox{90}{\scriptsize\phantom{AAa} Extracted inputs} &
        \includegraphics[width=\ww,frame]{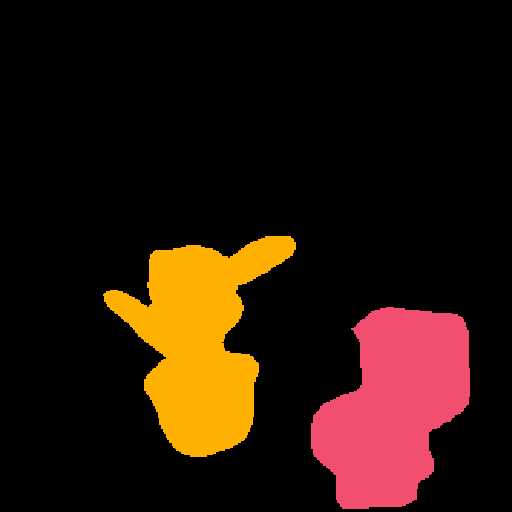} &
        \includegraphics[width=\ww,frame]{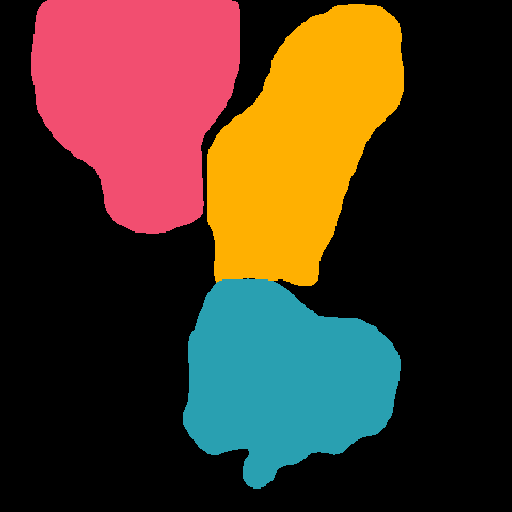} &
        \includegraphics[width=\ww,frame]{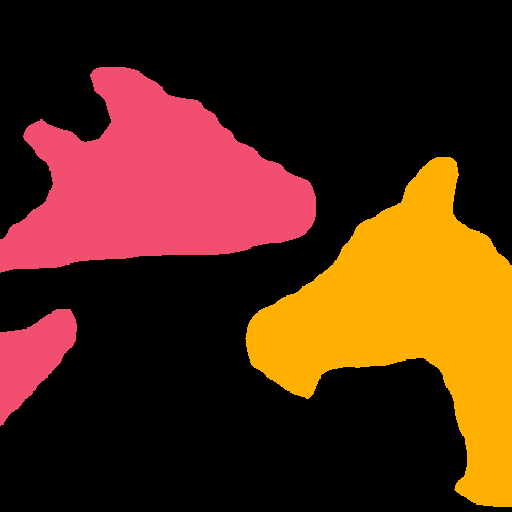} &
        \includegraphics[width=\ww,frame]{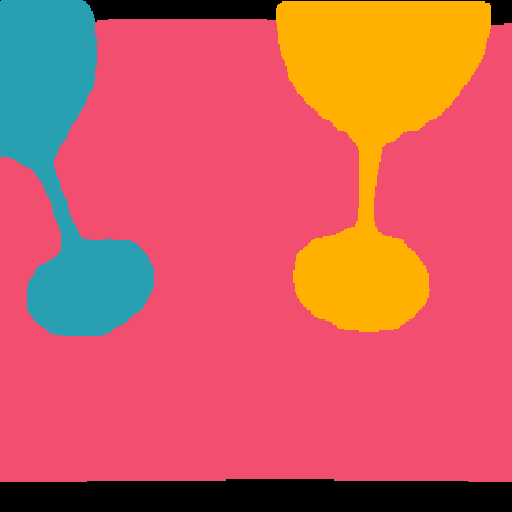} &
        \includegraphics[width=\ww,frame]{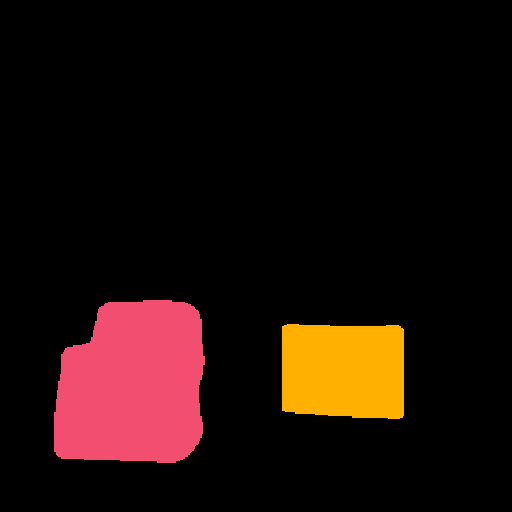} &
        \includegraphics[width=\ww,frame]{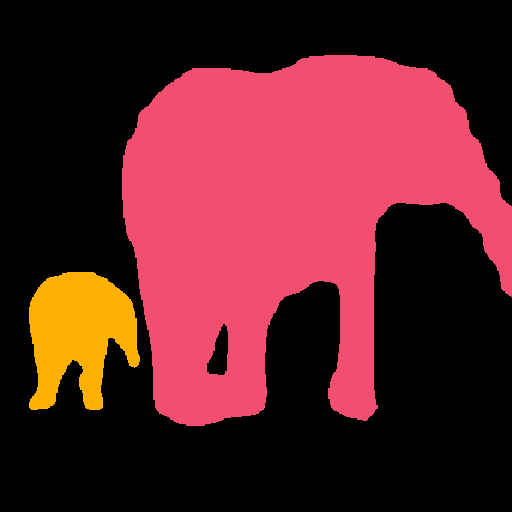}
        \\
        
        &
        \begin{tabular}{c}
            \maska{\scriptsize{``a potted plant''}} \\
            \maskb{\scriptsize{``a toilet''}} \\
            \\
            \\
        \end{tabular} &

        \begin{tabular}{c}
            \maska{\scriptsize{``a toilet''}} \\
            \maskb{\scriptsize{``a dog''}} \\
            \maskc{\scriptsize{``a dog''}} \\
            \\
        \end{tabular} &

        \begin{tabular}{c}
            \maska{\scriptsize{``a giraffe''}} \\
            \maskb{\scriptsize{``a giraffe''}} \\
            \\
            \\
        \end{tabular} &

        \begin{tabular}{c}
            \maska{\scriptsize{``dining table''}} \\
            \maskb{\scriptsize{``a wine glass''}} \\
            \maskc{\scriptsize{``a wine glass''}} \\
            \\
        \end{tabular} &

        \begin{tabular}{c}
            \maska{\scriptsize{``a chair''}} \\
            \maskb{\scriptsize{``a tv''}} \\
            \\
            \\
        \end{tabular} &

        \begin{tabular}{c}
            \maska{\scriptsize{``an elephant''}} \\
            \maskb{\scriptsize{``an elephant''}} \\
            \\
            \\
        \end{tabular}
        \\

        \rotatebox{90}{\scriptsize\phantom{AAAAa} NTLB \cite{paiss2022no}} &
        \includegraphics[width=\ww,frame]{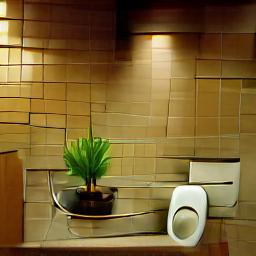} &
        \includegraphics[width=\ww,frame]{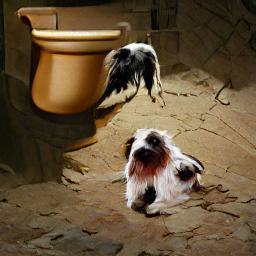} &
        \includegraphics[width=\ww,frame]{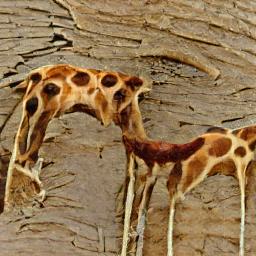} &
        \includegraphics[width=\ww,frame]{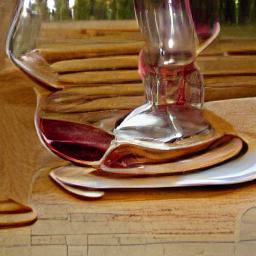} &
        \includegraphics[width=\ww,frame]{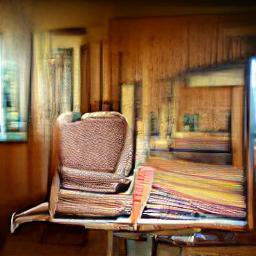} &
        \includegraphics[width=\ww,frame]{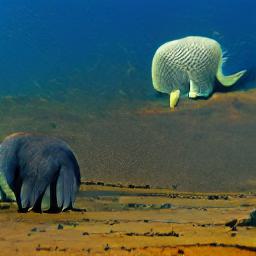}
        \\

        \rotatebox{90}{\scriptsize\phantom{AAAAa} MAS \cite{gafni2022make}} &
        \includegraphics[width=\ww,frame]{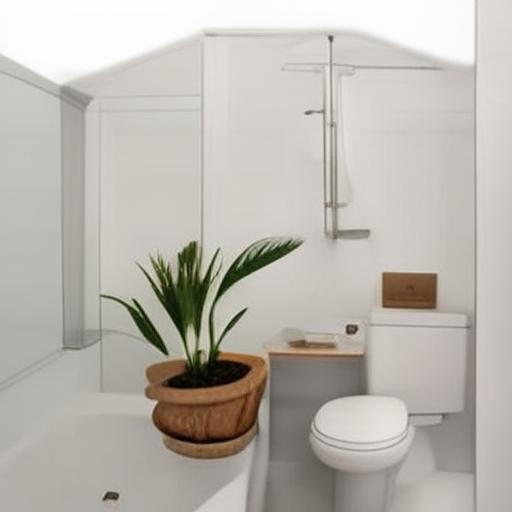} &
        \includegraphics[width=\ww,frame]{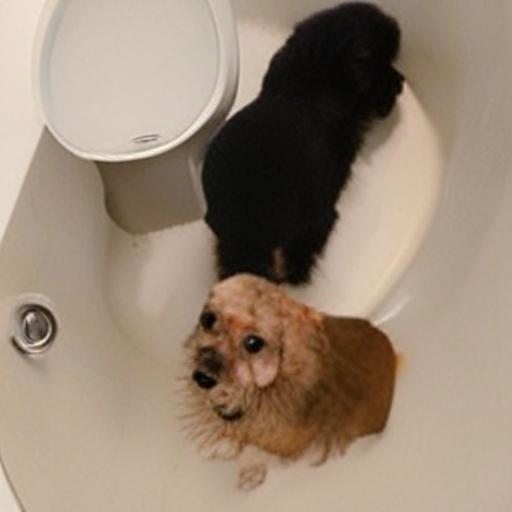} &
        \includegraphics[width=\ww,frame]{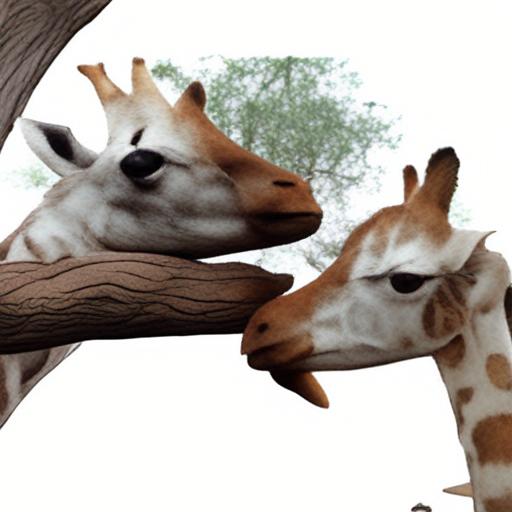} &
        \includegraphics[width=\ww,frame]{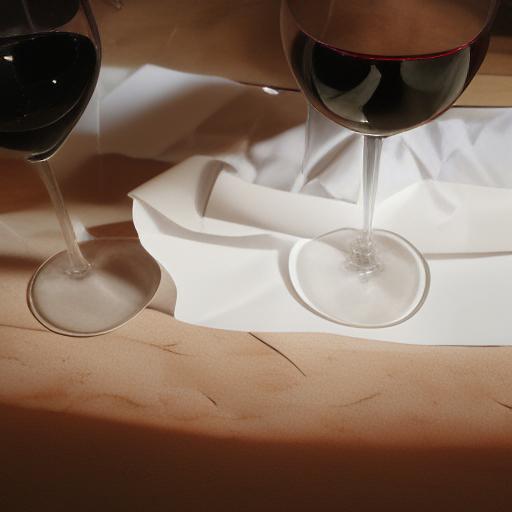} &
        \includegraphics[width=\ww,frame]{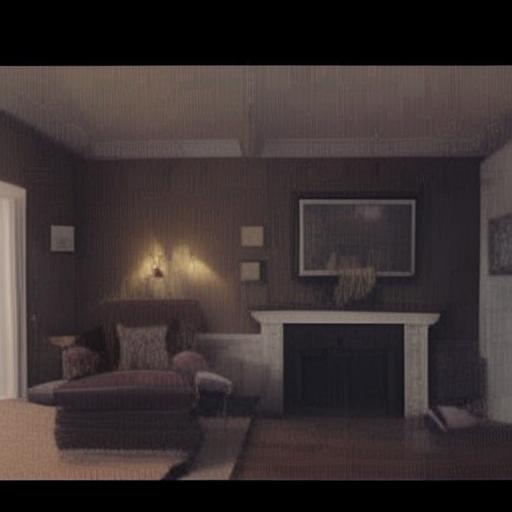} &
        \includegraphics[width=\ww,frame]{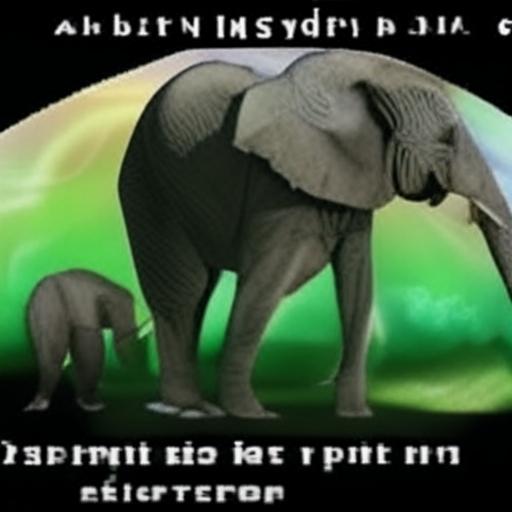}
        \\

        \rotatebox{90}{\scriptsize\phantom{A} MAS (rand-label) \cite{gafni2022make}} &
        \includegraphics[width=\ww,frame]{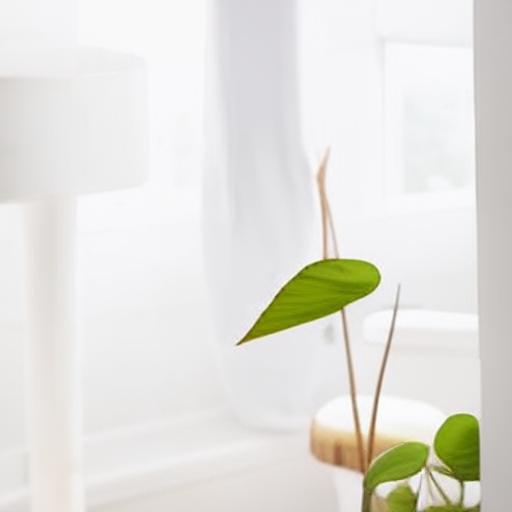} &
        \includegraphics[width=\ww,frame]{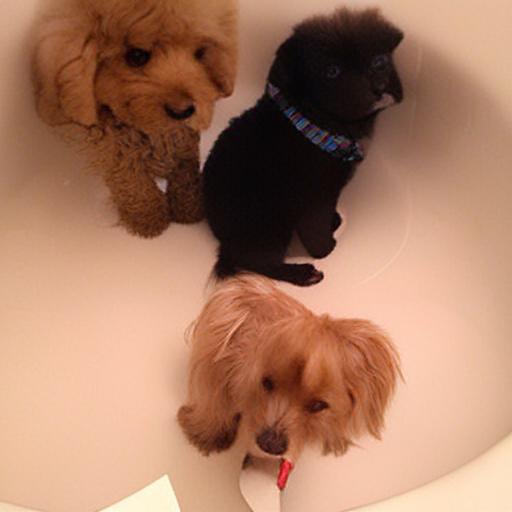} &
        \includegraphics[width=\ww,frame]{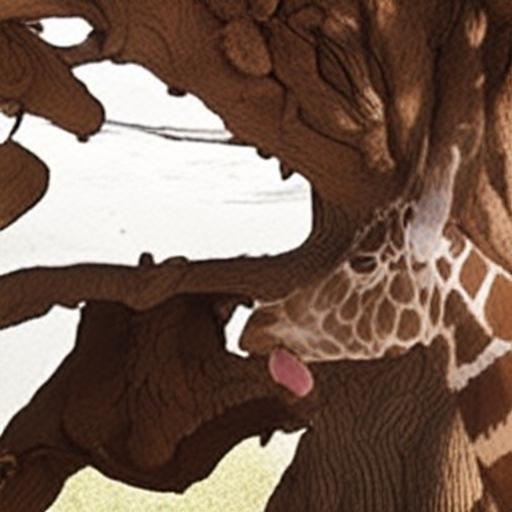} &
        \includegraphics[width=\ww,frame]{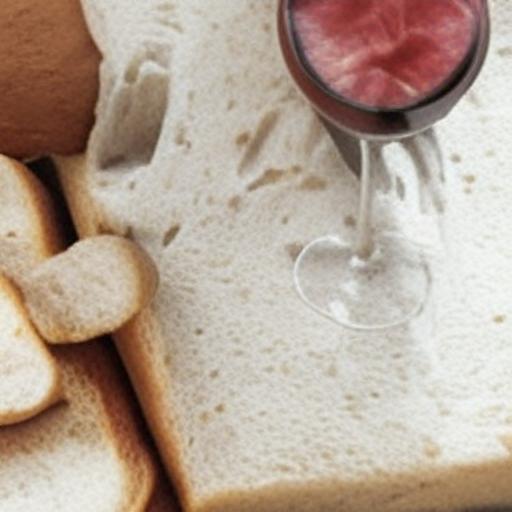} &
        \includegraphics[width=\ww,frame]{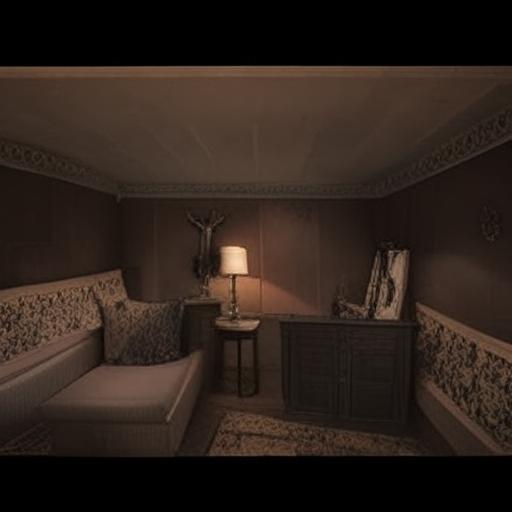} &
        \includegraphics[width=\ww,frame]{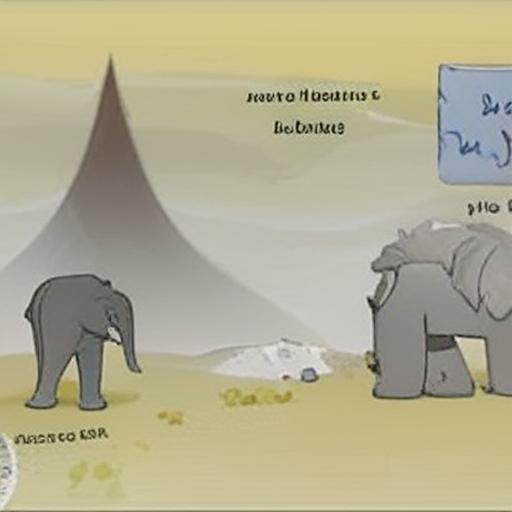}
        \\

        \rotatebox{90}{\scriptsize\phantom{AAA} \name (pixel)} &
        \includegraphics[width=\ww,frame]{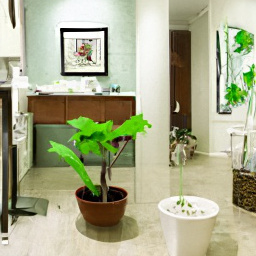} &
        \includegraphics[width=\ww,frame]{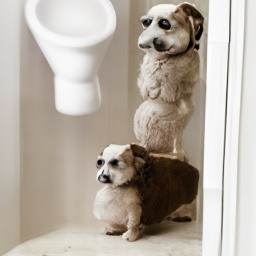} &
        \includegraphics[width=\ww,frame]{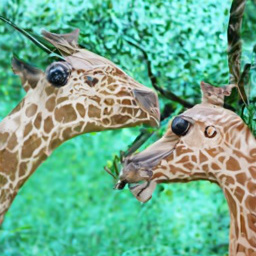} &
        \includegraphics[width=\ww,frame]{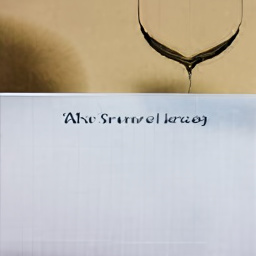} &
        \includegraphics[width=\ww,frame]{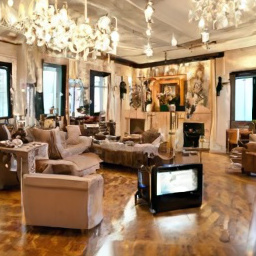} &
        \includegraphics[width=\ww,frame]{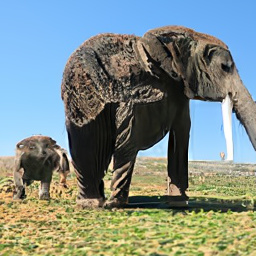}
        \\

        \rotatebox{90}{\textbf{\scriptsize{\phantom{AAA} \name (latent)}}} &
        \includegraphics[width=\ww,frame]{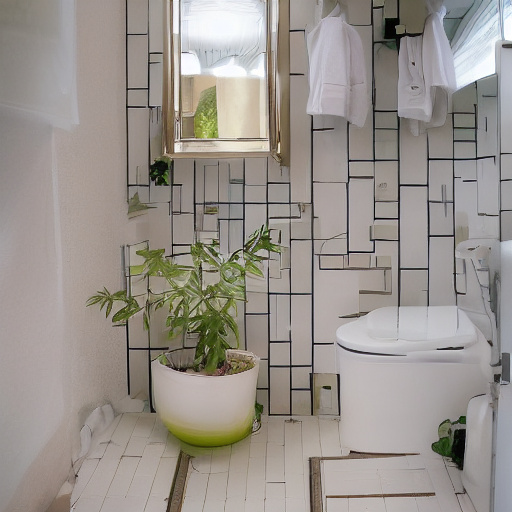} &
        \includegraphics[width=\ww,frame]{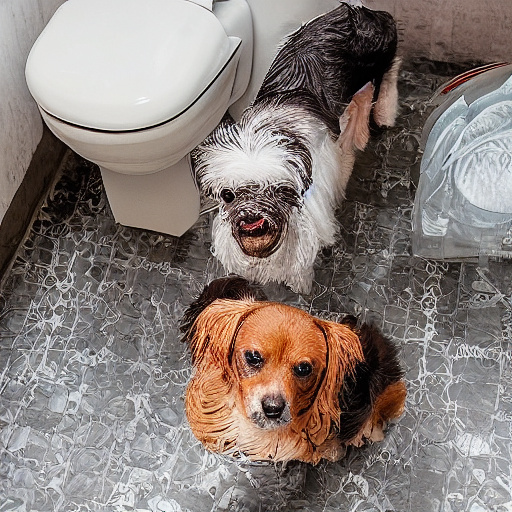} &
        \includegraphics[width=\ww,frame]{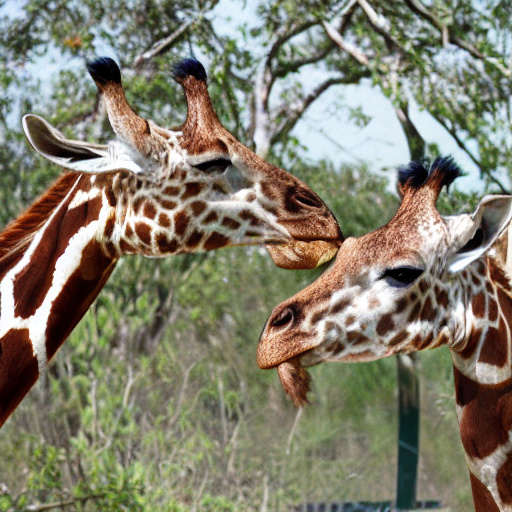} &
        \includegraphics[width=\ww,frame]{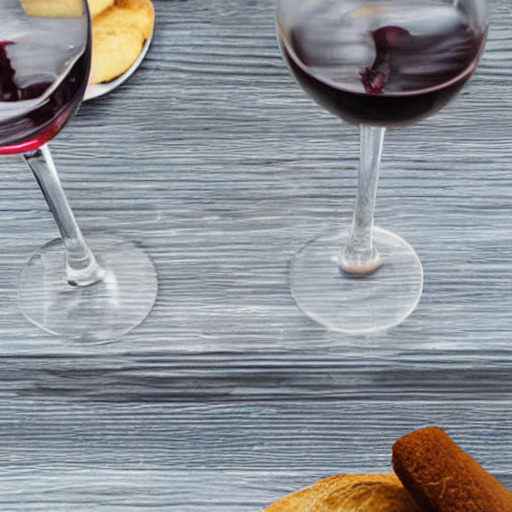} &
        \includegraphics[width=\ww,frame]{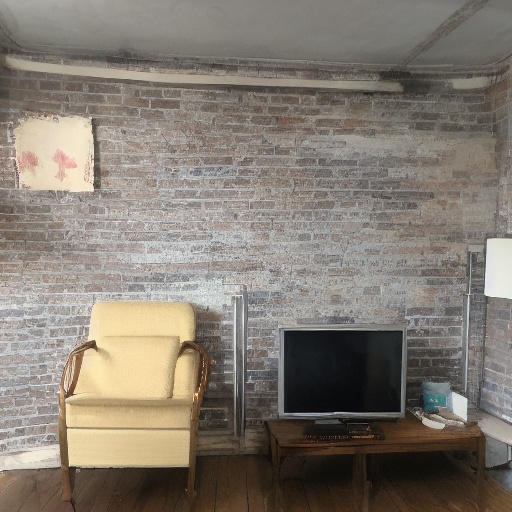} &
        \includegraphics[width=\ww,frame]{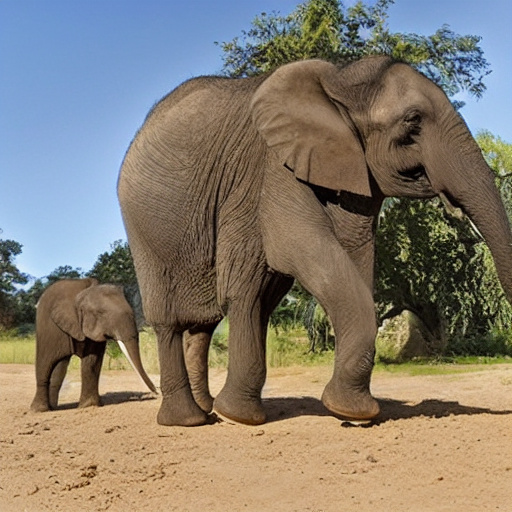}
        \\
    \end{tabular}
    
    \caption{\textbf{Qualitative comparison on automatically generated inputs:} in order to create realistic inputs comparison, we utilized a segmentation dataset \cite{lin2014microsoft} to create inputs (second row) that are based on real images (top row). Given those inputs, we generate images using the baselines and the two variants of our method. As we can see, our latent-based variant of our method  outperforms the baselines in terms of compliance with both the global and local texts, and in overall image quality.}
    \label{fig:coco_qualitative_comparison}
\end{figure*}

%% file: tables/inferece_time_and_parameters_comparison.tex
\begin{table*}[t]
    \begin{center}
        \begin{adjustbox}{width=2.0\columnwidth}
            \begin{tabular}{lccc}
                \toprule
        
                \textbf{Method} &
                \textbf{Consisting submodules} &
                \textbf{\# Parameters (B)} &
                \textbf{Inference time (sec)}
                \\
                
                \midrule
                No Token Left Behind \cite{paiss2022no} &
                CLIP (ViT-B/32) + model &
                0.15B + 0.08B = 0.23B &
                326 sec
                \\

                Make-A-Scene \cite{gafni2022make} &
                VAE + model &
                0.002B + 4B = 4.002B &
                76 sec
                \\

                \midrule
                \name (pixel) w/o prior &
                CLIP + model + upsample &
                0.43B + 3.5B + 1B = 4.93B &
                50 sec
                \\

                \name (pixel) w prior &
                CLIP + prior + model + upsample &
                0.43B + 1.3B + 3.5B + 1B = 6.23B &
                52 sec
                \\

                \name (latent) w/o prior &
                CLIP + model &
                0.43B + 0.87B = 1.3B &
                5 sec
                \\

                \textbf{\name (latent) w prior} &
                CLIP + prior + model &
                0.43B + 1.3B + 0.87B = 2.6B &
                7 sec
                \\

                \bottomrule
            \end{tabular}
        \end{adjustbox}
    \caption{\textbf{Inference time and parameters:} we compare the number of parameters and the inference time across the baselines and the different variants (including ablations) of our method. As we can see, \name (latent) is significantly faster than the rest of the baselines. In addition, it has fewer parameters than Make-A-Scene \cite{gafni2022make} and the \name (pixel) variant of our method. The inference times reported were computed for a single image on a single V100 NVIDIA GPU.}
    \label{tab:inference_time_and_paramters_comparison}
    \end{center}
\end{table*}

%% file: sections/appendix/additional_experiments.tex
\section{Additional Experiments}
\label{sec:additional_experiments}

\input{tables/metrics_comparison_mas_variants.tex}

In this section, we provide additional experiments that we have conducted. In \Cref{sec:manual_baselines} we describe manual baselines that may be used to generate images with free-form textual scene control. Then, in \Cref{sec:general_mas_variant} we present a general variant for Make-A-Scene and compare it against our method.
Finally, in \Cref{sec:local_prompts_concatenation_trick} we describe and demonstrate the local prompts concatenation trick.

\subsection{Manual Baselines}
\label{sec:manual_baselines}

\input{figures/elaborated_description_failure/fig_additional.tex}
\input{figures/interactive_editing_baseline/dog_beach.tex}
\input{figures/interactive_editing_baseline/bear_avocado.tex}

In order to generate an image with free-form textual scene control, one may try to operate existing methods in various manual ways. For example, as demonstrated in
\Cref{sec:introduction},
trying to achieve this task using an elaborated text prompt is 
overly optimistic.
We provided additional examples in \Cref{fig:additional_elaborated_description_failure}.

Another possible option to achieve this goal it to combine a text-to-image models with a local text-driven editing method \cite{avrahami2022blended,avrahami2022blended_latent,ramesh2022hierarchical} in a multi-stage approach: at the first stage, the user can utilize a text-to-image model to generate the background of the scene, e.g. Stable Diffusion or \DALLE~2. Then, the user can sequentially mask the desired areas and provide the local prompts using a local text-driven editing method, e.g. Blended Latent Diffusion or \DALLE~2. \Cref{fig:dog_beach_interactive_editing,fig:bear_avocado_interactive_editing} demonstrate that even though these approaches may place the object in the desired location, the composition of the entire scene looks less natural, because the model does not take into account the entire scene at the first stage, so the generated image of the background may not be easily edited for the desired composition. In addition, the objects correspond less to the local masks, especially in the \DALLE~2 case. Furthermore, the multi-stage approach is more cumbersome from the user point of view, because of its iterative nature.

Lastly, another approach is to utilize a sketch-to-image generation, as demonstrated in SDEdit \cite{meng2021sdedit}: the user can provide a \emph{dense} color sketch of the scene, then noise it to a certain noise level, and denoise it iteratively using a text-to-image diffusion model. However, this user interface is different from our interface in the following aspects: (1) the user need to provide a color for each pixel, whereas in our method the user may provide a local prompt that describe other aspects that are not color-related only. In addition, (2) in this approach, the user needs to construct a \emph{dense} segmentation map of the entire scene in advance, whereas in our method the user can provide only some of the areas and let the machine infer the rest. It is not clear how this can be done in the sketch-based approach.

\subsection{Random Label Make-A-Scene Variant}
\label{sec:general_mas_variant}

\input{figures/qualitative_comparison/fig_mas.tex}

In
\Cref{sec:quantitative_and_qualitative_results},
we presented a way to adapt Make-A-Scene (MAS)~\cite{gafni2022make} to our problem setting. The original Make-A-Scene work proposed a method that conditions a text-to-image model on a global text $\tglobal{}$ and a \emph{dense} segmentation map with \emph{fixed labels}. Hence, we converted it to our problem setting of \emph{sparse} segmentation map with \emph{open-vocabulary local prompts} by concatenating the local texts of the raw spatio-textual representation $RST$ into the global text prompt $\tglobal$.

However, the above version requires the user to provide an additional label for each segment, which is more than needed by our method and NTLB \cite{paiss2022no} baseline. Hence, we experimented with a more general version of Make-A-Scene we termed MAS (rand-label) that assigns a random label for each segment, instead of asking the user to provide an additional label. In \Cref{fig:qualitative_comparison_mas_variants} we can see that this method is able to match the local prompts even with random labels. We also evaluated this method numerically using the same automatic metrics and user study protocol described in
\Cref{sec:experimets}.
As can be seen in \Cref{tab:metrics_comparison_mas_variants}, this method achieves inferior results compared to the version that uses the ground-truth labels in both the automatic evaluation and the user study.

\subsection{Local Prompts Concatenation Trick}
\label{sec:local_prompts_concatenation_trick}

\input{figures/local_prompts_concat/fig.tex}

As described in
\Cref{sec:multi_conditional_cfg},
we noticed that the texts in the image-text pairs dataset contain elaborate descriptions of the entire scene, whereas we aim to ease the use for the end-user and remove the need to provide an elaborate global prompt in addition to the local ones, i.e., to not require the user to repeat the same information twice. Hence, in order to reduce the domain gap between the training data and the input at inference time, we perform the following simple trick: we concatenate the local prompts to the global prompt at inference time separated by commas. \Cref{fig:local_prompts_concat} demonstrates that this concatenation yields images that correspond to the local prompts better.

%% file: tables/metrics_comparison_mas_variants.tex
\begin{table*}[t]
    \begin{center}
        \setlength{\tabcolsep}{2pt}
        \begin{tabular}{lcccc|ccc}
            
            &
            \multicolumn{4}{c}{Automatic Metrics} &
            \multicolumn{3}{c}{User Study}
            \\

            \toprule
    
            \textbf{Method} & 
            Global $\downarrow$ & 
            Local $\downarrow$ &
            Local $\uparrow$ &
            FID  $\downarrow$&
            Visual &
            Global &
            Local
            \\

            &
            distance & 
            distance & 
            IOU &
            & 
            quality & 
            match  &
            match
            \\
            
            \midrule

            MAS \cite{gafni2022make} &
            0.7591 &
            0.7835 &
            \textbf{0.2984} &
            21.367 &
            81.25\% &
            70.61\% &
            57.81\%
            \\

            MAS (rand-label) \cite{gafni2022make} &
            0.7796 &
            0.7861 &
            0.1544 &
            29.593 &
            82.81\% &
            81.44\% &
            76.85\%
            \\

            \midrule

            \name (latent) &
            \textbf{0.7436} &
            \textbf{0.7795} &
            0.2842 &
            \textbf{6.7721} &
            - &
            - &
            -
            \\

            \bottomrule
        \end{tabular}
    \caption{\textbf{Metrics comparison:} We evaluated our method against the baselines using automatic metrics (left) and human ratings (right). The results for the human ratings (right) are reported as the percentage of the majority vote raters that preferred our latent-based variant of our method over the baseline. As we can see, MAS (rand-label) achieves inferior results compared to the standard version of MAS, in both the automatic metrics and the user study.}
    \label{tab:metrics_comparison_mas_variants}
    \end{center}
\end{table*}

%% file: figures/elaborated_description_failure/fig_additional.tex
\begin{figure*}[t]    
    \centering
    \setlength{\tabcolsep}{1pt}
    \renewcommand{\arraystretch}{0.5}
    \setlength{\ww}{0.5\columnwidth}
  
    \begin{tabular}{cccc}

        ``at the desert'' &
        \name &
        Stable Diffusion &
        \DALLE~2
        \\

        \includegraphics[width=\ww,frame]{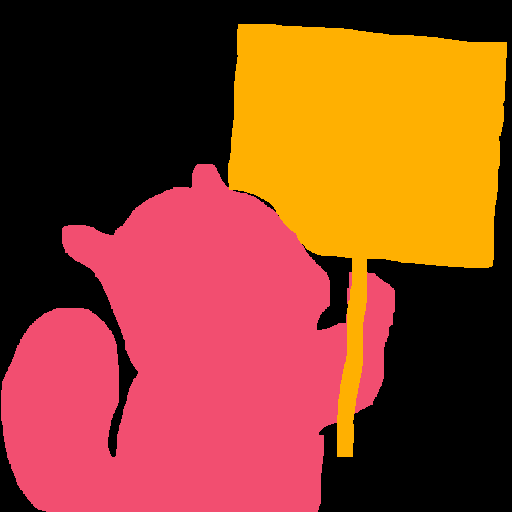} &
        \includegraphics[width=\ww,frame]{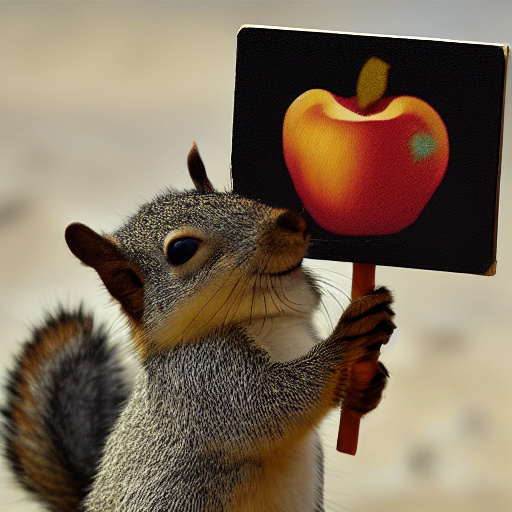}
        \phantom{.}
        &
        \includegraphics[width=\ww,frame]{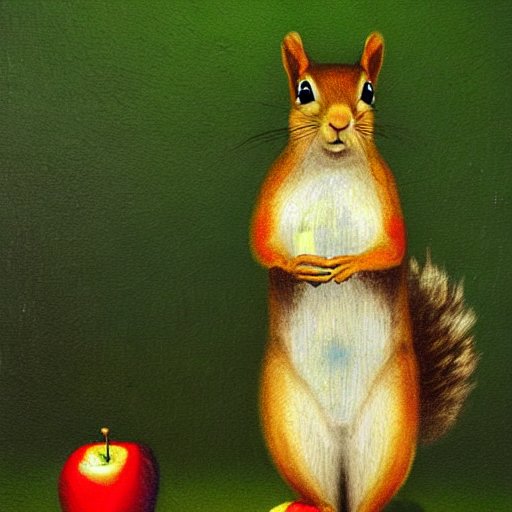} &
        \includegraphics[width=\ww,frame]{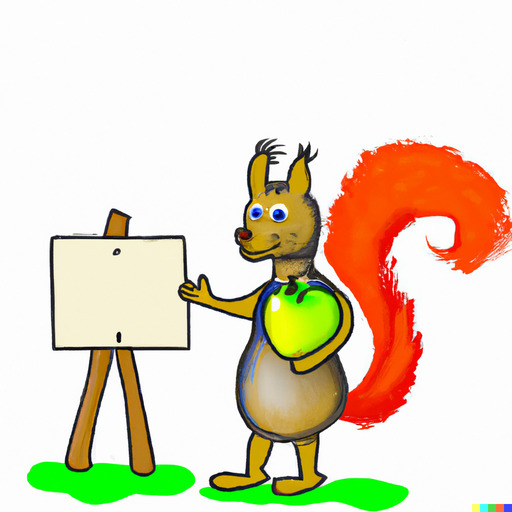}
        \\

        \maska{``a squirrel''} &&
        \multicolumn{2}{c}{``A squirrel stands on the left side of the frame, holding} \\

        \maskb{``a sign with an apple} &&
        \multicolumn{2}{c}{a sign with an apple painting on the right side of the frame''} \\

        \maskb{painting''} &&
        \\
        \\
        \\

        ``near the river'' &
        \name &
        Stable Diffusion &
        \DALLE~2
        \\

        \includegraphics[width=\ww,frame]{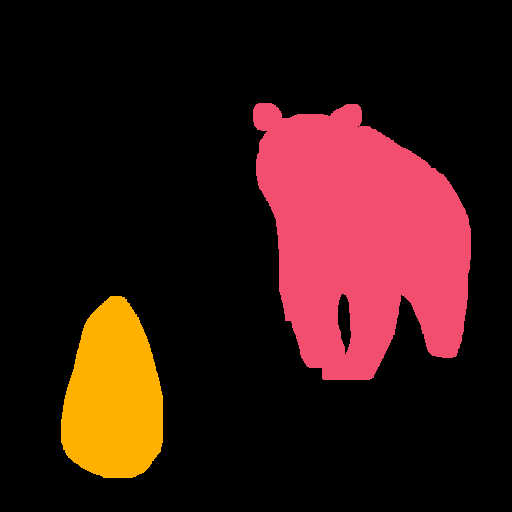} &
        \includegraphics[width=\ww,frame]{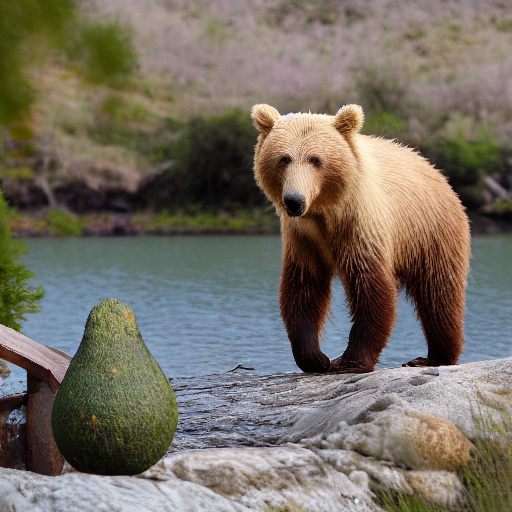}
        \phantom{.}
        &
        \includegraphics[width=\ww,frame]{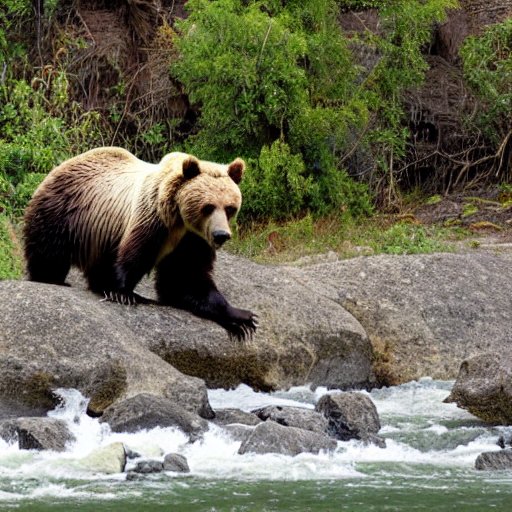} &
        \includegraphics[width=\ww,frame]{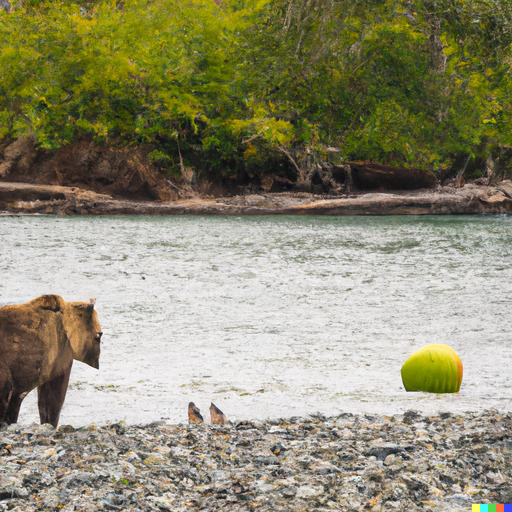}
        \\

        \maska{``a grizzly bear''} &&
        \multicolumn{2}{c}{``A grizzly bear stands near a river on the right side of the} \\

        \maskb{``a huge avocado''} &&
        \multicolumn{2}{c}{frame, looking at the huge avocado in left side of the frame''} \\
        \\
        \\
        \\

        ``a painting'' &
        \name &
        Stable Diffusion &
        \DALLE~2
        \\

        \includegraphics[width=\ww,frame]{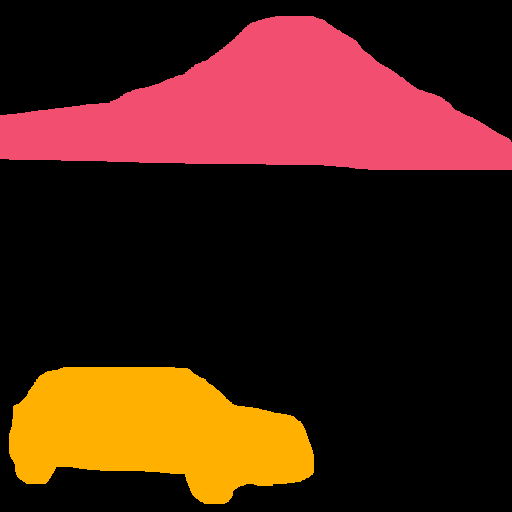} &
        \includegraphics[width=\ww,frame]{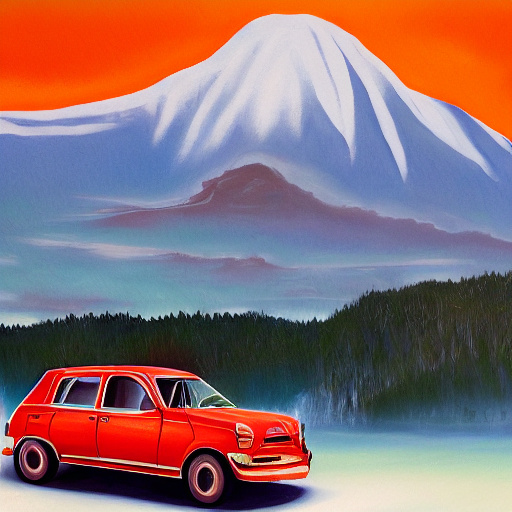}
        \phantom{.}
        &
        \includegraphics[width=\ww,frame]{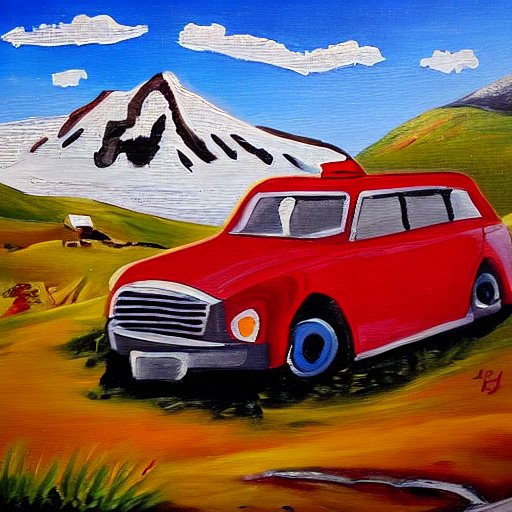} &
        \includegraphics[width=\ww,frame]{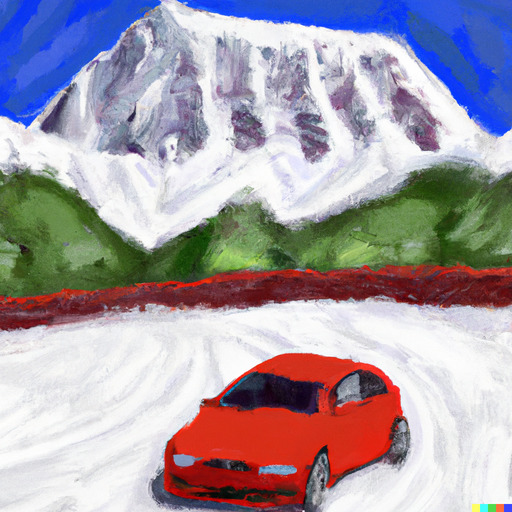}
        \\

        \maska{``a snowy mountain''} &&
        \multicolumn{2}{c}{``A painting of a snowy mountain in the} \\

        \maskb{``a red car''} &&
        \multicolumn{2}{c}{background and a red car in the front''} \\
        \\
    \end{tabular}
    
    \caption{\textbf{Lack of fine-grained spatial control:} A user with a specific mental image (left) can easily generate it with a \name representation but will struggle to do so with traditional text-to-image models (right) \cite{rombach2022high,ramesh2021zero}.}
    \label{fig:additional_elaborated_description_failure}
\end{figure*}

%% file: figures/interactive_editing_baseline/dog_beach.tex
\begin{figure*}[t]
    \centering
    
    \centering
    \setlength{\tabcolsep}{1pt}
    \renewcommand{\arraystretch}{0.5}
    \setlength{\ww}{0.39\columnwidth}
  
    \begin{tabular}{cccccc}

        &
        ``at the beach''
        \\

        \rotatebox{90}{\phantom{AAAa} \name} &
        \includegraphics[width=\ww,frame]{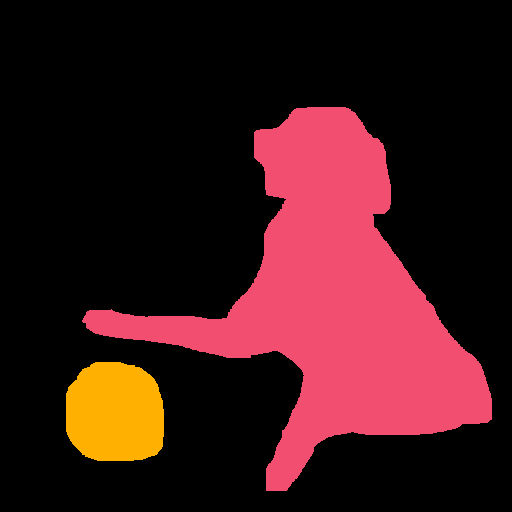} &
        \includegraphics[width=\ww,frame]{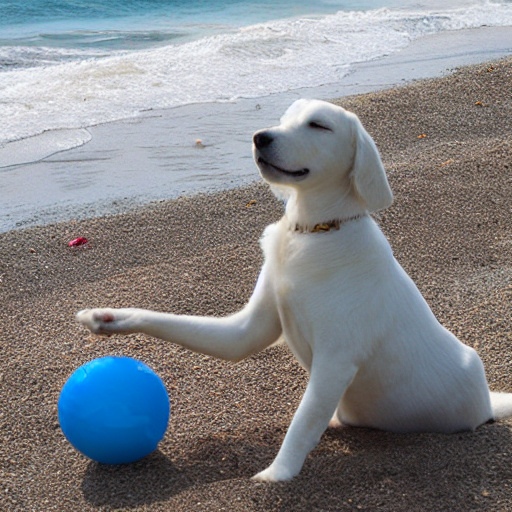}
        \\

        &
        \maska{``a white Labrador''}
        \\

        &
        \maskb{``a blue ball''} 
        \\
        \\
        \\
        \\

        &
        ``at the beach'' &
        ``a white Labrador'' &
        &
        ``a blue ball''
        \\

        \rotatebox{90}{\phantom{a} SD \cite{rombach2022high} + BLD \cite{avrahami2022blended_latent}} &
        \includegraphics[width=\ww,frame]{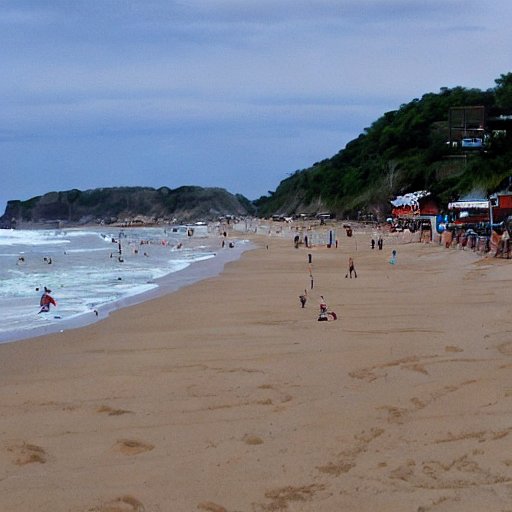} &
        \includegraphics[width=\ww,frame]{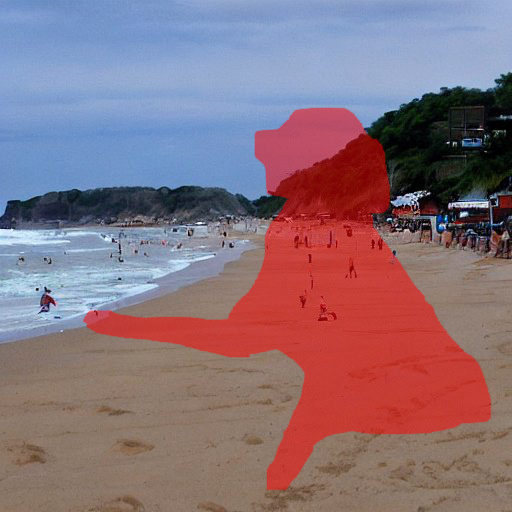} &
        \includegraphics[width=\ww,frame]{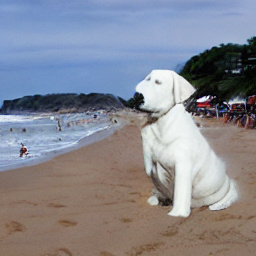} &
        \includegraphics[width=\ww,frame]{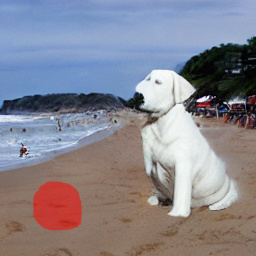} &
        \includegraphics[width=\ww,frame]{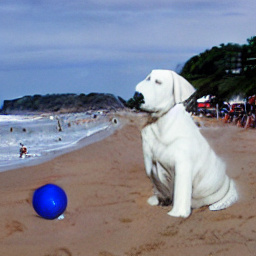}
        \\
        \\

        \rotatebox{90}{\phantom{AA} \DALLE~2 \cite{ramesh2022hierarchical}} &
        \includegraphics[width=\ww,frame]{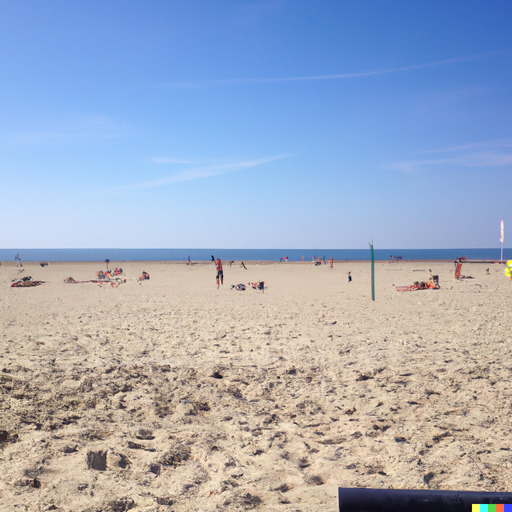} &
        \includegraphics[width=\ww,frame]{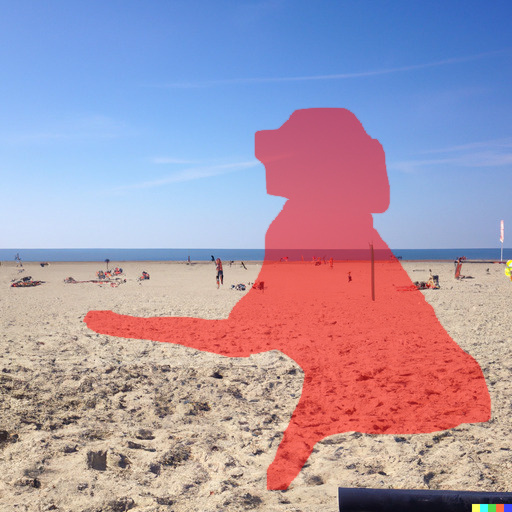} &
        \includegraphics[width=\ww,frame]{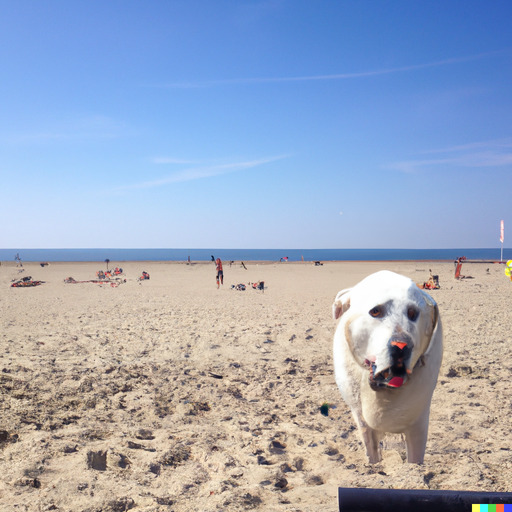} &
        \includegraphics[width=\ww,frame]{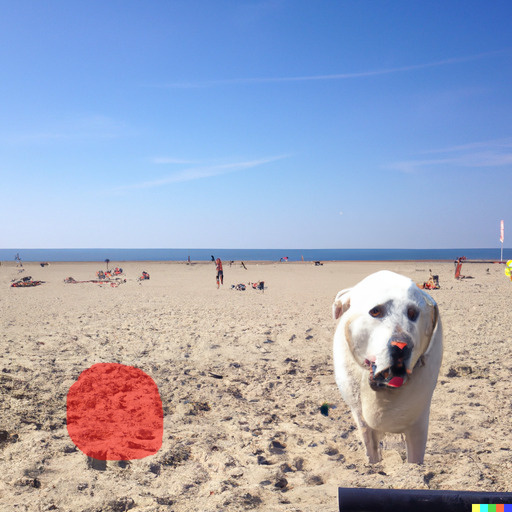} &
        \includegraphics[width=\ww,frame]{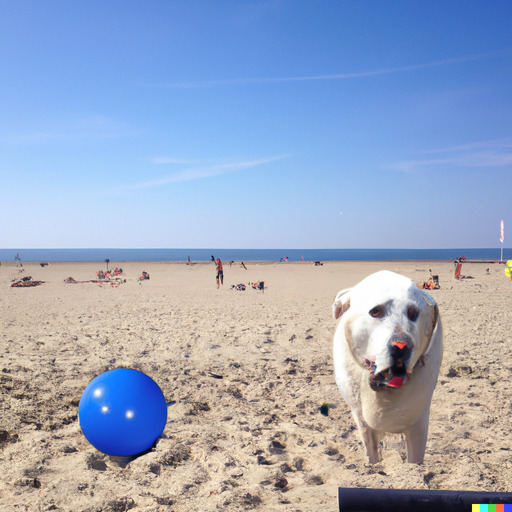}
        \\

        &
        Stage 1 &
        Stage 2 &
        Stage 3 &
        Stage 4 &
        Stage 5

    \end{tabular}
    
    \caption{\textbf{Interactive editing baseline:} An alternative way to achieve image generation with free-form textual scene control as in our method (first row) is by iterative editing: at the first stage, the user can utilize a text-to-image model to generate the background of the scene, e.g. Stable Diffusion (second row) or \DALLE~2 (third row). Then, the user can sequentially mask the desired areas and provide the local prompts using a local text-driven editing method, e.g. Blended Latent Diffusion (second row) or \DALLE~2 (third row).}
    \label{fig:dog_beach_interactive_editing}
\end{figure*}

%% file: figures/interactive_editing_baseline/bear_avocado.tex
\begin{figure*}[t]
    \centering
    
    \centering
    \setlength{\tabcolsep}{1pt}
    \renewcommand{\arraystretch}{0.5}
    \setlength{\ww}{0.39\columnwidth}
  
    \begin{tabular}{cccccc}

        &
        ``near a river''
        \\

        \rotatebox{90}{\phantom{AAAa} \name} &
        \includegraphics[width=\ww,frame]{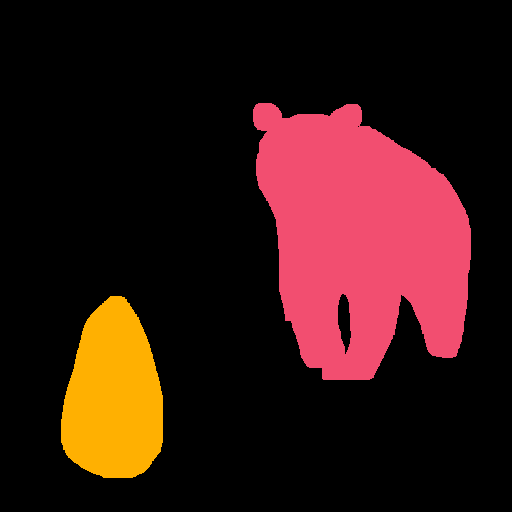} &
        \includegraphics[width=\ww,frame]{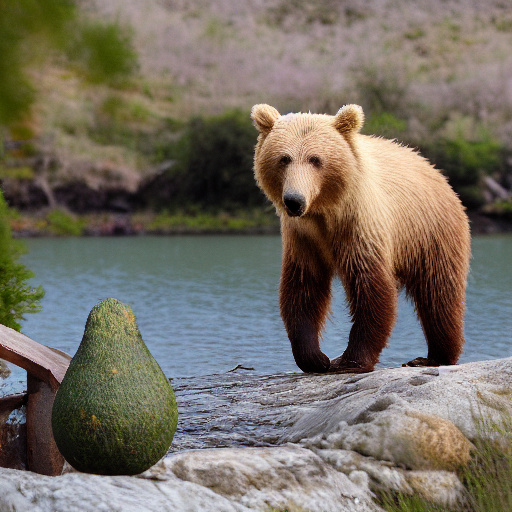}
        \\

        &
        \maska{``a grizzly bear''}
        \\

        &
        \maskb{``a huge avocado''}
        \\
        \\
        \\
        \\

        &
        ``near a river'' &
        ``a grizzly bear'' &
        &
        ``a huge avocado''
        \\

        \rotatebox{90}{\phantom{a} SD \cite{rombach2022high} + BLD \cite{avrahami2022blended_latent}} &
        \includegraphics[width=\ww,frame]{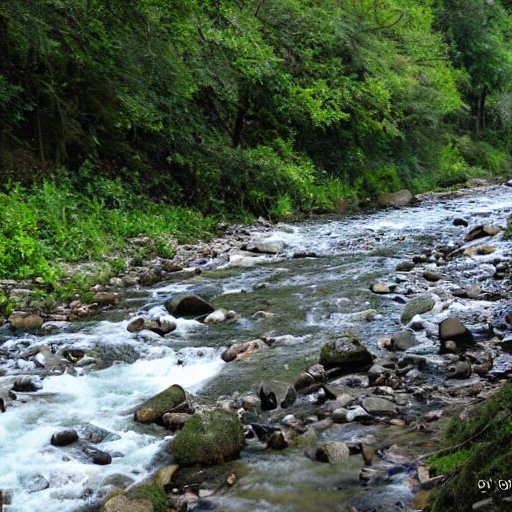} &
        \includegraphics[width=\ww,frame]{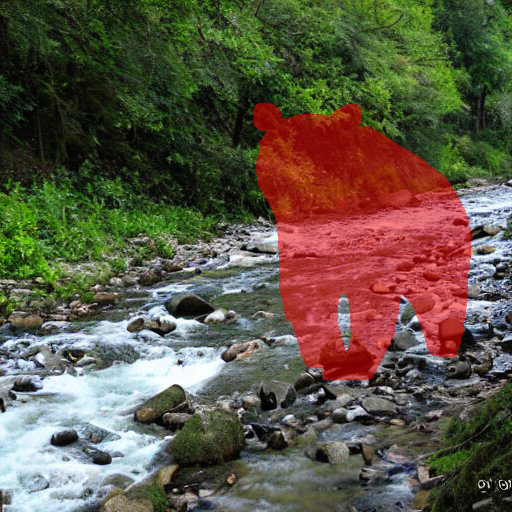} &
        \includegraphics[width=\ww,frame]{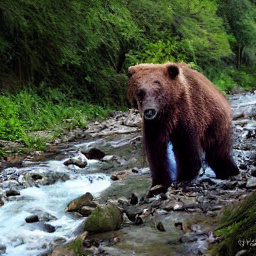} &
        \includegraphics[width=\ww,frame]{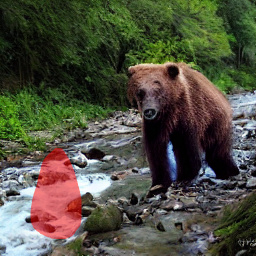} &
        \includegraphics[width=\ww,frame]{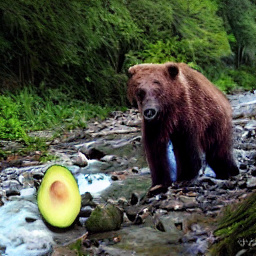}
        \\
        \\

        \rotatebox{90}{\phantom{AA} \DALLE~2 \cite{ramesh2022hierarchical}} &
        \includegraphics[width=\ww,frame]{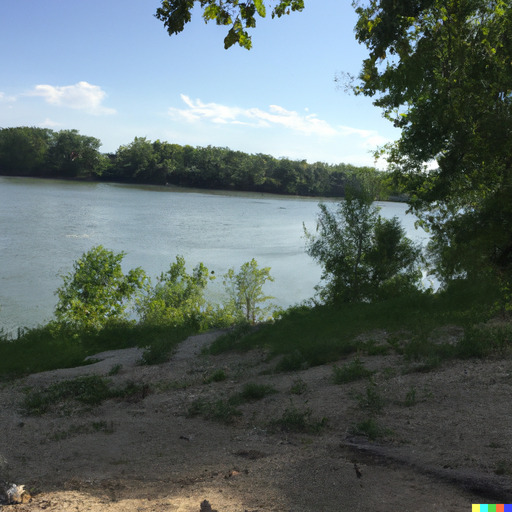} &
        \includegraphics[width=\ww,frame]{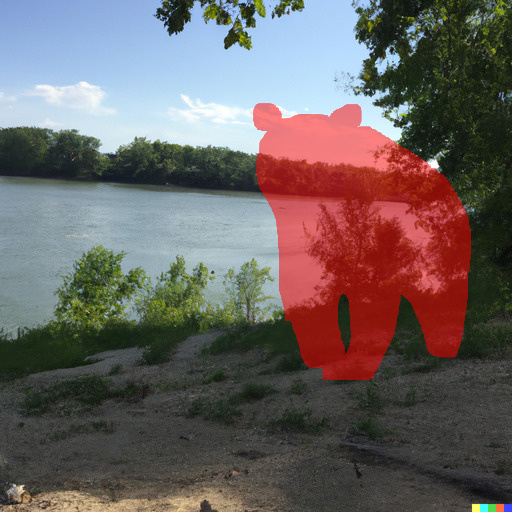} &
        \includegraphics[width=\ww,frame]{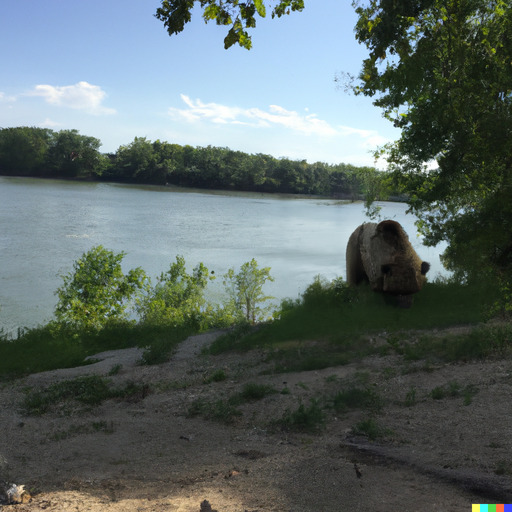} &
        \includegraphics[width=\ww,frame]{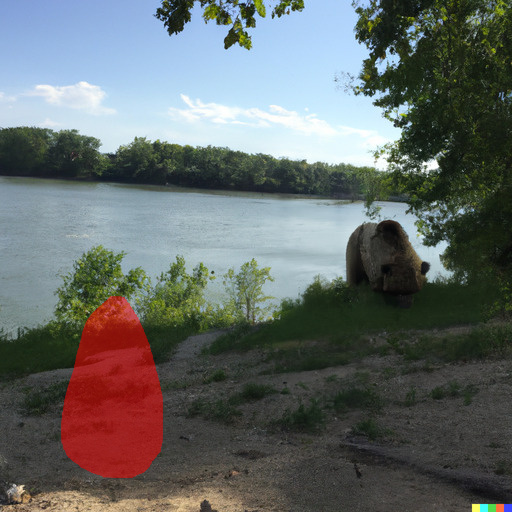} &
        \includegraphics[width=\ww,frame]{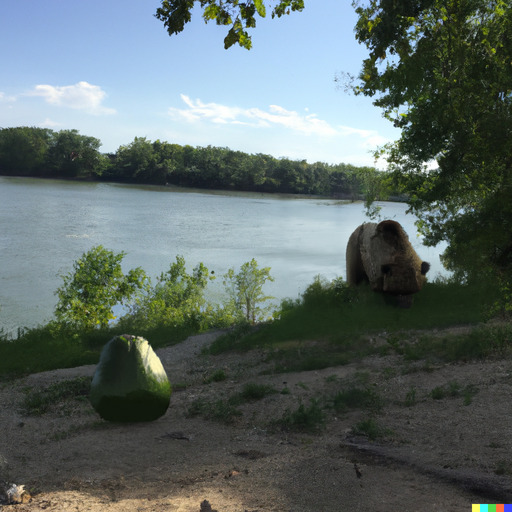}
        \\

        &
        Stage 1 &
        Stage 2 &
        Stage 3 &
        Stage 4 &
        Stage 5

    \end{tabular}
    
    \caption{\textbf{Interactive editing baseline:} An alternative way to achieve image generation with free-form textual scene control as in our method (first row) is by iterative editing: at the first stage, the user can utilize a text-to-image model to generate the background of the scene, e.g. Stable Diffusion (second row) or \DALLE~2 (third row). Then, the user can sequentially mask the desired areas and provide the local prompts using a local text-driven editing method, e.g. Blended Latent Diffusion (second row) or \DALLE~2 (third row).}
    \label{fig:bear_avocado_interactive_editing}
\end{figure*}

%% file: figures/qualitative_comparison/fig_mas.tex
\begin{figure*}[ht]
    \centering
    
    \centering
    \setlength{\tabcolsep}{1pt}
    \renewcommand{\arraystretch}{0.5}
    \setlength{\ww}{0.28\columnwidth}
  
    \begin{tabular}{cccccccc}
        
        &&&
        \scriptsize{``a sunny day after} &
        &
        \scriptsize{``a bathroom with} &
        \\

        &
        \scriptsize{``near a lake''} &
        \scriptsize{``a painting''} &
        \scriptsize{the snow''} &
        \scriptsize{``on a table''} &
        \scriptsize{an artificial light''} &
        \scriptsize{``near a river''} &
        \scriptsize{``at the desert''}
        \\

        \rotatebox{90}{\scriptsize\phantom{AAAAa} Inputs} &
        \includegraphics[width=\ww,frame]{figures/qualitative_comparison/assets/elephant/vis.jpg} &
        \includegraphics[width=\ww,frame]{figures/qualitative_comparison/assets/car_mountain/vis.png} &
        \includegraphics[width=\ww,frame]{figures/qualitative_comparison/assets/dogs/vis.png} &
        \includegraphics[width=\ww,frame]{figures/qualitative_comparison/assets/mug_plate/vis.png} &
        \includegraphics[width=\ww,frame]{figures/qualitative_comparison/assets/bathroom/vis.png} &
        \includegraphics[width=\ww,frame]{figures/qualitative_comparison/assets/bear_avocado/vis.png} &
        \includegraphics[width=\ww,frame]{figures/qualitative_comparison/assets/squirrel/vis.png}
        \\
        
        &
        \begin{tabular}{c}
            \maska{\scriptsize{``a black elephant''}} \\ 
            \\
            \\
            \\
            \\
        \end{tabular} &

        \begin{tabular}{c}
            \maska{\scriptsize{``a snowy mountain''}} \\ 
            \maskb{\scriptsize{``a red car''}} \\
            \\
            \\
            \\
        \end{tabular} &

        \begin{tabular}{c}
            \maska{\scriptsize{``a Husky dog''}} \\
            \maskb{\scriptsize{``a German }} \\
            \maskb{\scriptsize{Shepherd dog''}} \\
            \\
            \\
        \end{tabular} &

        \begin{tabular}{c}
            \maska{\scriptsize{``a mug''}} \\
            \maskb{\scriptsize{``a white plate }} \\
            \maskb{\scriptsize{with cookies''}} \\
            \\
            \\
        \end{tabular} &

        \begin{tabular}{c}
            \maska{\scriptsize{``a mirror''}} \\
            \maskb{\scriptsize{``a white sink''}} \\
            \maskc{\scriptsize{``a vase with}} \\
            \maskc{\scriptsize{red flowers''}} \\
            \\
        \end{tabular} &

        \begin{tabular}{c}
            \maska{\scriptsize{``a grizzly bear''}} \\
            \maskb{\scriptsize{``a huge avocado''}} \\
            \\
            \\
            \\
        \end{tabular} &

        \begin{tabular}{c}
            \maska{\scriptsize{``a squirrel''}} \\
            \maskb{\scriptsize{``a sign an }} \\
            \maskb{\scriptsize{apple painting''}} \\
            \\
            \\
        \end{tabular}

        \\

        \rotatebox{90}{\scriptsize\phantom{AAA} MAS \cite{gafni2022make}} &
        \includegraphics[width=\ww,frame]{figures/qualitative_comparison/assets/elephant/mas_fixed.jpg} &
        \includegraphics[width=\ww,frame]{figures/qualitative_comparison/assets/car_mountain/mas_fixed.jpg} &
        \includegraphics[width=\ww,frame]{figures/qualitative_comparison/assets/dogs/mas_fixed.jpg} &
        \includegraphics[width=\ww,frame]{figures/qualitative_comparison/assets/mug_plate/mas_fixed.jpg} &
        \includegraphics[width=\ww,frame]{figures/qualitative_comparison/assets/bathroom/mas_fixed.jpg} &
        \includegraphics[width=\ww,frame]{figures/qualitative_comparison/assets/bear_avocado/mas_fixed.jpg} &
        \includegraphics[width=\ww,frame]{figures/qualitative_comparison/assets/squirrel/mas_fixed.jpg}
        \\

        \rotatebox{90}{\scriptsize\phantom{a} MAS (rand-label) \cite{gafni2022make}} &
        \includegraphics[width=\ww,frame]{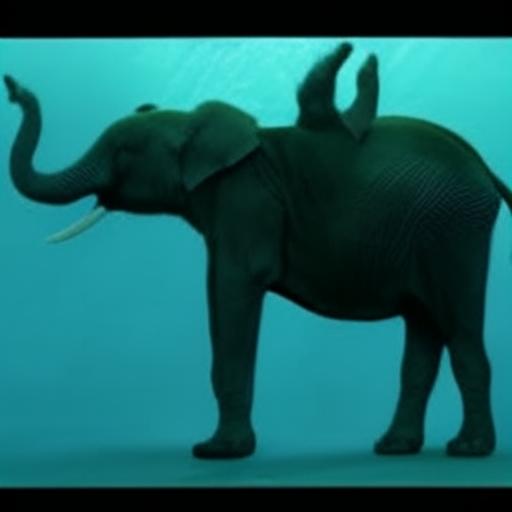} &
        \includegraphics[width=\ww,frame]{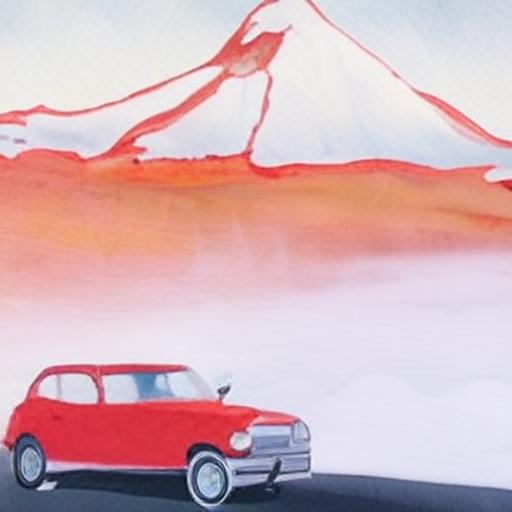} &
        \includegraphics[width=\ww,frame]{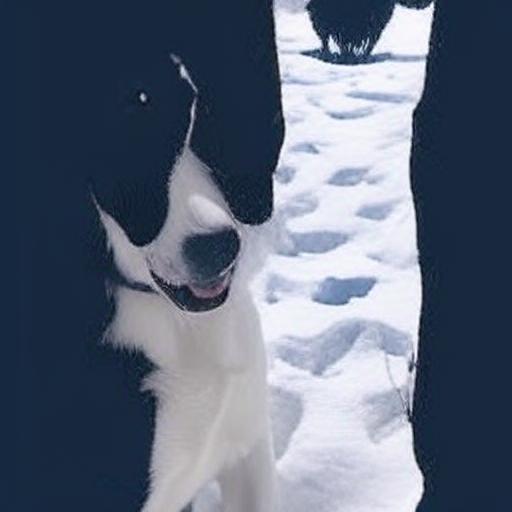} &
        \includegraphics[width=\ww,frame]{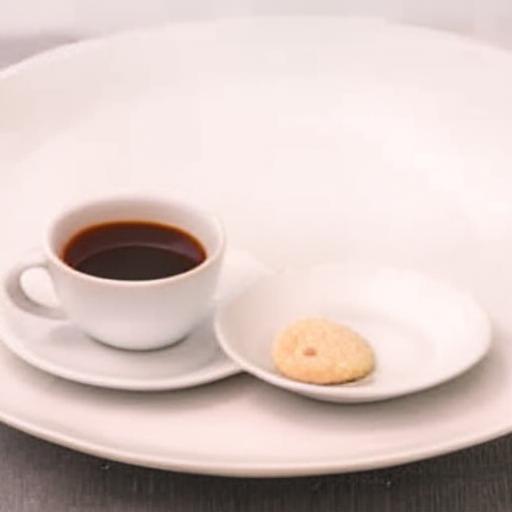} &
        \includegraphics[width=\ww,frame]{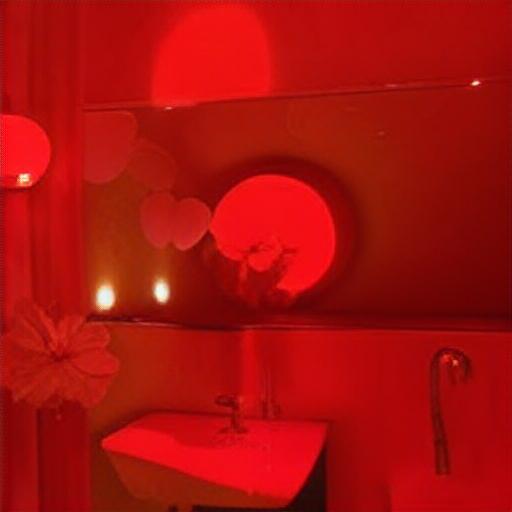} &
        \includegraphics[width=\ww,frame]{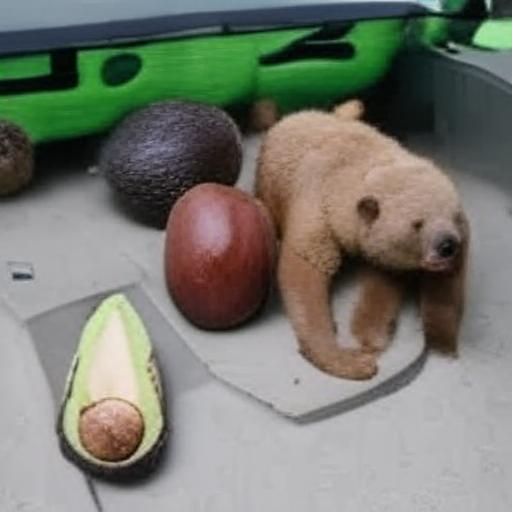} &
        \includegraphics[width=\ww,frame]{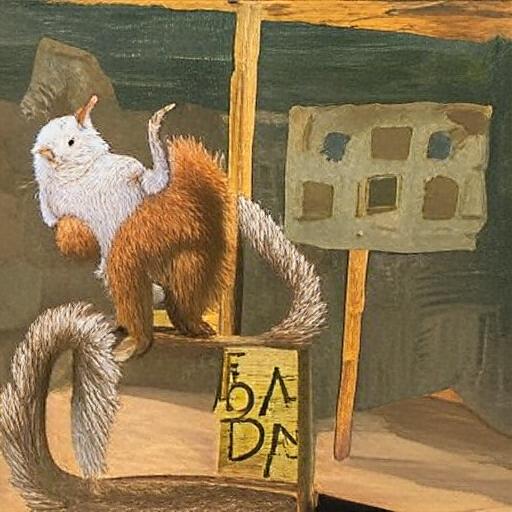}
        \\

        \rotatebox{90}{\scriptsize\phantom{AA} \name (latent)} &
        \includegraphics[width=\ww,frame]{figures/qualitative_comparison/assets/elephant/ours/5.jpg} &
        \includegraphics[width=\ww,frame]{figures/qualitative_comparison/assets/car_mountain/ours/3.jpg} &
        \includegraphics[width=\ww,frame]{figures/qualitative_comparison/assets/dogs/ours/4.jpg} &
        \includegraphics[width=\ww,frame]{figures/qualitative_comparison/assets/mug_plate/ours/1.jpg} &
        \includegraphics[width=\ww,frame]{figures/qualitative_comparison/assets/bathroom/ours/1.jpg} &
        \includegraphics[width=\ww,frame]{figures/qualitative_comparison/assets/bear_avocado/ours/3.jpg} &
        \includegraphics[width=\ww,frame]{figures/qualitative_comparison/assets/squirrel/ours/2.jpg}
        \\
    \end{tabular}
    
    \caption{\textbf{Qualitative comparison of Make-A-Scene variants:} Given the inputs (top row), we generate images using the two variants of Make-A-Scene (adapted to our task as described in \Cref{sec:general_mas_variant}) and our latent-based method. As we can see, \name (latent) outperforms these baselines in terms of compliance with both the global and local texts, and in overall image quality.}
    \label{fig:qualitative_comparison_mas_variants}
\end{figure*}

%% file: figures/local_prompts_concat/fig.tex
\begin{figure*}[t]
    \centering
    \setlength{\tabcolsep}{1pt}
    \renewcommand{\arraystretch}{0.5}
    \setlength{\ww}{0.6\columnwidth}
  
    \begin{tabular}{ccc}

        ``at the beach''
        \\

        \includegraphics[width=\ww,frame]{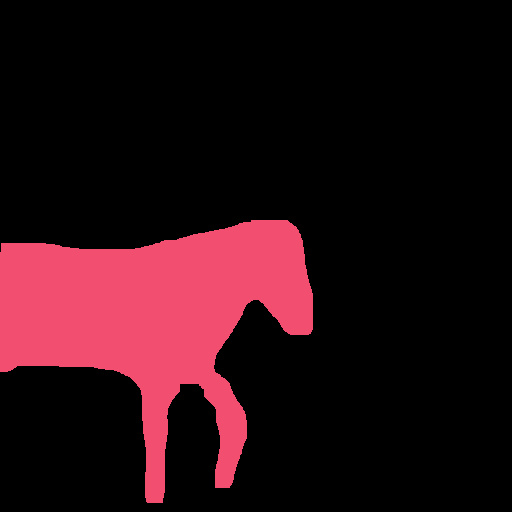} &
        \includegraphics[width=\ww,frame]{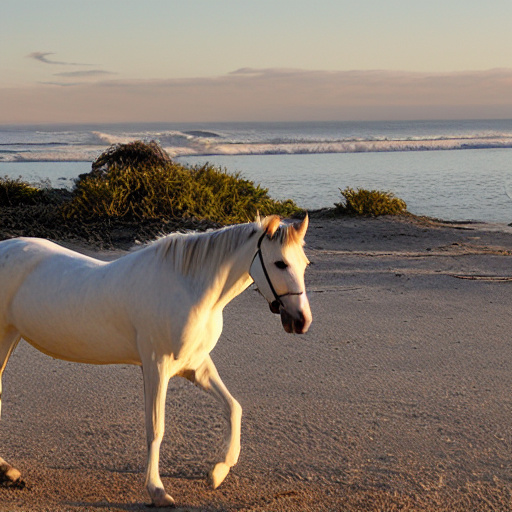} &
        \includegraphics[width=\ww,frame]{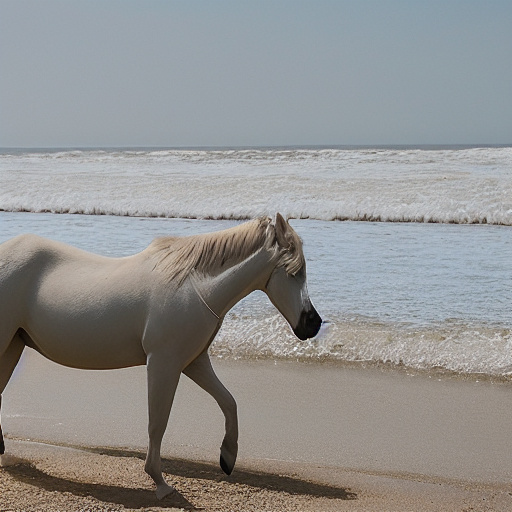}
        \\
        \maska{``a white horse''} &
        with local prompt concat &
        without local prompt concat
        \\
        \\
        \\

        ``in the forest''
        \\

        \includegraphics[width=\ww,frame]{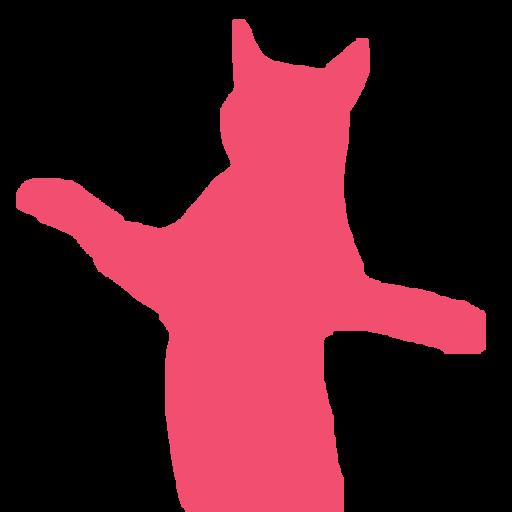} &
        \includegraphics[width=\ww,frame]{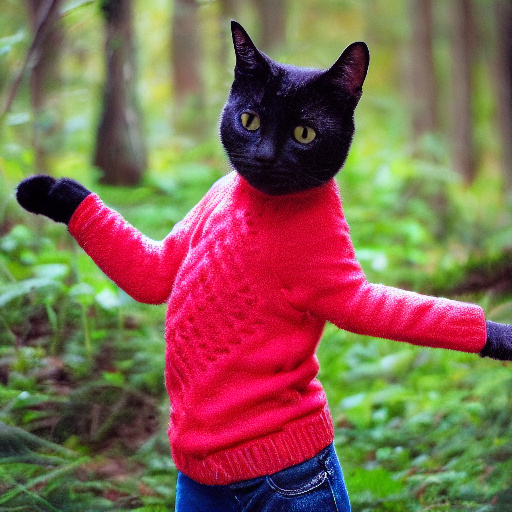} &
        \includegraphics[width=\ww,frame]{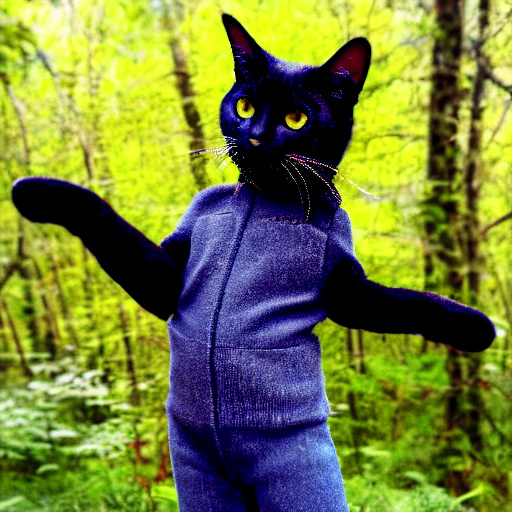}
        \\
        \maska{``a black cat with a red} &
        with local prompt concat &
        without local prompt concat
        \\ 
        \maska{sweater and a blue jeans''} \\
        \\
        \\

        ``near a river''
        \\

        \includegraphics[width=\ww,frame]{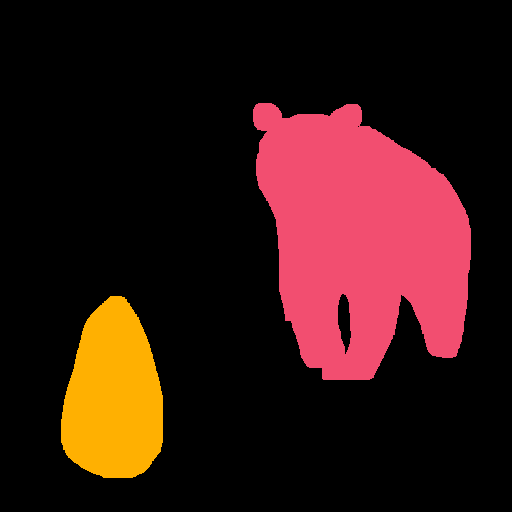} &
        \includegraphics[width=\ww,frame]{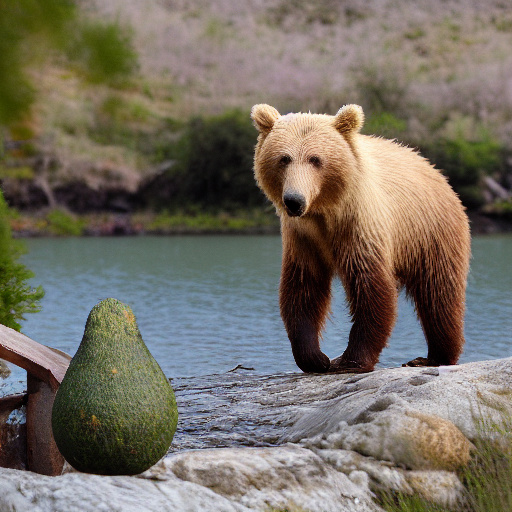} &
        \includegraphics[width=\ww,frame]{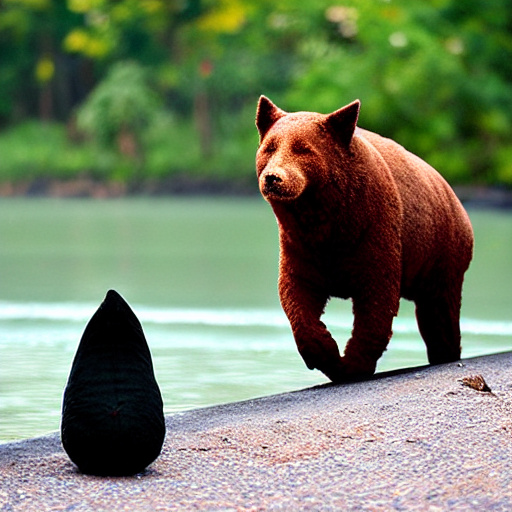}
        \\
        \maska{``a grizzly bear''} &
        with local prompt concat &
        without local prompt concat \\ 
        \maskb{``a huge avocado''} \\
        \\
    \end{tabular}
    
    \caption{\textbf{Local prompts concatenation:} concatenating the local text prompts to the global prompt during inference mitigates the train-inference gap and enables better alignment between the generated images and the local prompts.}
    \label{fig:local_prompts_concat}
\end{figure*}

%% file: sections/appendix/additional_related_work.tex
\section{Additional Related Work}
\label{sec:additional_related_work}
\textbf{Image-to-image translation:} Pix2Pix \cite{isola2017image, wang2018high} utilized a conditional GAN \cite{goodfellow2014gans,mirza2014conditional} to generate images from a paired image-segmentation dataset, which was later extended to the unpaired cased in CycleGAN \cite{zhu2017unpaired}. UNIT \cite{liu2017unsupervised} proposed to translate between domains using a shared latent space, which was extended to the multimodal \cite{huang2018multimodal} and few-shot \cite{liu2019few} cases. SPADE \cite{park2019semantic} introduced spatially-adaptive normalization to achieve better results in segmentation-to-image task. However, all of these works, do not enable editing with a free-form text description.

\textbf{Layout-to-image generation:} The seminal paper of Reed \etal \cite{reed2016learning} generated images conditioned on location and attributes and managed to show controllability over single-instance images, but generating complex scenes was not demonstrated. Later works extended it to an entire layout \cite{zhao2019image, sun2019image, sylvain2021object, sun2021learning}. However, these methods do not support fine-grained control using free-form text prompts. Other methods \cite{hong2018inferring, li2019object, hinz2018generating, hinz2020semantic} proposed to condition the layout also on a global text, but they do not propose a fine-grained free-form control for each instance in the scene. In \cite{pavllo2020controlling} an additional segmentation mask was introduced to control the shape of the instances in the scene, but they do not enable fine-grained free-form control for each instance separately. Recently \cite{frolov2022dt2i} proposed to condition a GAN model on free-form captions and location bounding boxes, and showed promising results on synthetic datasets' generation, in contrast, we focus on fine-grained segmentation masks to control the shape (instead of coarse bounding boxes), and on generating natural images instead of synthetic ones.

Concurrently to our work, eDiff-I \cite{balaji2022ediffi} presented a new text-to-image model that consists of an ensemble of expert denoising networks, each specializing in a specific noise interval. More related to our work, they proposed a training-free method, named paint-with-words, that enables users to specify the spatial locations of objects, by manipulating the cross-attention maps that correspond to the input tokens that they want to generate. Their method supports only rough segmentation maps, whereas our method focuses on the fine segmentation maps input case.